\documentclass[]{hy}

\usepackage{wrapfig}
\usepackage{tabularx}
\usepackage{textcomp}
\usepackage{stfloats}
\usepackage{url}
\usepackage{verbatim}
\usepackage{titlesec}
\usepackage{tocloft}
\usepackage{adjustbox}
\usepackage{multirow}
\usepackage{pifont}
\usepackage{tikz}
\usepackage{comment}
\usepackage{amsmath,amssymb} 
\usepackage{colortbl}  
\usepackage{color}
\usepackage{booktabs} 
\usepackage{natbib} 
\setcitestyle{square,comma,numbers,sort&compress}
\usepackage{graphicx}    
\usepackage{subcaption} 
\usepackage{multirow} 
\usepackage{booktabs} 
\usepackage{subcaption} 
\RequirePackage{xspace}
\makeatletter
\DeclareRobustCommand\onedot{\futurelet\@let@token\@onedot}
\def\@onedot{\ifx\@let@token.\else.\null\fi\xspace}

\makeatother

\definecolor{adptorange}{RGB}{248, 205, 172}
\definecolor{cmpblue}{RGB}{189, 215, 238}
\definecolor{cmpblue}{RGB}{189, 215, 238}

\definecolor{our_red}{RGB}{232,157,160}
\definecolor{our_blue}{RGB}{136,206,230}
\definecolor{our_orange}{RGB}{246,200,168}
\definecolor{our_green}{RGB}{178,211,164}

\definecolor{attn_code0}{RGB}{247,215,200}
\definecolor{attn_code1}{RGB}{238,169,139}
\definecolor{mlp_code0}{RGB}{204,201,221}
\definecolor{mlp_code1}{RGB}{102,95,153}

\definecolor{token_blue}{RGB}{84, 120, 140}

\usepackage{amsmath,amsfonts,bm}

\def\eqref#1{equation~\ref{#1}}

\def\1{\bm{1}}

\DeclareMathAlphabet{\mathsfit}{\encodingdefault}{\sfdefault}{m}{sl}
\SetMathAlphabet{\mathsfit}{bold}{\encodingdefault}{\sfdefault}{bx}{n}

\usepackage{amsmath,amsfonts,bm}

\def\eqref#1{equation~\ref{#1}}

\def\1{\bm{1}}

\DeclareMathAlphabet{\mathsfit}{\encodingdefault}{\sfdefault}{m}{sl}
\SetMathAlphabet{\mathsfit}{bold}{\encodingdefault}{\sfdefault}{bx}{n}

\usepackage{multirow}
\usepackage{diagbox}
\usepackage{makecell}
\usepackage{tabularx}
\usepackage{graphicx}
\usepackage{wrapfig}
\usepackage{array}
\usepackage{rotating}

\definecolor{aliceblue}{rgb}{0.94, 0.97, 1.0}
\definecolor{citecolor}{HTML}{0071BC}
\definecolor{linkcolor}{HTML}{ED1C24}
\definecolor{darkgreen}{HTML}{539165}

\makeatletter
\newcommand{\thickhline}{%
 \noalign {\ifnum 0=`}\fi \hrule height 1pt
 \futurelet \reserved@a \@xhline
}
\makeatother

\newcommand{\tablesize}{
  \fontsize{8.6pt}{10pt}\selectfont
}

\usepackage{pifont}

\usepackage{pifont}       
\usepackage{bbding}       
\usepackage{fontawesome}
\usepackage{xspace}

\usepackage{float}
\usepackage{enumitem}

\newlength\savewidth

\newcolumntype{x}[1]{>{\centering\arraybackslash}p{#1pt}}
\newcolumntype{y}[1]{>{\raggedright\arraybackslash}p{#1pt}}
\newcolumntype{z}[1]{>{\raggedleft\arraybackslash}p{#1pt}}

\renewcommand{\paragraph}[1]{\vspace{1.25mm}\noindent\textbf{\color{headingblue}#1}}

\usepackage{colortbl}
\usepackage{xcolor}
\usepackage{wrapfig}

\newcommand{\modelname}{UI-Mate\xspace}

\usepackage{algorithm}
\usepackage{listings}

\definecolor{codeblue}{rgb}{0.25, 0.5, 0.5}
\definecolor{codekw}{rgb}{0.35, 0.35, 0.75}
\lstdefinestyle{Pytorch}{
    language = Python,
    backgroundcolor = \color{white},
    basicstyle = \fontsize{9pt}{8pt}\selectfont\ttfamily\bfseries,
    columns = fullflexible,
    aboveskip=1pt,
    belowskip=1pt,
    breaklines = true,
    captionpos = b,
    commentstyle = \color{codeblue},
    keywordstyle = \color{codekw},
}

\definecolor{colSubject}{HTML}{D32F2F}   
\definecolor{colAction}{HTML}{F57C00}    
\definecolor{colDetail}{HTML}{388E3C}    
\definecolor{colSpatial}{HTML}{1976D2}   
\definecolor{colMood}{HTML}{7B1FA2}      
\definecolor{colKnow}{HTML}{AFB42B}      

\usepackage{xcolor}

\usepackage{multicol}
\usepackage{enumitem}

\usepackage{tikz}
\usetikzlibrary{tikzmark, calc, shadows.blur, shapes.geometric, fit, positioning}
\usepackage{xparse}
\pgfdeclarelayer{background}
\pgfdeclarelayer{floatbox}
\pgfdeclarelayer{foreground}
\pgfsetlayers{background,floatbox,main,foreground}

\newcounter{hcellcount}

\NewDocumentCommand{\hctext}{m}{\csname hctext@#1\endcsname}
\NewDocumentCommand{\sethctext}{mm}{\expandafter\gdef\csname hctext@#1\endcsname{#2}}

\definecolor{scoreRed}{RGB}{200, 0, 0}
\definecolor{grayText}{RGB}{120, 120, 120}

\definecolor{green}{HTML}{009000}
\definecolor{red}{HTML}{ea4335}

\definecolor{cvblue}{rgb}{0.15, 0.45, 0.68}

\title{UI-Mate: Advancing Open-Weight Foundation GUI Agents with In-Context Demonstrations}

\affiliation[]{Tencent Hy Frontier Team \\[0.4em]}

\abstract{
Foundation GUI agents hold immense potential for automating complex digital tasks, yet their deployment is hindered by two critical challenges: training-level data scarcity and distributional bias, alongside interaction-level prompt ambiguity and execution unreliability. Routine workflows rely heavily on user-specific tools and tacit conventions, leaving unstated instructions open to arbitrary variations across runs—so an agent that succeeds once may fail on the next attempt. We present \textbf{UI-Mate}, a foundation GUI agent designed to overcome these bottlenecks by integrating an environment-grounded training stack with in-context demonstration learning. UI-Mate incorporates three core contributions: \textbf{A Scalable Environment-Grounded Training Stack:} A closed-loop data engine that automates task generation, environment construction, rollout, filtering, and hierarchical capability balancing, feeding supervised fine-tuning (SFT) and online reinforcement learning (RL) across massively parallel environments via unified task–verifier bundles. \textbf{In-Context Demonstration Learning:} A mechanism that transforms multimodal demonstrations into flexible, subtask-level workflows rather than replaying rigid trajectories, adhering to demonstrated steps where they matter while autonomously re-planning from the live interface. \textbf{OSWorkerBench Benchmark \& Insights:} A benchmark of 100 long-horizon office tasks across 41 applications that supports instruction-only and demonstration-guided evaluation.
Its demonstration resources separate a 33-task self-demo setting, built from successful strong-agent rollouts of the same targets, from a 45-task variant-demo setting, built from human recordings of related but non-identical tasks. Experiments show that UI-Mate-27B sets a new open-weight state of the art on general computer-use benchmarks, scoring 77.0\% on OSWorld-Verified and 66.2\% on WindowsAgentArena. On OSWorkerBench, it reaches 41.0\% strict success and 76.9\% progress, outperforming its Qwen3.6-27B base by 17.7 and 24.5 points. On the 33-task self-demo subset, one demonstration raises strict success from 17.2\% to 35.4\% and progress from 67.9\% to 81.1\%, substantially improving long-horizon reliability. Project page:~\url{https://ui-mate.github.io}.
}
\newcommand{\teaserfigure}{%
    \vspace{-0.1in}
    \begingroup
    \noindent\begin{minipage}{\linewidth}
    \centering
\includegraphics[width=1.0\linewidth]{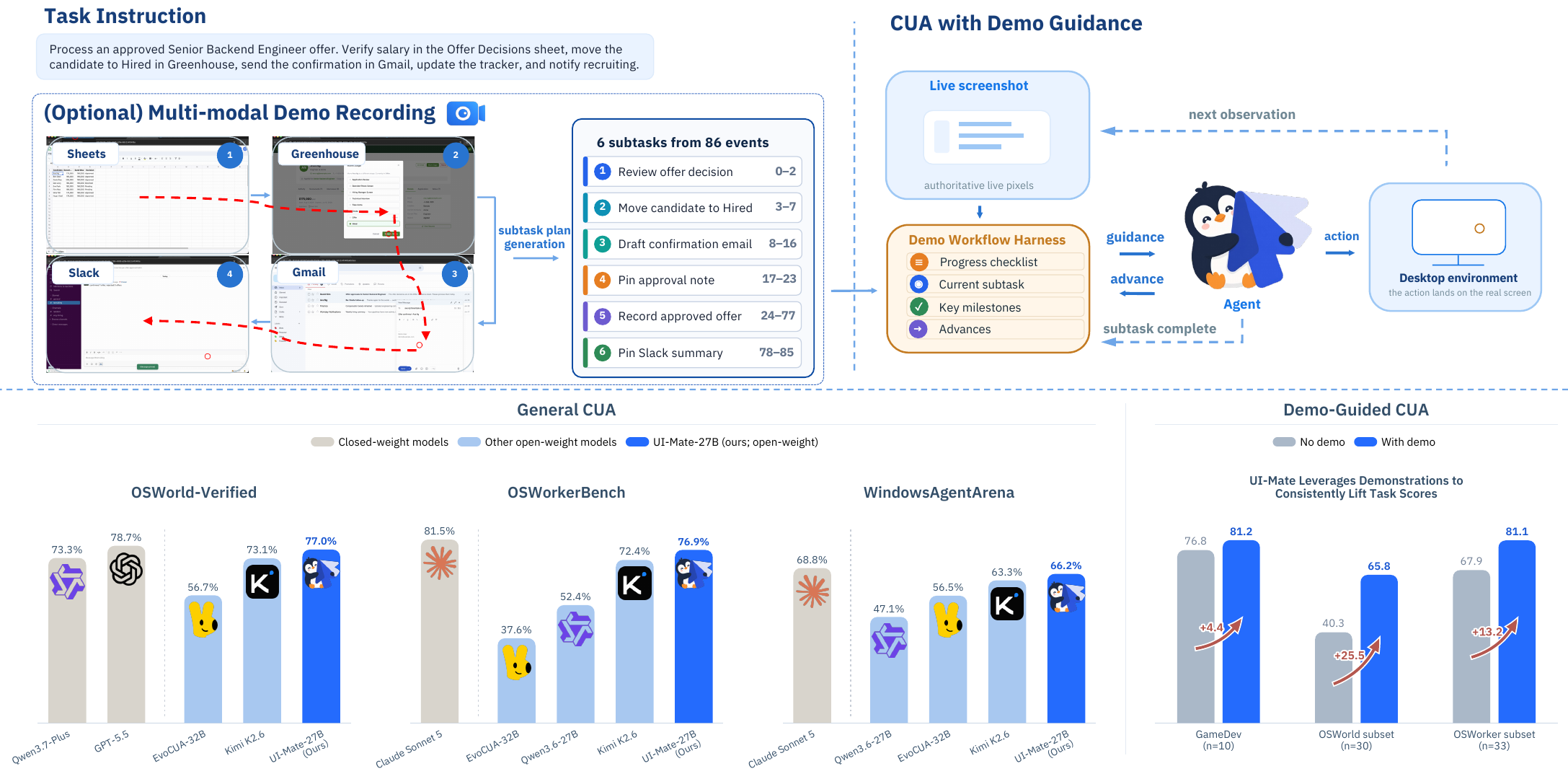}\par
    \captionsetup{font=small}
    \captionof{figure}{\textbf{UI-Mate combines strong general computer-use capabilities with demonstration-guided execution.}
    \textit{Top:} For an underspecified cross-application task, an optional multimodal demonstration is distilled into a subtask-level workflow. During execution, the harness provides the current subtask and key milestones; the agent grounds them in the live interface, acts, and advances the workflow upon completion.
    \textit{Bottom:} In the general CUA evaluation setting, UI-Mate-27B is competitive with leading open- and closed-weight systems across OSWorld-Verified, OSWorkerBench, and WindowsAgentArena. In the reported self-demo evaluation, one same-task demonstration raises task scores from 76.8 to 81.2 on GameDev, 40.3 to 65.8 on the OSWorld subset, and 67.9 to 81.1 on the 33-task OSWorkerBench subset.}%
    \label{fig:teaser}%
    \end{minipage}
    \endgroup
}

\date{\today}

\begin{document}
\thispagestyle{firstheader}
\maketitle 
\pagestyle{empty}

\clearpage
\begingroup
\setcounter{tocdepth}{3}
\hypersetup{linkcolor=headingblue}
\tableofcontents
\hypersetup{linkcolor=metabg}
\endgroup
\clearpage

\section{Introduction}

Foundation GUI agents promise to turn natural-language intent into multi-step action across software ecosystems.
Recent CUA systems~\cite{wang2026opencua,liu2026scalecua}, including UI-TARS~\cite{qin2025ui,ui-tars2} and Qwen-UI-Agent~\cite{zhou2026qwen_ui_agent}, show that vision-language models can perceive interfaces, ground actions, and execute extended workflows directly on screen.
Yet benchmark progress alone does not yield reliable real-world deployment, which remains constrained by two complementary bottlenecks.
The first is a \emph{training bottleneck}: scalable GUI learning requires not only trajectories, but also executable environments, reliable verifiers, and broad capability coverage.
The second is an \emph{interaction bottleneck}: an instruction usually specifies the desired outcome more readily than the user-specific procedure by which it should be achieved.
The former limits what an agent can learn, whereas the latter limits how consistently it applies what it already knows.

GUI interaction data are inseparable from the environments in which they are generated.
Unlike static text or image corpora, a trajectory is useful for learning only when its initial state can be instantiated, its actions can be executed, and its outcome can be verified.
Recent work has responded with scalable synthetic experience, cross-platform trajectory collection, and automated construction of verifiable environments~\cite{xue2026evocua,liu2026scalecua,zhang2026infiniteweb,wang2026cua}.
Scale alone, however, does not guarantee coverage.
Data production naturally favors what is inexpensive to instantiate: short, single-application tasks accumulate, whereas long-horizon workflows, cross-application information transfer, and recovery from execution errors remain sparse.
Training on this skewed distribution encourages narrow execution patterns.
The central data problem is therefore not merely how many trajectories to collect, but how to diagnose and correct what the corpus fails to cover.

The interaction bottleneck persists even for a capable agent because the missing information is absent from the instruction itself.
Routine work is shaped by a user's tools, file organization, templates, naming conventions, and required output formats rather than by a single universal procedure.
Encoding every such choice in a prompt can require as much effort as performing the task manually, so users provide concise instructions and leave procedural details implicit.
Those details may then be resolved differently across runs, causing an agent that succeeds once to fail on an ostensibly identical request.
Mean success rates obscure this distinction: occasional correct resolutions of ambiguity and consistently correct resolutions can produce the same average, although only the latter supports dependable delegation.
For end users, reliability is not secondary to capability but the means through which capability is experienced.

We address both bottlenecks with \textbf{UI-Mate}, an open-weight foundation GUI agent that combines environment-grounded training with in-context demonstration learning.
Figure~\ref{fig:teaser} summarizes the resulting system and its two evaluation regimes, while \S\ref{sec:overview} formalizes the task and presents the complete architecture.
On the training side, a closed-loop data pipeline (\S\ref{sec:data}) constructs tasks and executable environments, collects and filters rollouts, and uses a hierarchical capability tree to identify and rebalance gaps in coverage.
The resulting verified trajectories and task--verifier bundles support supervised fine-tuning followed by online agentic reinforcement learning in the general computer-use training stack (\S\ref{sec:training}).

On the interaction side, UI-Mate introduces DemoCUA (\S\ref{sec:democua}) to communicate procedural intent through a multimodal demonstration recorded from either a human or an agent.
Prior human-taught GUI agents have shown that screen recordings can expose reusable procedural knowledge~\cite{showui_aloha}.
UI-Mate converts such a recording into a subtask-level workflow rather than treating it as an action sequence to replay.
At execution time, the live screenshot remains authoritative: the agent follows demonstrated steps when they remain relevant, supplies omitted low-level actions, and re-plans when the target task diverges from the recording.
UI-Mate App (\S\ref{sec:uimate}) realizes this paradigm on the user's own desktop by recording a demonstration and subsequently running the agent against the same live environment.

Measuring the value of this guidance requires more than the instruction-only protocols used by established desktop benchmarks such as OSWorld~\cite{xie2024osworld} and WindowsAgentArena~\cite{bonatti2025windowsagentarena}.
We therefore introduce OSWorkerBench (\S\ref{sec:OSWorkerBench}), an office-centric benchmark of 100 long-horizon tasks spanning 41 normalized applications and 10 job families.
Its 67 Long-Memory tasks require delayed reuse of dynamic information, while 49 Multi-App tasks require substantive information transfer across at least three applications.
It supports instruction-only and demonstration-guided evaluation, with two distinct demonstration settings.
The 33-task \emph{self-demo} set pairs each target with a successful strong-agent rollout of that same task and supports the quantitative DemoCUA evaluation in this report.
The 45-task \emph{variant-demo} set instead pairs each target with a multimodal human recording of a semantically related but non-identical source task and is intended to measure procedural transfer.
In either paired comparison, the target instruction, initialized environment, interaction budget, and executable verifier remain fixed; only demonstration availability changes.

Our evaluation (\S\ref{sec:evaluations}) shows that UI-Mate-27B reaches 77.0\% on OSWorld-Verified and 66.2\% on Windows Agent Arena, establishing the strongest open-weight results among the systems in our comparison.
On OSWorkerBench, it achieves 41.0\% strict success and 76.9\% progress, improving over its Qwen3.6-27B base model by 17.7 and 24.5 percentage points, respectively.
On a 30-task subset of OSWorld-Verified, one demonstration raises the average task score from 40.3\% to 65.8\%.
We observe similarly substantial gains on the 33-task OSWorkerBench self-demo subset, introduced in this work, where progress increases from 67.9\% to 81.1\% and strict success from 17.2\% to 35.4\%.
These results support a central empirical finding: demonstrations improve not only average task completion, but also the consistency with which an agent realizes underspecified user intent.

In summary, this report makes three primary contributions:
\begin{itemize}
    \item \textbf{Environment-grounded data and training.} We develop a closed-loop pipeline that jointly constructs tasks and environments, filters trajectories, diagnoses capability coverage, and produces supervision for both SFT and online RL.
    \item \textbf{Demonstration-guided computer use.} We introduce an in-context learning formulation that converts multimodal human or agent demonstrations into adaptive subtask-level workflows, preserving procedural intent without reducing execution to rigid replay.
    \item \textbf{Benchmark and empirical evidence.} We introduce OSWorkerBench and its controlled paired protocol, and show that UI-Mate advances open-weight general computer use while demonstrations substantially improve long-horizon execution reliability.
\end{itemize}

\section{Overview}
\label{sec:overview}

\subsection{Task Formulation}
\label{sec:formulation}

\paragraph{General computer use.}
We define a computer-use task as a pair
\begin{equation}
\label{task_fomulation}
\mathcal{T} = (x, \mathcal{E}),
\end{equation}
where $x$ is a natural-language instruction and $\mathcal{E}$ is a computer-use environment: a machine with an operating system, installed applications, and user files, which the agent operates directly.

The agent interacts with $\mathcal{E}$ in discrete decision steps. At step $t$ it receives an observation $o_t$, the screenshot of the screen after the previous action has been executed. It then produces one response
\begin{equation}
y_t = (r_t, a_t) \sim \pi_{\theta}\!\left(\,\cdot \mid x,\, h_t,\, o_t\right),
\end{equation}
where $r_t$ is intermediate reasoning, $a_t$ is the action to execute, and $h_t$ is the interaction history. We keep $h_t$ to a bounded window of recent responses, so the input does not grow with the length of the task. We write $c_t=(x,h_t,o_t)$ for this step context.

A single response may carry more than one action,
\begin{equation}
a_t = \left(a_t^{(1)},\ldots,a_t^{(K_t)}\right),
\qquad
a_t^{(k)} \in \mathcal{A},
\end{equation}
where $\mathcal{A}$ is the action space. The $K_t$ actions run consecutively, and the agent receives no new observation until they finish. One response is therefore one \emph{decision turn}, and we measure interaction length in decision turns rather than in individual operations.

Execution ends when the agent emits a terminal action or exhausts its step budget. The resulting trajectory is
\begin{equation}
\tau = (x,\, o_1,\, y_1,\, \ldots,\, o_T,\, y_T).
\end{equation}
Whether the task succeeded cannot be read off $\tau$ alone. It is decided by an executable verifier that inspects the final state of $\mathcal{E}$ and returns a binary outcome $R(\tau)\in\{0,1\}$. The agent optimizes $\mathbb{E}\left[R(\tau)\right]$.

\paragraph{Demonstration-guided computer use.}
A demonstration-guided task additionally provides a demonstration $d$ of how the task, or a closely related one, has been carried out before.
We do not consume $d$ as a flat action sequence; it is segmented into an ordered set of subtasks
\begin{equation}
d = (s_1,\ldots,s_N),
\qquad
s_n = \bigl(\ell_n,\, v_n,\, u_n\bigr),
\end{equation}
where $\ell_n$ is a natural-language subtask goal, $v_n$ is a verifiable completion criterion
for it, and $u_n=\bigl(u_n^{(1)},\ldots,u_n^{(M_n)}\bigr)$ is the ordered sequence of action descriptions performed inside that subtask. Every $u_n^{(m)}$ is a textual description of an
operation together with the visual cues that identify its target; it carries no pixel coordinates and is therefore advisory rather than executable.
At each step the agent is shown only the part of $d$ that is currently relevant. A pointer $n_t\in\{1,\ldots,N\}$ tracks the active subtask, and the agent conditions on the \emph{workflow view}
\begin{equation}
g_t = \Phi(d, n_t) = \bigl(\underbrace{\ell_1,\ldots,\ell_N}_{\text{progress checklist}},\;
\ell_{n_t},\, v_{n_t},\, u_{n_t}\bigr),
\end{equation}
i.e.\ the goals of all subtasks annotated with completed / current / upcoming markers, but the detailed action steps of the active subtask only. 
The policy becomes
\begin{equation}
y_t = (r_t, a_t) \sim \pi_{\theta}\!\left(\,\cdot \mid x,\, h_t,\, o_t,\, g_t\right),
\end{equation}
and the action space is extended with one control action,
$\mathcal{A}^{+} = \mathcal{A}\cup\{\textsc{subtask\_complete}\}$. 
The agent emits \textsc{subtask\_complete} once $o_t$ satisfies $v_{n_t}$, which advances the pointer,
\begin{equation}
n_{t+1} =
\begin{cases}
\min(n_t+1,\,N), & a_t = \textsc{subtask\_complete},\\
n_t, & \text{otherwise},
\end{cases}
\end{equation}
so progress through the demonstration is driven by the observed screen state rather than by a fixed step counter.
Two properties of this formulation matter for what follows. First, $g_t$ is a \emph{prior}, not a target: $u_{n_t}$ may omit the low-level operations needed on the live screen, may reference elements that are absent, or may be misaligned with the current state, so the
correct $a_t$ is the one the observation $o_t$ demands and the observation retains veto power over the demonstration. Second, the objective is unchanged---the verifier still inspects the
final state of $\mathcal{E}$ and the agent still maximizes $\mathbb{E}[R(\tau)]$---because the demonstration alters only the conditioning of the policy. General computer use is thus the special case $g_t = \varnothing$, which occurs whenever no useful demonstration is available,
and a demonstration-guided agent must remain competent in that regime.

\subsection{UI-Mate System Overview}
\label{sec:system_overview}
UI-Mate delivers five connected components that together support the full
lifecycle of a computer-use agent: an environment-grounded data pipeline,
a training stack for general computer use, DemoCUA for learning from
in-context demonstrations, OSWorkerBench for controlled evaluation, and a
unified harness for deployment and evaluation.
Together, they connect executable data construction, policy training,
procedural adaptation, benchmark evaluation, and real-world execution.

\paragraph{Environment-grounded data pipeline.}
UI-Mate uses an environment-grounded data pipeline to produce diverse and
executable training data.
Starting from task instructions, the pipeline constructs the required
applications, files, and initial states, executes agent rollouts, and removes invalid or incomplete trajectories.
A hierarchical capability tree tracks data coverage and identifies
applications, operations, and workflows that remain underrepresented.
The pipeline produces verified trajectories for supervised fine-tuning and
executable task--verifier bundles for online reinforcement learning.
Rollout outcomes and capability diagnostics are fed back into task and
environment construction, allowing later collection to focus on weak or
missing capabilities.
In \S\ref{sec:data} we describe the pipeline in detail.

\paragraph{Training for general computer use.}
Built on the outputs of the data pipeline, the general computer-use training
stack combines supervised fine-tuning with agentic reinforcement learning.
Supervised fine-tuning teaches the interaction protocol, visual grounding,
application workflows, and multi-step execution, while reinforcement learning allows the policy to act in executable environments and learn from verifier scores assigned to complete trajectories.
Together, these stages produce a policy that can plan over long horizons,
track cross-application state, recover from errors, and complete tasks from
natural-language instructions.
In \S\ref{sec:training} we present the training method.

\paragraph{DemoCUA.}
DemoCUA enables the general policy to learn user-specific procedures from in-context multimodal demonstrations.
A recorded human execution or successful agent rollout is converted offline into a structured workflow containing subtasks, completion criteria, and visually grounded action descriptions.
During execution, the model receives the current workflow together with the
live screenshot and interaction history.
Because the screenshot remains the primary source for action selection, the
model can follow useful demonstrated steps, skip unnecessary ones, and revise the plan when the target task differs from the recording.
Training examples cover cases in which the demonstration agrees with, diverges from, or is unrelated to the current interface state, teaching the model to use the workflow as guidance rather than as a fixed script.

\paragraph{OSWorkerBench.}
OSWorkerBench evaluates both general computer-use ability and the ability to use a procedure supplied by a demonstration.
It contains 100 long-horizon office tasks across 41 applications, all available for instruction-only evaluation.
Its demonstration-guided mode has two distinct settings: 33 targets have self-demos obtained from successful strong-agent rollouts on those same tasks, while 45 targets have human-recorded variant-demos from related but non-identical tasks.
Within either paired protocol, the target is evaluated with and without its demonstration under the same instruction, environment, interaction budget, and executable verifier.
The performance difference therefore isolates the value of demonstration availability from the model's instruction-only ability, while only the variant-demo setting measures transfer across task variants.
The benchmark reports both strict task success and partial progress on long
workflows.
In \S\ref{sec:OSWorkerBench} we describe its design and construction.

\paragraph{Unified harness for deployment and evaluation.}
A shared harness connects the trained policy to executable computer
environments during both deployment and evaluation.
It manages the observation--reasoning--action loop, interaction history, model invocation, action execution, and trajectory recording, while platform-specific adapters support local desktops, virtual machines, and sandbox environments.
UI-Mate App uses this harness for general and demonstration-guided execution, demonstration recording, and user oversight of the execution trace.
The evaluation system uses the same interaction format in controlled environments and attaches executable verifiers to completed trajectories.
In \S\ref{sec:uimate} we describe UI-Mate App and its deployment.

The remainder of this report presents these components in turn, covering the
environment-grounded data pipeline, general computer-use training, DemoCUA,
OSWorkerBench, and the shared harness for evaluation and deployment.

\section{Data Pipeline}
\label{sec:data}

\begin{figure}[htp]
  \centering
  \includegraphics[width=0.95\linewidth]{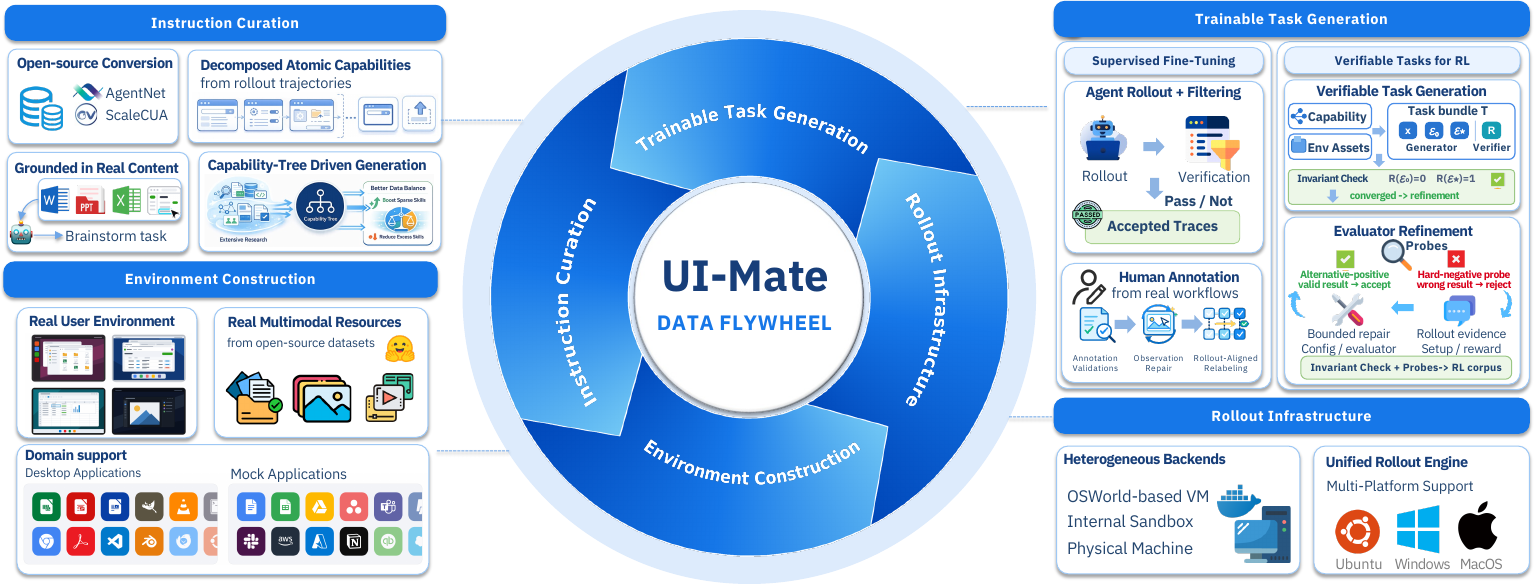}
  \caption{\textbf{Overview of the UI-Mate data flywheel.} 
  The pipeline jointly scales instruction curation, environment construction, trainable task generation, and rollout infrastructure. 
  SFT data combines filtered agent rollouts with validated and repaired human trajectories; RL data is built from verifiable task bundles whose evaluators are refined using complementary probes and rollout feedback. Coverage and outcome diagnostics are fed back to rebalance capabilities and continuously improve the diversity, difficulty, and reliability of the training distribution.
  }
  \label{fig:dataflywheel}
\end{figure}

\subsection{Task Instructions}
\label{sec:data_instructions}

Task instructions for rollout and training general computer-use agent must satisfy two requirements that pull in different
directions: they should reflect what people actually do on a computer, and
they should systematically cover the functional surface an agent is expected
to operate. We meet both with four complementary sources. Open-source
computer-use datasets, including AgentNet~\cite{wang2026opencua} and ScaleCUA~\cite{liu2025scalecua}, provide a broad foundation of everyday tasks
derived from real user activity. Atomic subtasks decomposed from failed or
stalled rollouts concentrate the distribution on
operations that agents demonstrably find difficult. Instructions generated
from real documents, spreadsheets, presentations, and static
websites~\cite{zhang2026infiniteweb} reference concrete entities and
relationships rather than synthetic placeholders, and support long-horizon
workflows that span multiple applications. Finally, capability trees built
from application specifications surface fine-grained operations that common
workflows would otherwise leave untouched. 
We deliberately bias the distribution toward everyday office use by incorporating both tasks performed by real users and tasks derived from authentic working materials. This combination promotes diversity and realism while ensuring systematic coverage of the capabilities required for GUI agents. The source of the task instructions is detailed in Appendix~\ref{appendix:task_instructions}.

\subsection{Environment Construction}
\label{sec:data_env}

\paragraph{From Instruction to Runnable Environment}
We design an automated pipeline that converts each instruction into an executable environment. An LLM identifies required files, such as documents, spreadsheets, presentations, or images, and generates executable code to create them when needed. The pipeline uploads the resources and configures the operating system, applications, and task-specific state. This produces the initial setup and environment resources needed for execution.
Real user environments vary substantially even when they support the same task. Randomized setup code therefore varies wallpapers, desktop layouts, application settings, and sidebar positions while preserving task feasibility. This diversifies the data and reduces dependence on incidental visual or interface configurations.

\paragraph{Grounding Environments with Real Resources}
LLM-generated resources are often shorter and more homogeneous than real files and may contain placeholders. These artifacts create shortcuts that distort realism and difficulty. We therefore index open-source documents, presentations, spreadsheets, images, videos, and audio. The indexed files better reflect the richness and irregularity of real working materials. During construction, the LLM retrieves and copies a real file through setup code, using synthesis when none exists. Tasks already grounded in real files or static websites use those resources directly. This approach simplifies access to realistic materials while reducing synthetic artifacts.

\subsection{Rollout and Filtering}
\subsubsection{Rollout  Infrastructure}
\label{sec:data_rollout}
Our rollout infrastructure is designed to execute large numbers of heterogeneous computer-use tasks efficiently and reliably. It supports Ubuntu, Windows, and macOS environments through a unified rollout interface, while accommodating the different virtualization and deployment constraints of each operating system.
To support parallel rollouts and complex reinforcement-learning environments, we develop an internal cloud virtual machine backend optimized for lightweight environment provisioning and high-throughput parallel execution. It exposes programmable interfaces for environment creation, interaction, observation, and teardown, allowing workers to be allocated dynamically while maintaining isolation between concurrent rollouts.
It also manages the complete rollout lifecycle, including resource upload, environment setup, agent interaction, observation collection, and reward computation. To sustain high-throughput data collection and reinforcement learning, rollouts are scheduled independently onto available workers, allowing large numbers of long-horizon trajectories to proceed concurrently across heterogeneous machines and backend types.

\subsubsection{Trajectory Filtering}
\label{sec:data_filtering}

The filtering of the rollout trajectories undergoes two stages, including task and environment validity checks and step-level outcome filtering. For the first stage, a multimodal judge will check the setup configuration as well as the initial environment state, and retain a trajectory only when its task and environment are valid, and rollout evidence supports every expected deliverable.
Before assessing agent behavior, the judge rejects ambiguous or infeasible tasks, tasks inconsistent with the setup or already satisfied in the initial state, and trajectories with malformed actions, missing visual observations, or environment failures. This prevents task and infrastructure defects from entering supervision.
For the second step-level outcome verification stage, the judge extracts independently verifiable deliverables from each instruction and tracks supporting evidence across GUI observations and actions. It retains a trajectory only when every deliverable is supported. Tracking evidence throughout the rollout prevents plausible final states or unsupported agent narration from masking partial failure and captures intermediate evidence absent from the final frame.

\subsection{Capability Tree and Data Diagnosis}
\label{sec:data_capability}
\begin{wrapfigure}{r}{0.47\textwidth}
    \centering
    \vspace{-15mm}
    \includegraphics[width=\linewidth]{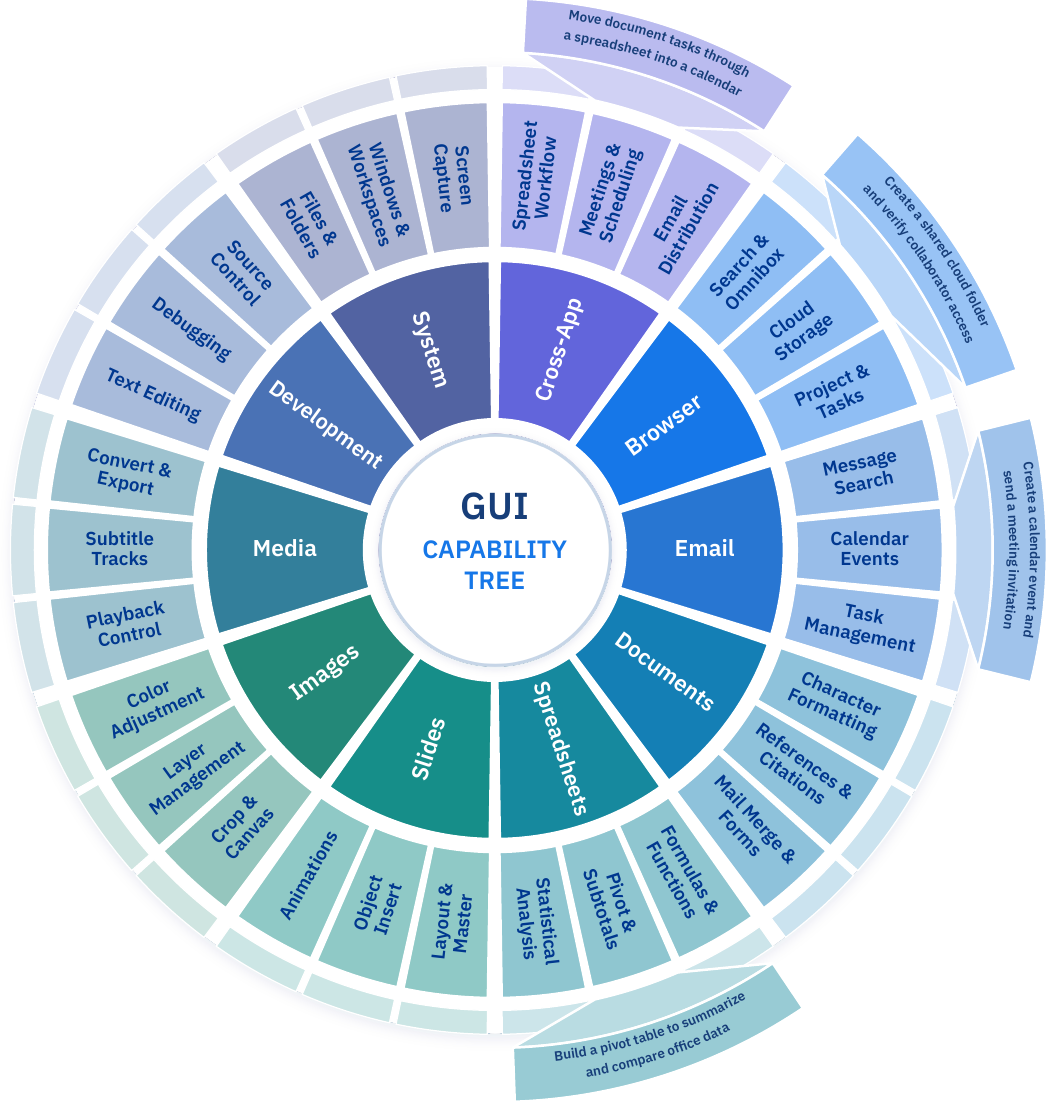}
    \caption{Representative subset of the collected GUI capability tree. The hierarchy proceeds from application domains to fine-grained capabilities and representative workflows; sector sizes are schematic.}
    \label{fig:gui_capability_tree}
    \vspace{-16mm}
\end{wrapfigure}

Large-scale data production favors inexpensive tasks, while application-level statistics obscure underrepresented behaviors. We therefore map each task to a shared capability taxonomy and associate its rollout outcomes with the corresponding capabilities, making coverage gaps actionable for subsequent data generation.

\subsubsection{Extracting Capabilities}

For each application, we maintain an inventory of atomic user-facing capabilities. We initialize it from application documentation and expand it with behaviors found in training instructions.
During annotation, we match each instruction to an existing entry and create a new one only when no suitable match exists.
We maintain a separate cross-application domain for behaviors that connect applications, such as transferring and transforming information between them.

\subsubsection{Constructing the Capability Tree}

As a flat inventory grows, matching becomes ambiguous and data budgets become difficult to allocate consistently. We therefore organize capabilities into three levels: application, coarse capability, and fine-grained operation. Each task is first routed to a coarse capability and then matched within the corresponding subtree, reducing confusion among near-duplicates. We keep the coarse level stable for budgeting, add fine-grained nodes as needed, and periodically consolidate redundant or inactive nodes. Equivalent entries are aligned across applications to expose shared interaction patterns.

\subsubsection{Data Rebalancing}

We rebalance the corpus using target coverage, observed data density, rollout success, and filtering rejection rates. Low density relative to the target triggers additional generation, whereas oversupplied capabilities are down-weighted. Sufficient volume combined with low rollout success or high rejection instead signals task or environment defects. We treat task length as a separate sampling dimension to prevent abundant short tasks from masking long-horizon gaps. Newly generated tasks pass through the same annotation and evaluation pipeline, and their outcomes update the next allocation cycle. This closes the loop between capability diagnosis and data generation.

\subsection{Human Annotation}
\label{sec:data_annotation}
Automated rollouts provide scale, but remain biased toward tasks that current agents can execute and environments that are inexpensive to instantiate. We therefore complement them with human-annotated trajectories from real workflows across operating systems and applications, capturing natural interaction patterns and long-tail behavior underrepresented in synthetic data. Because raw annotations may contain inconsistent labels, recording artifacts, and supervision that differs from the rollout format, we subject them to validation, observation repair, and aligned relabeling.

\paragraph{Annotation Validation and Observation Repair}
Deterministic checks verify structural integrity and action--parameter agreement, while multimodal review assesses task completion and whether each action matches the visible state transition. We pay particular attention to observation timing: a pre-action screenshot may leak the target through cursor or hover state, or capture an unstable interface transition. Affected steps are repaired by selecting an earlier buffered frame for target leakage or a later one for incomplete rendering, with time-bounded recovery from the source recording when needed. Repaired frames are revalidated; only trajectories with irreparable corruption or annotation-tool contamination are removed. This repair-first policy preserves costly supervision without admitting visual shortcuts or inconsistent actions.

\paragraph{Rollout-Aligned Relabeling}
Human recordings provide action types and arguments but not the reasoning and natural-language action descriptions expected by the policy. A teacher model adds them using the same system prompt, interaction history, observation folding, and action space as model rollout. The outputs are generated jointly and sequentially, while all human-annotated action parameters remain fixed. Consistency checks reject action mismatches and privileged annotation leakage, yielding the same decision-step representation without altering the executable human supervision.

\subsection{Verifiable Tasks for Reinforcement Learning}
\label{sec:data_verifiable}

Reinforcement learning from verifiable rewards (RLVR) requires more than instructions and runnable environments: every rollout must terminate in a verifiable signal. Authoring such tasks manually is expensive, because the instruction, the configuration, and the reward must agree on the same initial conditions and completion criteria. We therefore automate construction in two stages: \emph{generation} turns capability and environment assets into diverse, self-contained task bundles, and \emph{refinement} independently stress-tests and repairs the evaluators. 
Both stages operate on a common data contract. Beyond the instruction $x$ and environment $\mathcal{E}$ of
Equation~\ref{task_fomulation}, a task usable for RL fixes the initial state $\mathcal{E}_0$ from which rollouts begin,
a reference completion state $\mathcal{E}^\star$ the generator can reach, and a per-task executable verifier
$R$ evaluated on environment state. Every such task must satisfy the execution invariant
\begin{equation}
R(\mathcal{E}_0) = 0,
\qquad
R(\mathcal{E}^\star) = 1 .
\label{eq:invariant}
\end{equation}
Generation uses this invariant for internal consistency; refinement asks whether the evaluator faithfully
captures the instruction rather than merely agreeing with the reference.

\subsubsection{Verifiable Task Generation}

\paragraph{Scaling instructions and executable configurations.}
We reuse the instruction sources and environment-construction machinery described in \S\ref{sec:data_instructions} and \S\ref{sec:data_env}, but couple them more tightly for RL. Capability-guided sampling controls which behaviors are exercised and how they are composed, while configuration synthesis instantiates the resources, preconditions, and application state required to execute each instruction. The same capability can therefore be realized under different files, user contexts, application states, and system configurations, increasing interaction diversity rather than merely producing paraphrases. Conversely, complex instructions retain the environmental dependencies needed to make them genuinely executable instead of being simplified to what is easiest to initialize or evaluate. This joint expansion of instruction and configuration space is the primary mechanism by which we scale the coverage and difficulty of verifiable tasks.

\paragraph{Decoupled construction of environments and rewards.}
Inspired by the generator--discriminator formulation of CUA-Gym~\cite{wang2026cua}, we separate the roles of realizing a task and judging its outcome. A generator constructs the executable configuration together with the initial and reference completion states, whereas a verifier agent derives the reward from the task specification. This information boundary keeps evaluation faithful---seeing only the specification and the resulting state, the verifier judges whether the user-visible objective is met, not how the reference completion arose. We further extend this separation with application-aware configuration synthesis: the state that determines task success may live in different places depending on the application — user artifacts, application profiles, structured stores (e.g., databases), or operating-system state — and the generator resolves this per task while exposing all such state to the verifier through a single, common evaluator interface. Co-designing configuration observability and reward semantics makes verifiability a construction-time constraint, rather than an annotation added after rollout collection.

\paragraph{Generation-time convergence checks.}
A bundle is retained only once its components converge on the same task semantics. Static and model-based checks reject unsupported preconditions, inconsistent artifacts, weak proxy rewards, and configurations that do not expose the state required for evaluation; execution then tests the invariant directly. This removes malformed or trivially satisfied tasks cheaply, but establishes only internal consistency, since the reference completion and the reward can share a semantic blind spot---making convergence a prerequisite for, not a substitute for, independent refinement.

\subsubsection{Evaluator Refinement}

\paragraph{Independent probes beyond self-consistency.}
Refinement takes converged bundles as read-only inputs and re-examines instruction--reward alignment along a reasoning path independent of generation. We first make evaluator decisions observable by decomposing a
scalar pass/fail result into per-requirement diagnostics, then apply two complementary probes. A \emph{hard-negative} probe perturbs a successful state into a plausible but incorrect completion and should be rejected; acceptance indicates an under-specified reward. An \emph{alternative-positive} probe constructs a different legitimate completion and should be accepted; rejection indicates an over-strict reward. With the
original initial and reference states, the probes provide evidence about both false acceptance and false rejection instead of validating a single solution.
 
\paragraph{Rollout feedback and bounded repair.}
Real rollouts expose discrepancies that generation cannot anticipate: the configuration may instantiate an unintended state, a precondition may not survive initialization, a valid outcome may be represented differently by the application, or the reward may respond to a shortcut. Flagged tasks enter a bounded repair loop rather than being discarded; repairs may target the configuration, reference state, evaluator, or---when the objective itself is ambiguous---the instruction. Every modification must pass the invariant and
the probes again, and changes that improve one signal while degrading evaluator observability or task compliance are reverted. Only tasks free of hard setup defects and supported by both positive and negative evidence are promoted to the RL corpus.
 
\paragraph{Scaling as a coupled design objective.}
These components make scaling a coupled objective rather than three independent throughput problems: added generation capacity contributes new capabilities, interaction contexts, and difficulty instead of paraphrases, easy-to-verify tasks, or noisy rewards, broadening the frontier of the training distribution while preserving the executability and reliability RLVR requires.

\section{Training UI-Mate for General Computer Use}
\label{sec:training}

\subsection{Training Recipe}
\label{sec:training_recipe}
\modelname is trained in two stages: supervised fine-tuning (SFT),
followed by agentic reinforcement learning (RL).
SFT learns from filtered offline trajectories
(\S\ref{sec:data}) and equips the model with capabilities for GUI
interaction: producing well-formed and executable responses under the
interaction protocol, grounding semantic targets to screen coordinates,
and selecting actions based on screenshots and interaction history.

Agentic RL improves the model through online interaction in the verifiable
environments (\S\ref{sec:data_verifiable}).
It uses task-level rewards to optimize for successful task completion rather
than action imitation.
This stage improves long-horizon planning, state tracking, error recovery,
and reliable task completion.

\subsection{Supervised Fine-Tuning}
\label{sec:sft}
\paragraph{Training data.}
Our SFT corpus is constructed using the data pipeline described in
\S\ref{sec:data}.
For each task instruction, we instantiate a runnable environment
(\S\ref{sec:data_env}) and execute an agent rollout using the
infrastructure (\S\ref{sec:data_rollout}).
We retain only trajectories for which step-level verification confirms
all required outcomes (\S\ref{sec:data_filtering}).
The retained trajectories are sampled according to the capability tree in
\S\ref{sec:data_capability}, with balanced coverage across
applications, capability levels, and task lengths.
The resulting training mixture covers atomic operations,
single-application workflows, and cross-application tasks, and includes
trajectories both with and without explicit intermediate
reasoning~\citep{wang2026opencua}.

\paragraph{Objective.}
Each training instance is a single decision turn of a retained trajectory. Given the
instruction $x$, the interaction history $h_t$ carried into turn $t$, and the current
screenshot $o_t$, the model reproduces the target response $y_t$, which contains its
reasoning followed by the action to execute. Writing $c_t=(x,h_t,o_t)$ for the turn context,
SFT minimizes the next-token prediction loss over the response tokens,
\begin{equation}
\mathcal{L}_{\mathrm{SFT}}(\theta)
=
-\,\mathbb{E}_{(c_t,y_t)\sim\mathcal{D}_{\mathrm{SFT}}}
\left[\frac{1}{|y_t|}
\sum_{k=1}^{|y_t|}
\log \pi_{\theta}\!\left(y_{t,k}\mid y_{t,<k},\,c_t\right)
\right],
\end{equation}
where $y_{t,k}$ is the $k$-th token of the response and $|y_t|$ its length. The turn context
carries no loss and acts purely as conditioning.

\subsection{Agentic Reinforcement Learning}
\begin{figure*}[h]
    \centering
    \includegraphics[width=\textwidth]{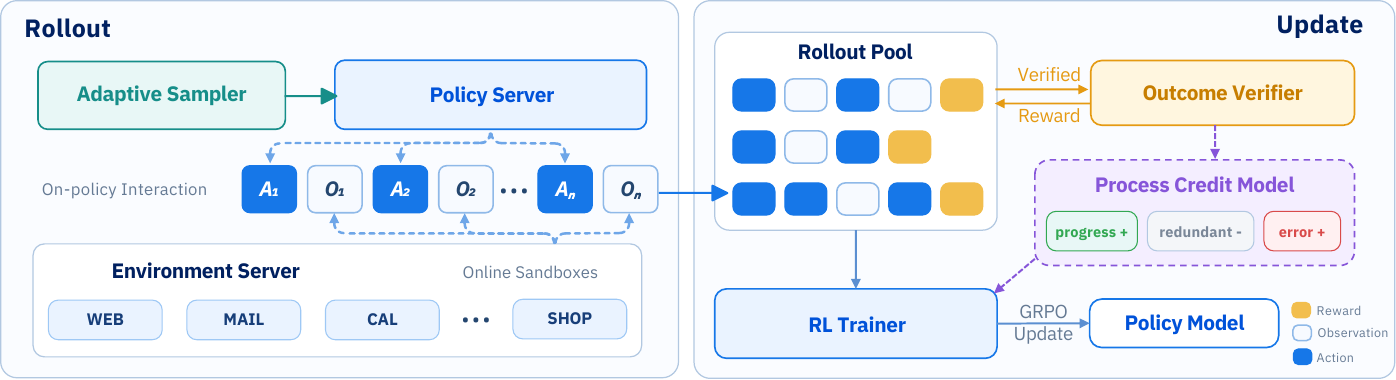}
    \caption{
    \textbf{Agentic RL system of \modelname.} Our RL Pipeline can be divided into rollout stage (left) and update stage (right). Adaptive sampler drives online sandbox rollouts under a fixed policy snapshot; completed trajectories are pooled per task group for outcome verification, optional process credit assignment, and asynchronous GRPO updates.
    }
    \label{fig:training_overview}
\end{figure*}

Starting from an SFT policy, we optimize \modelname online in the verifiable GUI environments described in \S\ref{sec:data_verifiable}, using the decision-turn interaction defined in \S\ref{sec:formulation}. Each completed trajectory is scored by an environment verifier. Figure~\ref{fig:training_overview} summarizes the overall RL pipeline,
which combines adaptive task sampling and grouped online rollouts with
verifier-based credit assignment and asynchronous GRPO updates.
Decision-turn centering and token-level normalization form the
outcome-only credit-assignment path, while an optional
\emph{Process Credit Model} (PCM) further localizes verifier-derived
credit and provides process-level diagnostics.

\subsubsection{Online GUI RL Preliminaries}

\paragraph{Grouped online interaction.}
Following \S\ref{sec:formulation}, let
$\mathcal{T}=(x,\mathcal{E})$ denote a selected verifiable task. At decision turn $t$ of trajectory $i$, the rollout-policy snapshot $\pi_{\bar\theta}$ samples
\begin{equation}
y_{i,t}=(r_{i,t},a_{i,t})
\sim
\pi_{\bar\theta}\!\left(\cdot\mid c_{i,t}\right),
\qquad
c_{i,t}=(x,h_{i,t},o_{i,t}).
\end{equation}
Each response $y_{i,t}$ is one decision turn even when $a_{i,t}$ contains multiple consecutive operations. A trajectory terminates when the model emits a terminal action or reaches the interaction budget.
To construct a rollout group, we use the same fixed policy snapshot to run the selected task in multiple independent environment instances.
Each run produces one complete trajectory. We retain both successful and failed task executions, discarding only rollouts corrupted by environment or infrastructure failures. The remaining $N$ trajectories form the rollout group
$\mathcal{G}=\{\tau_i\}_{i=1}^{N}$. 
Because they share the same task and policy snapshot, their outcomes can be compared directly. 

\paragraph{Group-relative outcome supervision.}
The executable verifier for the selected task inspects the final environment state and returns a binary outcome
$R_i=R(\tau_i)\in\{0,1\}$, where $1$ indicates task success and $0$ indicates failure.
We adopt Group Relative Policy Optimization
(GRPO)~\citep{shao2024deepseekmath,Guo2025_GRPO}, which estimates relative advantages by comparing outcomes within the rollout group $\mathcal{G}$, without requiring a learned value model.
Groups in which all trajectories receive the same outcome are excluded because they provide no relative learning signal~\citep{
DBLP:conf/nips/YuZZYZYDFLLLLLL25}.

\subsubsection{Trajectory-to-Token Credit Assignment}

A verifier supplies one outcome per trajectory, while the actor is optimized over tokens produced across many decision turns. Applying the same trajectory-level GRPO advantage to every decision turn raises two issues. First, under outcome-only supervision, trajectories within the same group can contain different numbers of decision turns. Failed trajectories often contain more turns because they involve more trial and error. Broadcasting the same advantage to every turn therefore gives longer trajectories more weight, resulting in a decision-turn bias. We address this bias with decision-turn centering. 
Second, even after centering, an outcome alone does not reveal which steps are critical to success or failure; PCM uses process evidence to identify these steps and reweight their learning signals accordingly.

\paragraph{Decision-turn centering.}
To correct the decision-turn bias, we compute the group baseline under the decision-turn measure:
\begin{equation}
\mu^{\mathrm{turn}}
=
\frac{\sum_{i=1}^{N}T_iR_i}
{\sum_{i=1}^{N}T_i},
\qquad
A_{i,t}^{\mathrm{base}}
=
R_i-\mu^{\mathrm{turn}},
\end{equation}
where $T_i$ is the number of decision turns in $\tau_i$ and
$t\in\{1,\ldots,T_i\}$. 
The resulting advantage is constant within each trajectory but is indexed by decision turn. This approximately centers the advantage under the decision-turn measure to make sure $\mathbb{E}[A]\approx 0$. Within each task group, we center only the mean and do not divide by the group standard deviation. This preserves the reward scale and avoids unstable amplification when group outcomes are nearly uniform.

\paragraph{Trajectory-merge process credit.}
The outcome-only objective can optimize the policy directly from the final reward $R_i$, but the outcome does not reveal which decision steps were critical to success or failure.
As a result, every step in the same trajectory receives the same
advantage.
Building on process supervision~\citep{
DBLP:journals/corr/abs-2505-18121,
DBLP:journals/corr/abs-2602-11524,
DBLP:conf/www/XiLLZCWJZGWJGZH26}, we optionally apply PCM during RL training.
When enabled, PCM uses teacher annotations to identify critical steps and reweight their learning signals accordingly.
PCM only redistributes the learning signal across decision steps and does not change the final task-level reward $R_i$.
Without PCM, training follows the outcome-only objective without
step-level reweighting.

The teacher first extracts observable milestones from verified successful trajectories and merges equivalent subgoals and alternative valid branches into a shared task structure.
This structure is then used to annotate every trajectory.
For each failed trajectory, the teacher selects the closest valid branch and labels each milestone as \texttt{completed},
\texttt{attempted\_failed}, \texttt{not\_attempted}, or
\texttt{skipped\_by\_branch}.
The last two labels prevent blocked or irrelevant subgoals from being treated as policy failures.

At the trajectory level, failures are classified as policy-related, external, mixed, or unclear.
At the decision-turn level, annotations identify progress, causal errors, recovery, and redundancy.
Related errors are grouped to distinguish their root causes from
downstream symptoms.
If a task group contains no verified successful trajectory, PCM is not applied and training follows the outcome-only path.

Let $e_{i,t}$ denote the resulting process evidence for decision step
$t$.
PCM converts this evidence into a nonnegative importance weight:
\begin{equation}
w_{i,t}
:=
\operatorname{clip}\!\left(
b+\phi\!\left(
e_{i,t},
\operatorname{sgn}\!\left(A_{i,t}^{\mathrm{base}}\right)
\right),
0,w_{\max}
\right),
\qquad
\widetilde w_{i,t}
:=
\frac{w_{i,t}}
{\frac{1}{T_i}\sum_{u=1}^{T_i}w_{i,u}}.
\end{equation}
Here $b$ and $w_{\max}$, with $0<b\leq w_{\max}$, set the base and maximum weight scales. 
The mapping $\phi$ rewards steps that make progress or recover from mistakes in successful trajectories. In failed trajectories, it penalizes the steps that introduce an error and the later steps that fail to correct it. Redundant steps and steps blocked by the environment are also penalized. We normalize the resulting weights within each trajectory so that their average is one. If all weights become zero after clipping, we assign equal weights to all steps.

A sign-aware selector $m_{i,t}\in\{0,1\}$ retains decision turns with positive evidence or causal responsibility for failure. PCM produces
\begin{equation}
A_{i,t}^{\mathrm{proc}}
:=
m_{i,t}\,
A_{i,t}^{\mathrm{base}}\,
\widetilde w_{i,t}\,
s_i,
\end{equation}
where $s_i\in[0,1]$ discounts failures only partially attributable to the policy and equals one otherwise. 
With valid annotations, $A_{i,t}^{\mathrm{credit}}=A_{i,t}^{\mathrm{proc}}$; otherwise, $A_{i,t}^{\mathrm{credit}}=A_{i,t}^{\mathrm{base}}$. Nonselected responses remain as context but carry no actor loss. PCM thereby concentrates learning on informative decisions and provides process-level data insights.

\paragraph{Token-level normalization.}
Beyond decision-turn-level credit assignment, varying response lengths introduce another source of bias because the actor objective is computed over generated tokens. Failed trial-and-error turns often require a larger thinking budget and therefore contain more response tokens. Their advantages are consequently repeated over more tokens and can disproportionately affect the token-level objective. We address this response-length bias by normalizing the advantages over all response tokens in the optimization batch~\citep{slime_github}. This aligns the advantage statistics with token-level aggregation, controls the scale of policy updates, and further improves training stability.

\subsubsection{Asynchronous Group-Relative Optimization}

Synchronous training waits for all trajectories in a rollout group to finish.
This is inefficient because GUI rollouts vary widely in duration across environments and task horizons~\citep{xu2026singlestream}.
Instead, rollout workers use the latest available policy snapshot, and the learner starts optimization as soon as enough complete trajectories from the same task group are buffered. We never truncate individual trajectories, and all trajectories compared within a group share the same task configuration and policy snapshot.

Asynchronous training introduces policy staleness because the rollout policy may lag behind the current learner policy. 
For compactness, let $y=(y_1,\ldots,y_{|y|})$ denote a generated response, $y_{<k}$ its prefix before token $k$, $c$ its turn context, and $\bar\theta$ the rollout-policy snapshot that generated it. 
We measure the train--rollout mismatch at token $k$ by
\begin{equation}
\rho_k(\theta)
=
\frac{
\pi_\theta(y_k\mid y_{<k},c)
}{
\pi_{\bar\theta}(y_k\mid y_{<k},c)
}.
\end{equation}
where ratios $\rho_k(\theta)$ far from one indicate substantial train--rollout mismatch.
We apply IcePop~\cite{lingteam2025icepop} and SeqClip~\cite{huang-xu-wang-2026-gradloc} as complementary mismatch filters. IcePop rejects isolated tokens with abnormal likelihood ratios, whereas SeqClip evaluates the geometric mean of $\rho_k(\theta)$ over the response to detect coherent policy drift. Let $M_k\in\{0,1\}$ indicate that token $k$ passes both filters.
The resulting PPO-style clipped actor objective~\citep{schulman2017proximal} is
\begin{equation} 
\label{eq:async_clip_objective} 
\mathcal{L}_{\mathrm{RL}}(\theta) = 
-\,\mathbb{E}_{y} \left[ \frac{1}{|y|} \sum_{k=1}^{|y|} M_k \min\!\left( \rho_k(\theta)A_k, 
\operatorname{clip}\!\left( \rho_k(\theta),1-\epsilon,1+\epsilon \right)A_k \right) \right]. 
\end{equation} Here $\epsilon>0$ is the PPO clipping radius. Tokens rejected by either mismatch filter have $M_k=0$ and therefore contribute no actor loss. The mismatch filters determine whether a stale token is eligible for learning, while PPO clipping bounds the contribution of each retained token; the two mechanisms therefore address different aspects of train--rollout mismatch.

\subsubsection{Adaptive Curriculum Sampling}

As the policy improves, uniform sampling increasingly spends rollouts on solved tasks. We therefore use a domain-level \emph{adaptive curriculum sampler} that combines broad coverage with emphasis on weak application domains~\citep{DBLP:journals/corr/abs-2601-22448,DBLP:journals/corr/abs-2505-16282,DBLP:journals/corr/abs-2508-11360,DBLP:conf/iclr/QiLILSSYYY00D25,lv2026scalecua}. 
It only reallocates rollout computation over the fixed RL corpus; task construction and corpus rebalancing remain in \S\ref{sec:data}.

\paragraph{Base and adaptive allocation.}
At each training iteration, we select a fixed number of tasks, with each task producing one rollout group.
We divide this task budget between base and adaptive sampling.
Base sampling follows the domain distribution of the RL corpus to maintain broad coverage.
Adaptive sampling allocates its portion to weak domains identified from recent on-policy results.
The two components select disjoint tasks, so no task appears twice in the same batch.

\paragraph{Estimating weak domains.}
We estimate domain performance over a recent rollout window.
Only domains with enough valid trajectories and a low environment-error
rate are included, preventing infrastructure failures from being
mistaken for policy weakness.
Let $\mathcal{D}_{\mathrm{elig}}$ denote these eligible domains, and let
$V_d$ and $S_d$ be the numbers of valid and successful trajectories in
domain $d$.
We compute the domain success rate $\widehat p_d$ and the overall success
rate $\bar p$ as
\begin{equation}
\widehat p_d
=
\frac{S_d}{V_d},
\qquad
\bar p
=
\frac{
\sum_{d\in\mathcal{D}_{\mathrm{elig}}}S_d
}{
\sum_{d\in\mathcal{D}_{\mathrm{elig}}}V_d
}.
\end{equation}
A domain is considered weak when
$\bar p-\widehat p_d\geq\delta$, where $\delta>0$ is the minimum required success-rate gap.

We activate adaptive sampling only when at least two domains have
reliable estimates and at least one weak domain is identified.
Once these weak domains are identified, the adaptive component samples tasks from them without replacement, subject to a per-domain cap.
When candidate tasks have similar priority, we prefer tasks whose recent rollout groups contain both successful and failed trajectories, since these partially learned tasks provide informative group-relative signals.
If the statistics are insufficient, no weak domain is identified, or a weak domain runs out of eligible tasks, the unused adaptive budget is reassigned to base sampling.

\section{DemoCUA: Learning from In-Context Demonstrations}
\label{sec:democua}

\subsection{Demonstration Representation}

\begin{figure}[htp]
  \centering
  \includegraphics[width=\linewidth]{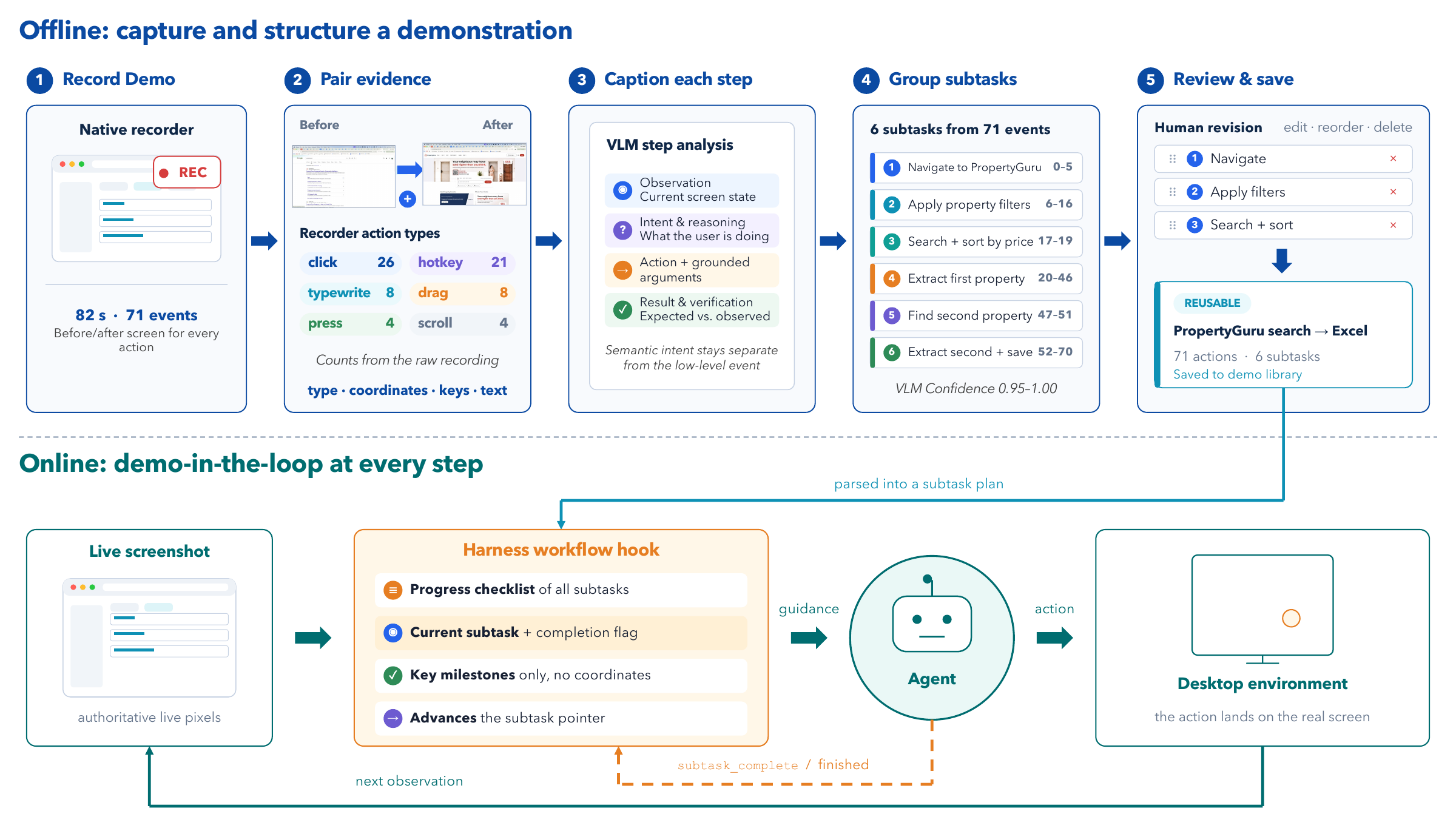}
  \caption{
    \textbf{Demonstration representation.} 
    Offline (top), a human recording is annotated and segmented into subtasks with goals and verifiable completion criteria. 
    Online (bottom), the agent follows the current subtask using live screenshots and advances upon completion. 
  }
  \label{fig:demo-workflow}
\end{figure}

\paragraph{How to Get the Demo.}
A demonstration is a recorded successful desktop execution, with every mouse and keyboard action and a screenshot before and after each one.
Its source depends on the evaluation setting: a \emph{self-demo} may be a successful rollout from a stronger GUI agent on the same task, whereas a \emph{variant-demo} is recorded by a human on a related but non-identical task.
The raw trace is normalized into a uniform action-and-frame format, then annotated offline by a vision-language model along four axes: screen state, the intent, the action taken, and how its target was located visually. 
The annotated trace is finally segmented into a few coherent subtasks, each with a short goal and an explicitly checkable completion criterion.
Figure~\ref{fig:demo-workflow} summarizes this offline pipeline and how the resulting demonstration is consumed online at every step.

\begin{figure}[htp]
  \centering
  \includegraphics[width=\linewidth]{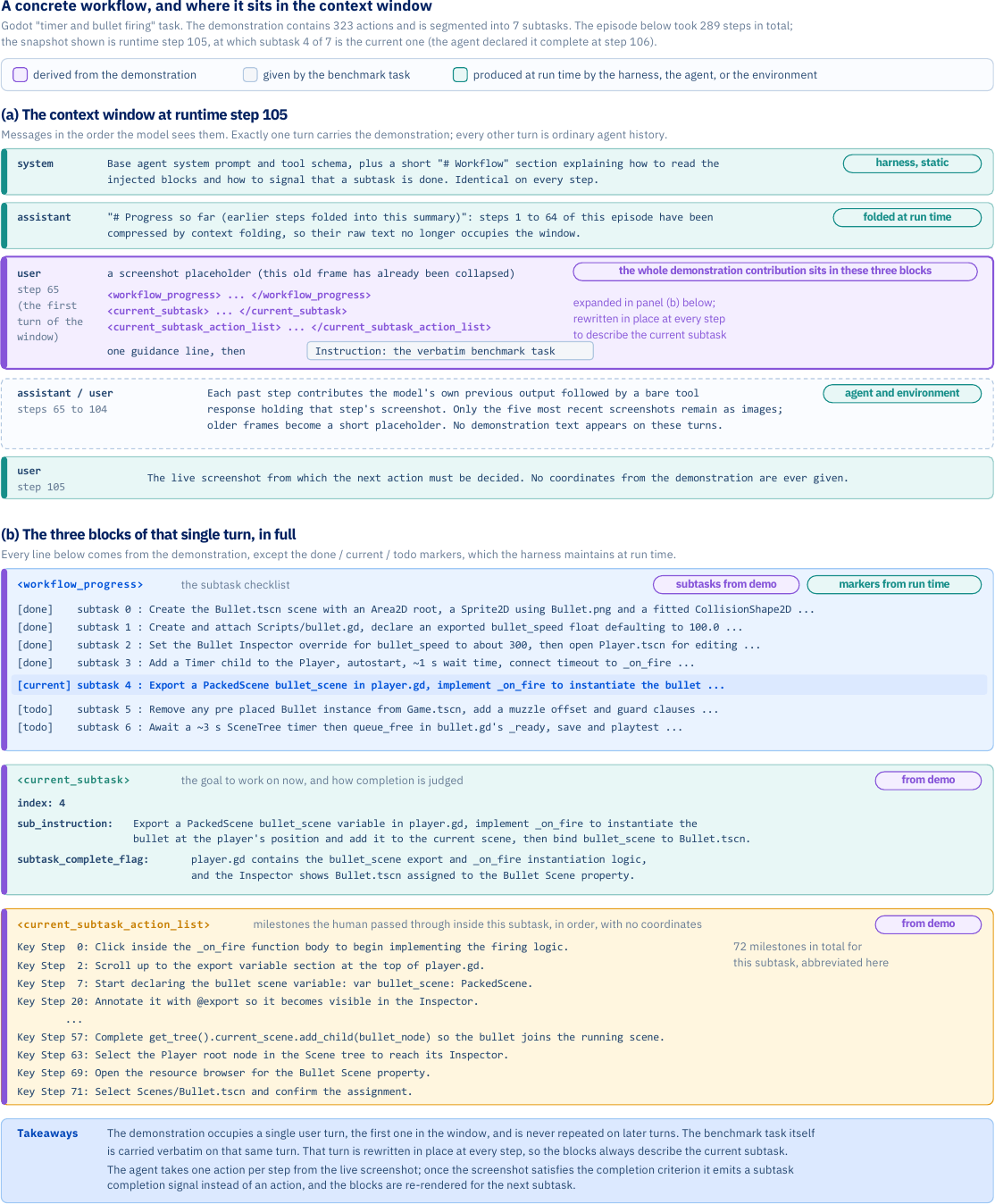}
  \caption{
    A complete demo-workflow for a Godot task in which the agent adds timer-based bullet firing to a simple game. 
    Panel (a) shows the context available to the agent at one point during execution. It includes the original task instruction, one user turn containing the demonstration, and the agent’s interaction history. 
    Panel (b) breaks down the demonstration into a subtask checklist, the current subtask and its completion criterion, and key milestones.
  }
  \label{fig:democua_context_example}
\end{figure}

\paragraph{How to Use the Demo.}
A demonstration serves as a guide rather than a script to replay. 
Before each model call, the harness provides a short summary of completed, current, and upcoming subtasks, along with key goals and milestones. 
It omits pixel coordinates and low-level actions, so the agent must rely on the live screenshot and can adapt when the interface changes. 
The agent marks each subtask as complete before moving to the next. Figure~\ref{fig:democua_context_example} shows a complete example and the context provided to the agent.

\subsection{Training with Demonstration}
\label{sec:train_w_demo}
\paragraph{Data Generation.}
We build a three-stage pipeline to generate both self-demo (i.e. identical tasks) and variant-demo (i.e. similar tasks) data, consisting of Rollout, Score, and Filter \& Repair.
(1) The Rollout stage collects observation-to-action trajectories on perturbed AgentNet data~\cite{wang2026opencua}. 
Demo-guided prompting and a subtask-report protocol are used to improve rollout success.
(2) The Score stage evaluates each rollout using a hybrid mechanism: a VLM judge assesses semantic completion, while rule-based metrics verify format validity. Each rollout is scored at three levels—trajectory, subtask, and demo-following—as detailed in Table~\ref{tab:scoring_rules}.
(3) The Filter \& Repair stage retains data that meet task-specific quality thresholds.
Instead of discarding trajectories with fixable errors, it recovers them through rule-based and VLM-based repair specified in Table~\ref{tab:repair_rules}, improving data efficiency.

\definecolor{demobg}{RGB}{246,248,252}
\definecolor{demoframe}{RGB}{188,201,222}
\lstdefinestyle{democua}{
  basicstyle=\small\ttfamily,
  backgroundcolor=\color{demobg},
  frame=single, rulecolor=\color{demoframe},
  framesep=6pt, xleftmargin=6pt, xrightmargin=6pt,
  columns=fullflexible, keepspaces=true, breaklines=true,
}
\begin{figure}[!h]
\centering
\begin{lstlisting}[style=democua]
System :  tools: computer_use(click|type|scroll|... , subtask_complete, finished)
          # Demonstration Workflow  -- contract; the live screenshot is authoritative
User 1 :  <image>
          <workflow_progress>            all subtasks; done / current / upcoming
          <current_subtask>              sub_instruction + subtask_complete_flag
          <current_subtask_action_list>  ordered action steps of THIS subtask only
          Instruction: {task}
\end{lstlisting}
\caption{DemoCUA SFT Format: the demonstration is pinned to the first user turn as a
snapshot of the workflow state.}
\label{fig:democua-sft-layout}
\end{figure}


\paragraph{Demo-Augmented Training Data Format.}
Each training sample represents one decision step in a trajectory. 
Figure~\ref{fig:democua-sft-layout} shows two changes to our standard CUA format. 
First, the system prompt adds a demonstration workflow contract and a new \texttt{computer\_use} action, \texttt{subtask\_complete}. 
Second, the first user turn includes a snapshot of the demonstration workflow at the supervised step. This snapshot contains the subtask checklist and its progress, the current subtask and its completion criterion, and its ordered action steps (see Figure~\ref{fig:democua_context_example} for an example).
The remaining format is unchanged. 
Assistant turns retain the \texttt{<think>}/\texttt{<action>}/\texttt{<tool\_call>} structure, while subsequent user turns contain only screenshots. 
Thus, the multi-turn format remains identical to the demonstration-free baseline. 
When a subtask is completed, the agent emits \texttt{subtask\_complete} after observing the post-action screenshot. Each episode ends with an explicit \texttt{finished} turn.


\paragraph{Demo-Augmented Training Data Category Ratios.}
Because the demonstration workflow and target query are closely aligned, the model may simply copy the next workflow step without checking the screenshot. 
This shortcut fails when the workflow and live interface diverge. 
To prevent this shortcut, we introduce three types of demonstration-workflow--screen relationships: 
(1) full-alignment, where the model follows the workflow; 
(2) partial-misalignment, where the model corrects mismatched steps using the screenshot;
(3) irrelevance, where it ignores the workflow and acts from the screenshot alone. 
Note that full-alignment remains the majority case so that the workflow still provides useful guidance.


\paragraph{Training from Incomplete Demonstration Workflows to Prevent Shortcut Learning.}
We keep the full trajectory as the supervision target, but show only key actions in the demonstration workflow. 
Intermediate actions, such as focus clicks, scrolling, and popup dismissal, are omitted. 
Thus, even in full-alignment cases, the model cannot simply copy the workflow; it must infer the missing actions from the screenshot. 
The workflow provides milestones, while the model learns how to reach them.



\subsection{Inference with Demonstration}
\label{sec:demo-inference}

At inference, we provide the subtask's complete action sequence without LLM-based key-action extraction. 
The action list is more detailed than during training. 
This mismatch is intentional: omitting actions during training prevents blind copying, while including them at inference provides fuller guidance once the model relies on the screenshot. 
It also removes a model call and extraction errors.

\paragraph{Long-Horizon Context Management.}
Long episodes can produce interaction histories that exceed the model's context window. 
We control context growth through proactive folding and reactive truncation. 
Before each inference call, the harness estimates the request size using approximately one token per three text characters and a fixed token cost for each screenshot. 
When the estimated usage approaches a configurable threshold, an additional LLM call summarizes the oldest interaction steps into a compact progress note. Recent steps remain verbatim, while the most recent screenshots are retained separately rather than summarized.

\paragraph{Limitations.}
A current limitation is that the workflow appears at the beginning of the context. Each subtask update therefore invalidates the shared prefix and prevents efficient KV-cache reuse. 
Future work will move the workflow to the end, allowing the context to grow append-only and reuse the cache throughout the episode.

\section{OSWorkerBench: Realistic Cross-Application Office Workflows with Multimodal Demonstrations}
\label{sec:OSWorkerBench}

Most existing GUI benchmarks evaluate agents under an instruction-only protocol, requiring them to infer and execute a workflow from a natural-language request~\cite{xie2024osworld,bonatti2025windowsagentarena}. While this setting is essential for measuring general computer-use competence, it does not directly test whether an agent can use an example of how a procedure should be carried out. This capability is particularly important for personalized workflows, proprietary or organization-specific applications, and complex processes whose conventions may be absent from public training data and cumbersome to specify completely in text. In such settings, a demonstration can expose both the intended actions and the resulting visual state transitions. To evaluate demonstration-guided execution alongside standard instruction following, we introduce \textbf{OSWorkerBench}, \textit{an office-centric benchmark of long-horizon, cross-application workflows in enterprise productivity environments}.

OSWorkerBench comprises 100 realistic office tasks spanning 41 normalized applications and 10 consolidated job families (Figure~\ref{fig:WorkerBench_job_families}), all of which support standard instruction-only evaluation. To capture the complexity of real-world office work, we further identify two independently annotated, potentially overlapping capability subsets: 67 \textbf{Long-Memory} tasks require delayed reuse of dynamic information or sustained tracking of constraints and workflow state, while 49 \textbf{Multi-App} tasks require faithful transfer of dynamic, multi-field information across at least three logical applications. OSWorkerBench supports two evaluation modes: \textbf{\emph{instruction-only}}, in which the target instruction is the only procedural input, and \textbf{\emph{demonstration-guided}}, in which one multimodal demonstration is additionally supplied. The demonstration-guided mode contains two settings with different purposes. The \textbf{\emph{self-demo}} setting pairs 33 targets with successful strong-agent rollouts of those same tasks and is used for the quantitative DemoCUA evaluation in Section~\ref{sec:democua-evaluation-setup}. The more challenging \textbf{\emph{variant-demo}} setting pairs 45 targets with human recordings of semantically related but non-identical source tasks. The numbers 33 and 45 therefore refer to different demonstration collections, not to a partition of the 100 benchmark tasks.  The 45 variant-demo targets are deliberately challenging and long-horizon: under instruction-only evaluation, Kimi-2.6 requires more than 100 observation-to-action decision turns per task on average before termination (see Section~\ref{sec:decision_turn_analysis} for the decision-turn definition and full trajectory analysis). These cases are not demonstration-only tasks: the same targets and evaluators can be used with and without guidance. To the best of our knowledge, \textbf{OSWorkerBench is the first CUA benchmark to provide multimodal demonstrations as explicit one-shot guidance during evaluation}. We report systematic results for the 33-task self-demo setting in Section~\ref{sec:evaluations}; systematic evaluation of the 45-task variant-demo setting is left to future work.

\subsection{Benchmark Design and Construction Pipeline}

\begin{figure*}[t]
  \centering
  \includegraphics[width=\textwidth]{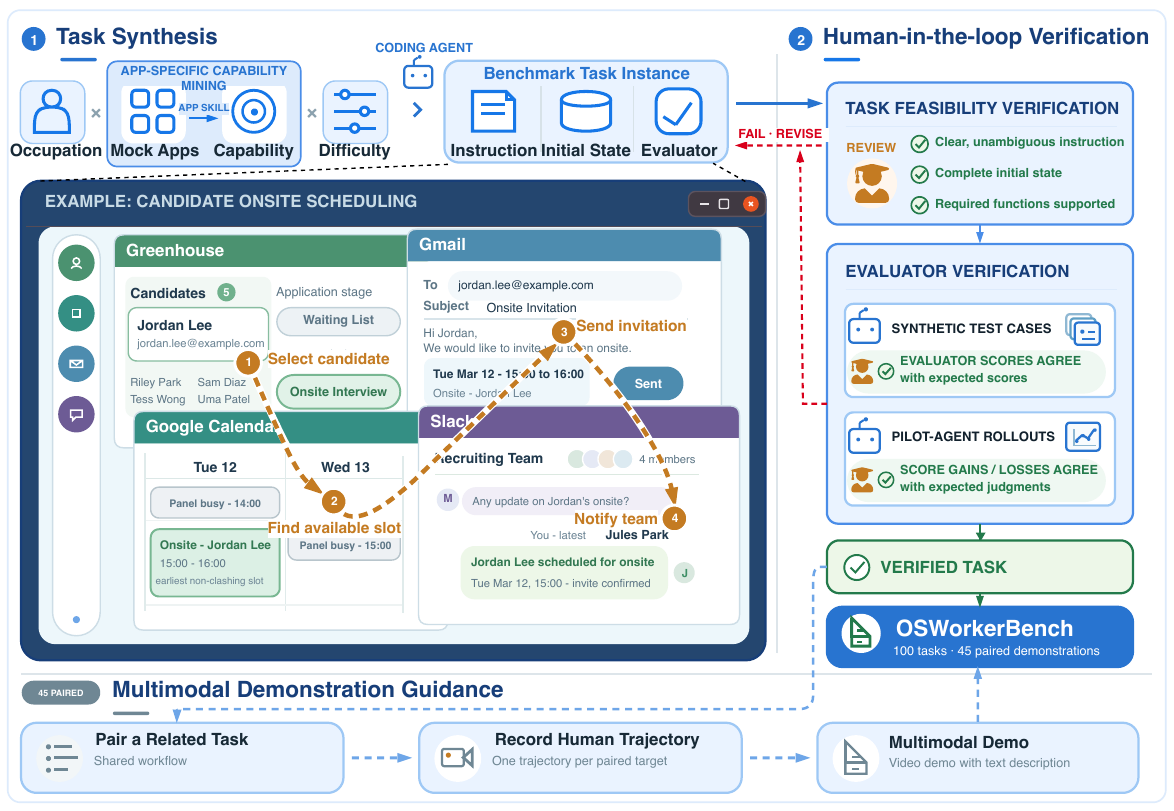}
  \caption{\textbf{OSWorkerBench benchmark construction pipeline.} Candidate task instances are synthesized and validated through human review, evaluator tests, and pilot-agent rollouts. Forty-five targets are additionally paired with human-recorded demonstrations of related but non-identical tasks for the variant-demo setting. The 33 same-task strong-agent rollouts used in the self-demo evaluation are constructed separately and are not depicted.}
  \label{fig:workerbench_pipeline}
\end{figure*}

Figure~\ref{fig:workerbench_pipeline} provides an overview of the OSWorkerBench construction pipeline, which proceeds from capability-grounded task synthesis and dense evaluator generation through human-in-the-loop verification and, for 45 selected targets, human-recorded variant-demo pairing. The 33 self-demos are produced separately by retaining successful strong-agent rollouts on the finalized target tasks.

\paragraph{Capability-grounded workflow synthesis.}
OSWorkerBench synthesizes mock-app tasks from an occupation, a set of applications, verified application capabilities, and a target difficulty. For each application, we inspect its implemented UI and state schema to identify functions that are both executable through the interface and programmatically observable, and use these verified functions as workflow building blocks. The synthesizer requires at least one explicit cross-application dependency rather than merely co-locating unrelated actions. In the onsite-scheduling example in Figure~\ref{fig:workerbench_pipeline}, the candidate selected in Greenhouse determines what must be scheduled, Calendar availability constrains the Gmail invitation, and the confirmed schedule supplies the content of the Slack notification.

Difficulty is assigned during synthesis rather than inferred retrospectively from model performance. We control three complementary axes: \emph{breadth}, the number of applications and transitions among them; \emph{depth}, the number and diversity of nontrivial UI access points that require navigation or discovery; and \emph{reasoning}, the amount of task state that must be derived, filtered, or deliberately left unchanged rather than directly copied. A task's construction-time difficulty profile jointly varies these axes through its applications and information transfers, access points, records and distractors, decision branches, dependencies, and estimated human-reference horizon. Harder tasks increase multiple dimensions together rather than scaling any single factor in isolation. The horizon is estimated from the GUI actions in a complete human reference solution and is not an observed model trajectory length.

Once the workflow and difficulty profile are fixed, a coding-agent pipeline realizes the specification as a natural-language instruction, a deterministic initial state, and an executable evaluator. The setup and evaluator are authored separately under a shared task and answer-key contract, while CUA-Gym's state-injection design provides an isolated, resettable session for reproducible execution~\cite{wang2026cua}.

\paragraph{Dense evaluator generation and human-in-the-loop verification.}
To evaluate task outcomes without requiring imitation of a reference trajectory, each evaluator measures functional outcomes rather than similarity to a reference trajectory by decomposing task completion into independently verifiable checkpoints. Record-level checks detect missing, incorrect, or extra outputs; gates enforce prerequisites among dependent outcomes; and weights prioritize primary business outcomes over cross-application outputs and auxiliary confirmations. Most checkpoints compare the final and initial application states, including session-scoped mock-application data exposed through the unified API. When backend state is insufficient, task-specific evaluators inspect spreadsheet cells, generated files, images, compound documents, or conditional outputs. Evaluators contain 1--13 checkpoints (mean 4.86; median 5), enabling dense progress measurement while reserving strict success for tasks satisfying all required final-state conditions.

Before deployment, each evaluator is tested on controlled terminal states. The untouched initial state must score $0$ and a golden completion $1$. Checkpoint-specific and valid partial states test intermediate scores, while negative cases introduce missing prerequisites, wrong values or branches, distractors, and over-action errors. Each case applies a controlled mutation to the initialized state, with its expected score fixed from the rubric before execution. The production evaluator is then run unchanged across all cases, and a per-task manifest records every mutation and expected score for reproducibility.

Human reviewers additionally verify instruction clarity, setup completeness, task feasibility, checkpoint coverage, weights, and expected partial scores. As an independent execution check, representative agents, including Kimi-2.6 and other CUAs, run each task from its initialized environment. Reviewers compare their trajectories and terminal states against checkpoint-level and aggregate scores. Any mismatch triggers revision of the instruction, setup, or evaluator, followed by rerunning all affected validation checks.

\subsection{Multimodal Demonstration Guidance}

\paragraph{Self-demo setting.}
The self-demo collection contains 33 multi-application targets used in the reported DemoCUA evaluation.
For each target, a stronger GUI agent first completes that same task successfully.
Its rollout is converted into a multimodal demonstration that preserves the screenshots and action types while omitting concrete pixel coordinates from the workflow shown to the evaluated agent.
The target is then reset and evaluated with and without this same-task demonstration.
Self-demo measures how effectively a model can use a demonstrated execution path; because the source and target tasks are identical, it does not by itself measure procedural transfer across task variants.

\paragraph{Variant-demo setting.}
OSWorkerBench additionally selects 45 procedurally demanding, multi-stage targets for demonstration-guided procedural transfer.
These workflows require information obtained in one application to guide subsequent decisions or artifacts in another, testing intermediate-state retention, stage ordering, and reliable cross-application handoffs.
Each target is paired with one successful human demonstration from a semantically related but non-identical source task.
Source and target share a transferable workflow structure but differ in entities, values, initial states, and execution details, preventing direct replay.
Each raw demonstration contains its source instruction and, for every tool call, the pre-action screenshot, action type and arguments, and post-action screenshot.
Natural-language reasoning is excluded.
Before inference, the trace is converted into the coordinate-free subtask workflow described in Section~\ref{sec:democua}.
These 45 human-recorded pairs define the variant-demo benchmark resource; they are distinct from the 33 same-task demonstrations used for the reported self-demo results.

\paragraph{Controlled evaluation protocols.}
All 100 tasks can be evaluated under the \textbf{\emph{instruction-only}} protocol.
Under either demonstration-guided setting, the associated targets are rerun after adding exactly one demonstration to the context.
Guided and unguided runs use identical target instructions, initialized environments, interaction budgets, and evaluators; only demonstration availability differs.
On the 33-task self-demo set, the paired difference measures the value of same-task execution guidance.
On the 45-task variant-demo set, it measures transfer of an observed procedure to a related but non-identical task.
The remaining 55 tasks lack a variant-demo pairing but remain part of the full instruction-only benchmark.

\subsection{Task Coverage and Characteristics}

\begin{figure*}[t]
  \centering
  \begin{subfigure}[t]{0.49\textwidth}
    \centering
    \includegraphics[width=\linewidth]{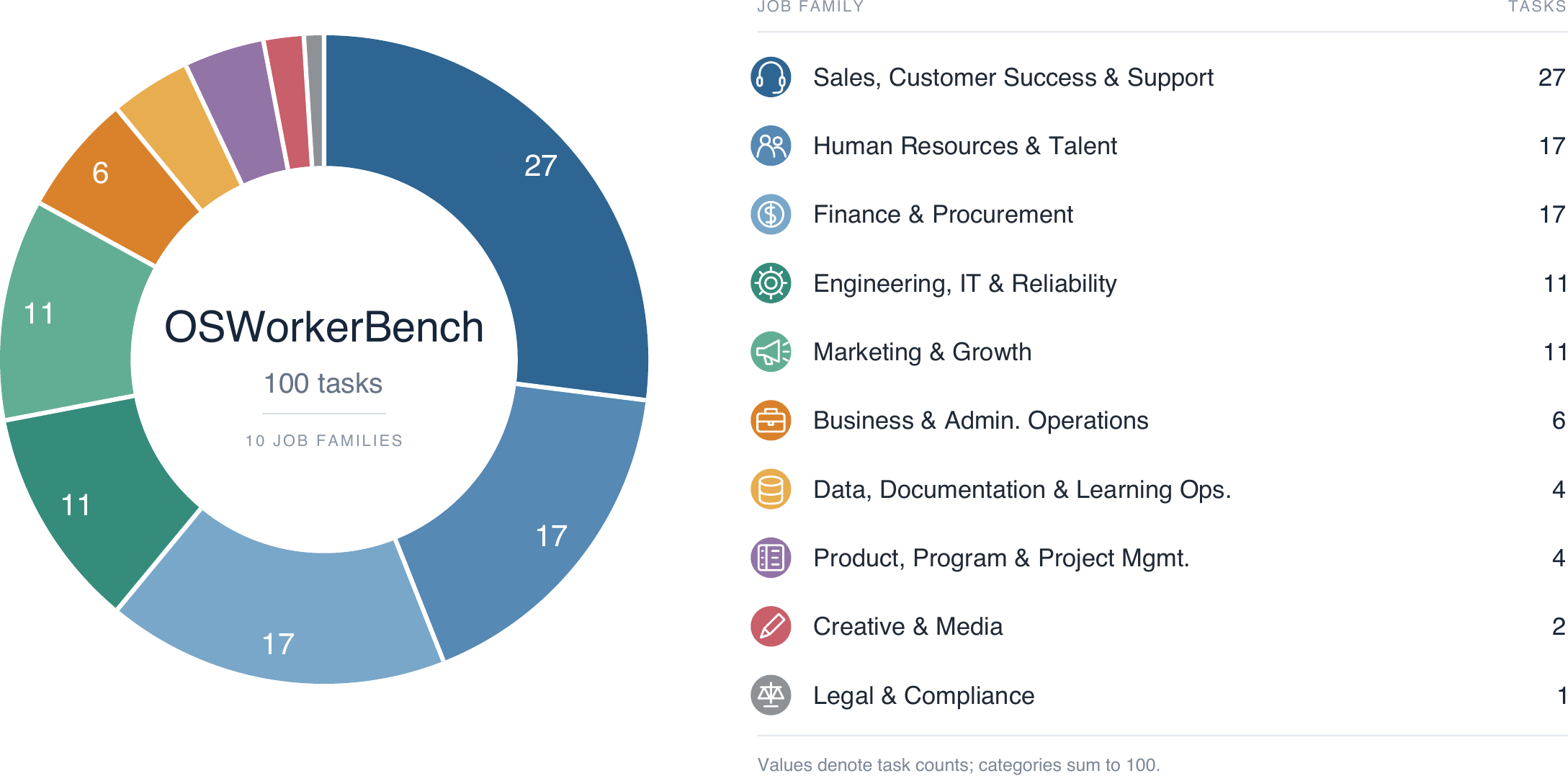}
    \caption{Primary job-family distribution. Each task is assigned to exactly one family according to its main business objective.}
    \label{fig:WorkerBench_job_families}
  \end{subfigure}
  \hfill
  \begin{subfigure}[t]{0.49\textwidth}
    \centering
    \includegraphics[width=\linewidth]{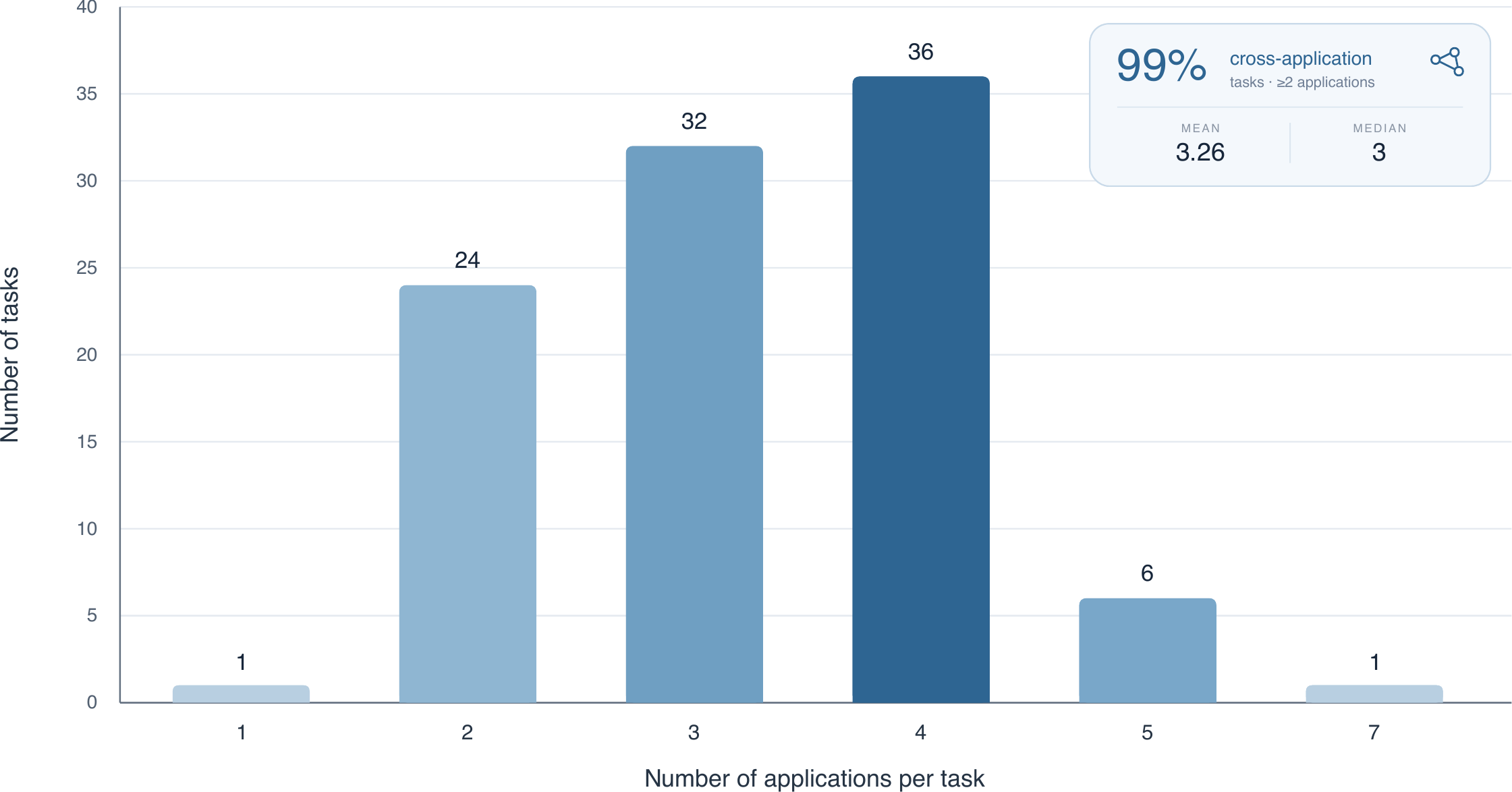}
    \caption{Number of distinct required applications per task across the full benchmark.}
    \label{fig:WorkerBench_app_breadth}
  \end{subfigure}

  \vspace{2mm}
  \begin{subfigure}[t]{0.49\textwidth}
    \centering
    \includegraphics[width=\linewidth]{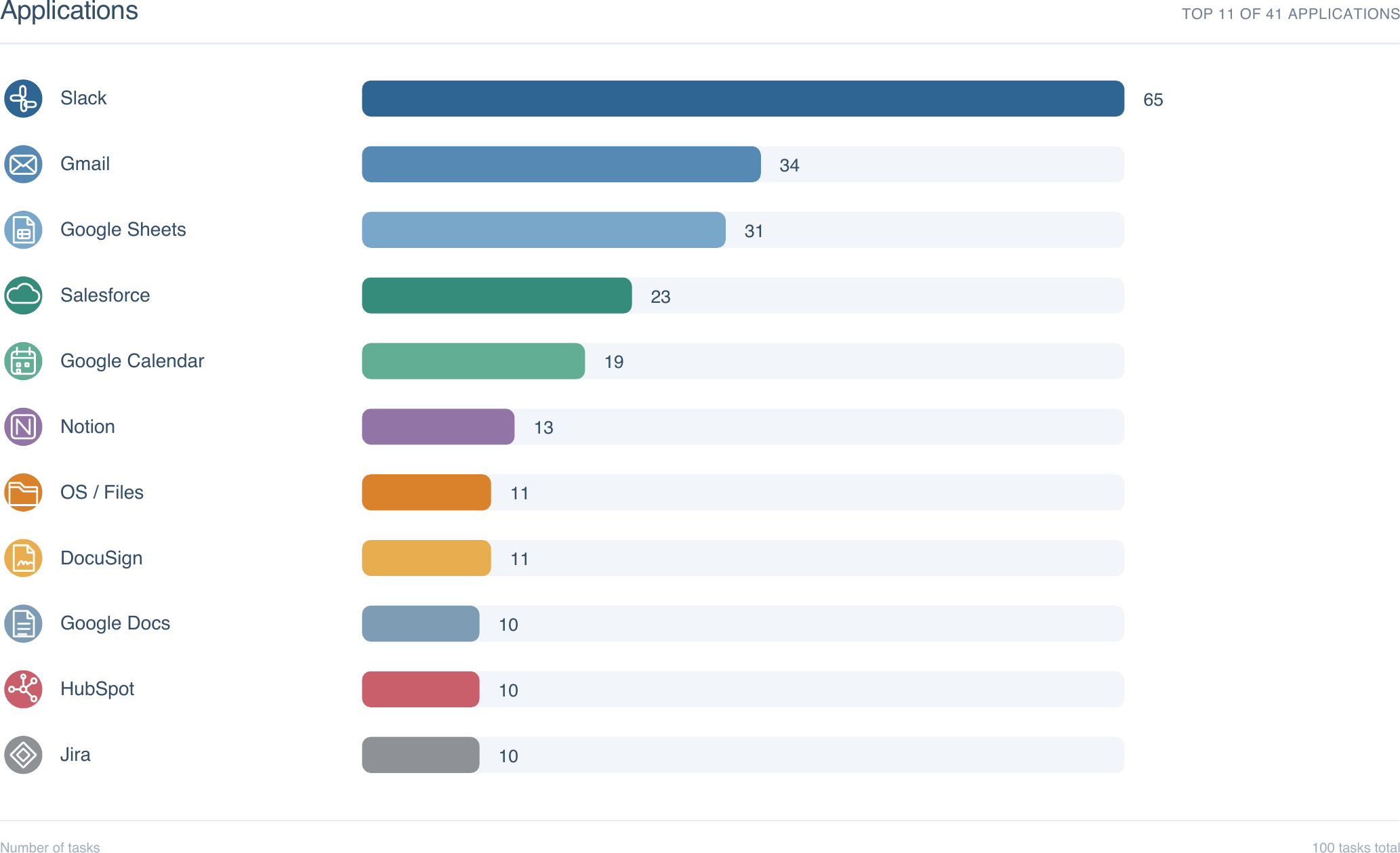}
    \caption{Most frequent normalized applications across the benchmark.}
    \label{fig:WorkerBench_app_frequency}
  \end{subfigure}
  \hfill
  \begin{subfigure}[t]{0.49\textwidth}
    \centering
    \includegraphics[width=\linewidth]{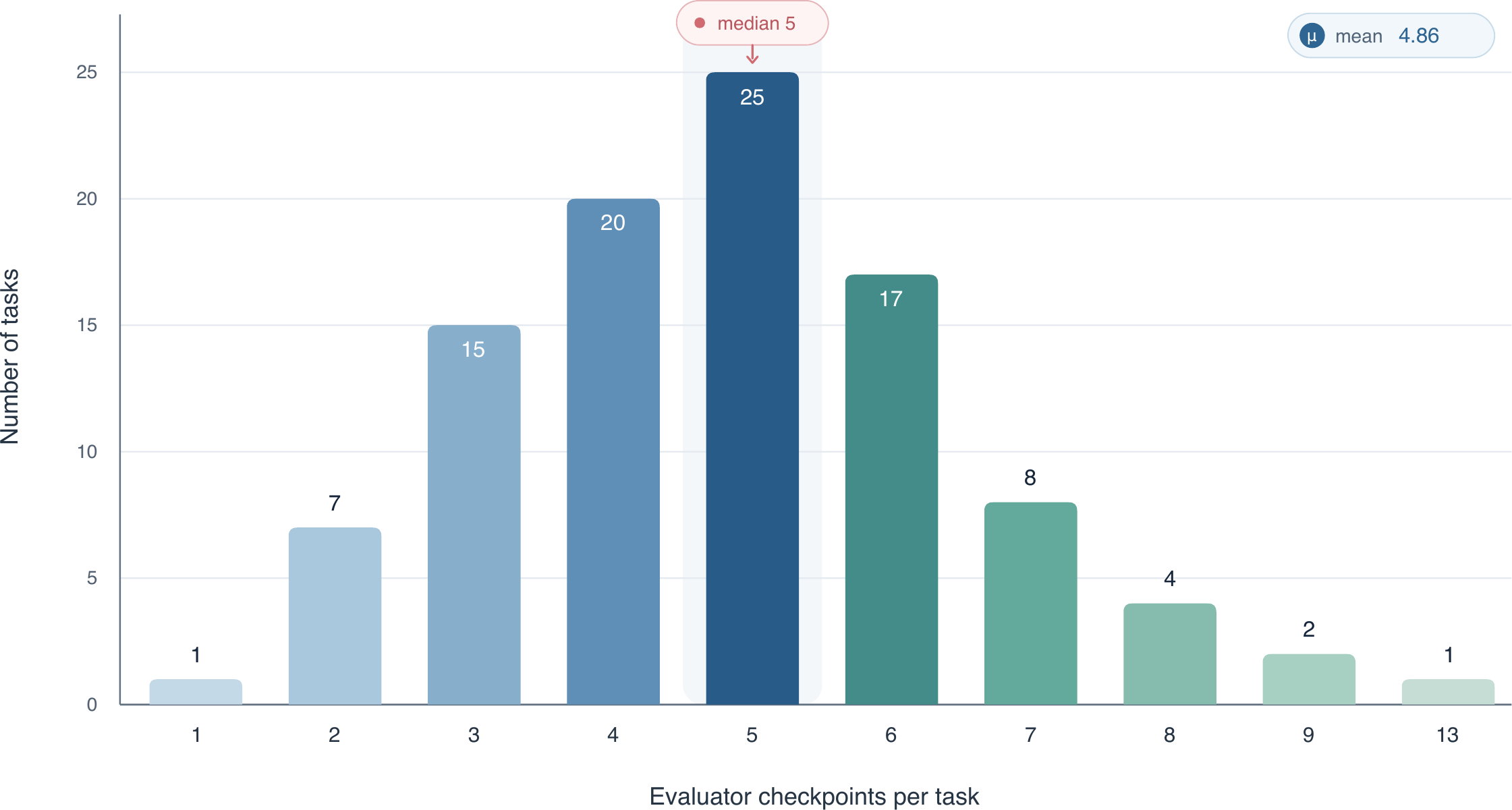}
    \caption{Distribution of evaluator checkpoints per task.}
    \label{fig:WorkerBench_evaluator_granularity}
  \end{subfigure}

  \caption{\textbf{OSWorkerBench dataset overview.} All statistics are computed over the canonical 100-task set. Application aliases and mock variants are normalized before counting.}
  \label{fig:WorkerBench_overview}
\end{figure*}

We assign each task to one mutually exclusive job family according to its primary business outcome, rather than the applications it happens to use. As shown in Figure~\ref{fig:WorkerBench_job_families}, sales, customer success, and support form the largest family with 27 tasks. Human resources and talent and finance and procurement each contribute 17 tasks, followed by engineering, IT, and reliability and marketing and growth with 11 tasks each. The remaining 17 tasks cover business and administrative operations, data, documentation and learning operations, product/program/project management, creative and media work, and legal and compliance. This distribution reflects a deliberate focus on enterprise work while retaining a meaningful long tail of professional scenarios.

\paragraph{Capability-oriented task slices.}
Job-family and application statistics describe where a task is situated, but they do not reveal the information dependencies that make the workflow difficult. We therefore annotate every task along two complementary requirement-level dimensions: whether successful completion demands long-term retention of task state, and whether it demands substantive information transfer across applications. These tags characterize properties of the task specification rather than the behavior of a particular agent: they are assigned independently of model scores and observed failure modes. Each case is semantically reviewed against its instruction and available setup, evaluator, and trajectory evidence, while automated scripts are used only to align evidence, count applications, and validate label consistency.

\paragraph{Long-memory workflows (67 cases).}
Real office processes often separate the moment when information is discovered from the moment when it must be used. We label a task as Long-Memory when it requires either (i) reading or deriving a dynamic value, retaining it across substantive intermediate operations or application switches, and reproducing it consistently in a later evaluated outcome; or (ii) maintaining multiple constraints, decisions, completed stages, and pending items across a multi-stage workflow. Both patterns create an explicit dependency between earlier observations and later actions. Merely producing a long trajectory, taking detours, repeating clicks, or eventually failing does not qualify a task for this subset.

\paragraph{Cross-application information flow (49 cases).}
Likewise, opening several applications does not by itself constitute meaningful cross-application reasoning: the actions may be independent and require no information to flow between tools. We reserve the Multi-App tag for tasks that involve at least three distinct logical applications and require the agent to obtain at least two dynamic facts, a multi-field record collection, or a substantive passage of text from one application and faithfully reproduce, organize, or transform it in another. A single transferred identifier, constants already supplied in the instruction, and unrelated fixed actions across several applications are excluded. Application counts include only tools that must be read or modified to complete the evaluated workflow; browser and desktop shells, incidental exploration, and background-only applications are excluded, while multiple tabs or windows of the same product count once. The Long-Memory and Multi-App subsets are assigned independently and may overlap: the former captures temporal dependencies within a workflow, whereas the latter captures information dependencies across tools.

\paragraph{Cross-application breadth.}
Because demonstrations alter the evaluation context rather than the required workflow, Figure~\ref{fig:WorkerBench_app_breadth} aggregates all 100 tasks. Cross-application execution is a defining property of OSWorkerBench: 99 tasks require at least two applications, 68 require three or four, and the mean is 3.26 applications per task (median 3; maximum 7). This count is broader than the 49-task Multi-App subset, which additionally requires dynamic information transfer across applications.

Application frequencies exhibit a hub-and-spoke structure (Figure~\ref{fig:WorkerBench_app_frequency}). Slack appears in 65 tasks, followed by Gmail (34), Google Sheets (31), Salesforce (23), and Google Calendar (19), where the most common pairs are Google Sheets--Slack and Gmail--Slack (23 tasks each). These hubs support communication and state synchronization across a long tail of 41 applications. Together, application breadth and evaluator granularity characterize intrinsic task structure, with evaluators containing 1--13 checkpoints (Figure~\ref{fig:WorkerBench_evaluator_granularity}).

\paragraph{Long-horizon interaction difficulty.}
OSWorkerBench is deliberately designed around realistic office workflows that unfold over many dependent stages rather than isolated GUI operations. To reach the final business outcome, an agent must carry forward dynamically discovered information, preserve constraints across application switches, and track both completed and pending steps throughout the workflow. This structure produces genuinely long interaction horizons: in the Kimi-2.6 rollouts, the median trajectory contains 68 observation--decision turns (mean 88.3), and 38 of the 100 trajectories extend to at least 100 turns. OSWorkerBench therefore directly stresses long-memory, cross-application state tracking, and sustained end-to-end workflow control. Section~\ref{sec:decision_turn_analysis} provides the decision-turn definition and full trajectory analysis.

\subsection{Executable Evaluation}

Each task is paired with an evaluator that measures functional outcomes rather than similarity to a reference trajectory. Eighty-eight tasks use state-based evaluators over initialized enterprise application backends. The remaining 12 use task-specific evaluators for spreadsheets, images, compound document deliverables, or conditional workflows. This mixture preserves deterministic final-state verification while accommodating outputs whose correctness cannot be represented as a single application-state predicate.

We report \emph{strict task success}, which requires all final-state conditions, and \emph{partial progress}, the checkpoint-weighted score
$S_{\mathrm{partial}}=(\sum_i w_i c_i)/(\sum_i w_i)$, where $c_i\in[0,1]$ is the score and $w_i$ the weight of checkpoint $i$. For the Long-Memory and Multi-App subsets, we report only strict success because task-level labels may correspond to only a subset of checkpoints, making partial scores capability-ambiguous. Guided and unguided runs use identical evaluators, ensuring that measured gains reflect task outcomes rather than action imitation. We plan to release the task specifications, the 33 self-demo and 45 variant-demo pairings, taxonomy metadata, and evaluators.

\section{Evaluations}
\label{sec:evaluations}

\begin{table*}[t]
  \tablesize
  \centering
  \renewcommand{\arraystretch}{1.3}
  \caption{\textbf{Overall average scores (\%) on OSWorld-Verified.} We report the overall average score of each model under the OSWorld evaluation protocol. Size denotes the publicly disclosed model scale when available. Higher is better. Our models are shaded. $^{*}$ denotes a result reproduced by following the official OSWorld repository.}
  \label{tab:osworld_results}
  \begin{tabularx}{0.75\textwidth}{
      >{\raggedright\arraybackslash}p{5.2cm}
      >{\centering\arraybackslash}X
      >{\centering\arraybackslash}X
  }
    \toprule
    \textbf{Model} & \textbf{Size} & \textbf{Average Score} \\
    \midrule

    \rowcolor[gray]{0.92}
    \multicolumn{3}{l}{\emph{General-purpose multimodal models}} \\
    Kimi-K2.6                & 1T-A32B & 73.1 \\
    Qwen3.7-Plus  & Closed-source & 73.3 \\
    Qwen3.6-27B             & 27B     &  52.5$^{*}$ \\
    GPT-5.5                 & Closed-source & 78.7 \\
    Claude Sonnet 5         & Closed-source & 81.2 \\
    Claude Opus 4.8         & Closed-source & 83.4 \\
    \rowcolor[gray]{0.92}
    \multicolumn{3}{l}{\emph{Specialized computer-use agents}} \\
    EvoCUA              & 32B     & 56.7 \\
    UI-TARS-1.5         & 7B      & 25.4 \\
    ScaleCUA-Qwen3.5     & 9B      & 68.7 \\
    \midrule
    \rowcolor[gray]{0.92}
    \multicolumn{3}{l}{\emph{Our models}} \\
    \rowcolor{our_blue!20}
    \textbf{UI-Mate-9B (Ours)}  & 9B  & 66.2 \\
    \rowcolor{our_blue!20}
    \textbf{UI-Mate-27B (Ours)} & 27B & 77.0 \\
    \bottomrule
  \end{tabularx}
\end{table*}

\subsection{Evaluation Setup}
\subsubsection{Benchmarks}
\paragraph{Public Benchmarks.}
We evaluate UI-Mate on OSWorld-Verified~\cite{xie2024osworld} and WindowsAgentArena (WAA)~\cite{bonatti2025windowsagentarena} as public references for general computer-use performance.

\paragraph{OSWorkerBench.}
We additionally evaluate on OSWorkerBench (Section~\ref{sec:OSWorkerBench}).
Under the instruction-only protocol, all models receive only the target instruction, and we report strict binary success and checkpoint-based progress over all 100 tasks.
We further report binary success on two overlapping requirement-based subsets: 67 Long-Memory tasks and 49 Multi-App tasks.
Demonstration-guided evaluation uses separate pairings: the reported quantitative experiment uses 33 same-task strong-agent rollouts in the self-demo setting, while the benchmark also provides 45 human-recorded demonstrations of related but non-identical tasks for the variant-demo setting. A systematic aggregate result on the 45-task variant-demo set is left for future work exploration.

\subsubsection{Baselines}
We compare UI-Mate with representative baselines, divided into two groups. The general-purpose group comprises Kimi-K2.6~\cite{moonshotai2026kimik26}, Qwen3.6-27B~\cite{qwen2026qwen36}, Qwen3.5-9B~\cite{qwen2026qwen35}, GPT-5.5~\cite{openai2026gpt55}, Claude Sonnet 5~\cite{anthropic2026claudesonnet5}, and Claude Opus 4.8~\cite{anthropic2026claudeopus48}. These models span different scales, model families, and deployment regimes, providing strong reference points for computer-use capabilities that emerge from broadly trained multimodal models. The specialized group includes EvoCUA-32B~\cite{xue2026evocua}, UI-TARS-1.5-7B~\cite{qin2025ui}, and ScaleCUA-Qwen3.5-9B~\cite{lv2026scalecuascalingcomputeruse}, each explicitly trained or adapted for graphical user-interface interaction. Together, these baselines position UI-Mate relative to both general-purpose multimodal models and specialized computer-use agents. Because the systems differ in architecture, training data, and agent scaffolding, the comparisons reflect end-to-end system performance rather than the isolated effect of model scale or computer-use specialization. Unless otherwise stated, all systems are evaluated in the same initialized target environments with identical task instructions, interaction budgets, and executable evaluators. Where a demonstration-guided comparison is reported, each system receives the same demonstration through its model-specific input format; the quantitative OSWorkerBench comparison in this report uses the 33-task self-demo collection.

\begin{table*}[t]
  \tablesize
  \centering
  \renewcommand{\arraystretch}{1.3}
  \caption{\textbf{Main results on OSWorkerBench under the \emph{instruction-only} protocol.} We report the average checkpoint-based progress score across all 100 tasks, followed by strict binary success rates on the overlapping requirement-based Multi-App (49 tasks) and Long-Memory (67 tasks) subsets and on the full benchmark. Models are grouped by weight availability, with model sizes omitted when they are not publicly disclosed. All values are percentages, and higher is better. The best baseline result in each column is shown in bold.}
  \label{tab:OSWorkerBench_results}

  \begin{tabularx}{\textwidth}{
    >{\raggedright\arraybackslash}p{3.8cm}
    >{\centering\arraybackslash}p{1.2cm}
    >{\centering\arraybackslash}X
    @{}
    *{3}{>{\centering\arraybackslash}X}
  }
    \toprule
    \multirow{2}{*}{\textbf{Model}} &
    \multirow{2}{*}{\makecell{\textbf{Model}\\\textbf{Size}}} &
    \multicolumn{1}{c}{\textbf{Progress Score} (\%)} &
    \multicolumn{3}{c}{\textbf{Binary Success Rate} (\%)} \\
    \cmidrule(lr){3-3}\cmidrule(lr){4-6}

    & & \makecell{\textbf{Overall}\\\textbf{(100)}}
    & \makecell{\textbf{Multi-}\\\textbf{App (49)}}
    & \makecell{\textbf{Long-}\\\textbf{Memory (67)}}
    & \makecell{\textbf{Overall}\\\textbf{(100)}} \\
    \midrule

    \rowcolor[gray]{0.92}
    \multicolumn{6}{l}{\emph{Closed-weight models}} \\
    GPT-5.6-Sol
      & & \textbf{87.67} & \textbf{65.31} & \textbf{67.16} & \textbf{71.00} \\
    Claude Opus 4.8
      & & 81.54 & 53.06 & 55.22 & 62.00 \\
    Claude Sonnet 5
      & & 81.46 & 42.86 & 50.75 & 55.00 \\

    \midrule
    \rowcolor[gray]{0.92}
    \multicolumn{6}{l}{\emph{Open-weight models}} \\
    UI-TARS-1.5~\cite{qin2025ui}
      & 7B & 9.22 & 0.00 & 0.00 & 4.33 \\
    Qwen3.5~\cite{qwen2026qwen35}
      & 9B & 18.11 & 2.04 & 1.49 & 5.05 \\
    EvoCUA~\cite{xue2026evocua}
      & 32B & 37.62 & 2.04 & 4.48 & 16.00 \\
    \mbox{ScaleCUA-Qwen3.5~\cite{lv2026scalecuascalingcomputeruse}}
      & 9B & 38.27 & 3.40 & 7.96 & 16.33 \\
    Qwen3.6~\cite{qwen2026qwen36}
      & 27B & 52.35 & 7.48 & 12.94 & 23.33 \\
    Kimi-K2.6~\cite{moonshotai2026kimik26}
      & 1T & 72.42 & 18.37 & 25.37 & 40.67 \\

    \addlinespace[1pt]
    \rowcolor{our_blue!20}
  [\dimexpr\tabcolsep+0.8pt\relax]
  [\dimexpr\tabcolsep+0.8pt\relax]
\textbf{UI-Mate (Ours)}
  & 9B & 66.55 & 16.33 & 25.37 & 34.00 \\
\rowcolor{our_blue!20}
  [\dimexpr\tabcolsep+0.8pt\relax]
  [\dimexpr\tabcolsep+0.8pt\relax]
\textbf{UI-Mate (Ours)}
  & 27B & \textbf{76.86} & \textbf{28.57}
  & \textbf{32.84} & \textbf{41.00} \\

    \bottomrule
  \end{tabularx}
\end{table*}

\subsubsection{Evaluation Protocols}

We use benchmark-specific evaluation configurations as described below.

\paragraph{OSWorld-Verified.} We evaluate our model on OSWorld-Verified~\cite{xie2024osworld}, a benchmark designed to assess multimodal agents on open-ended tasks in real computer environments. OSWorld-Verified performance reflects the complete interactive process: interpreting visual observations, grounding actions in the graphical interface, executing multi-step operations, and maintaining progress over long-horizon workflows.
We follow the official OSWorld codebase and evaluation protocol. 
Evaluation is conducted using the official Docker/AWS provider configurations. The Docker provider supplies isolated, reproducible desktop environments with KVM acceleration where available, while the AWS provider uses OSWorld’s official host-client architecture to support large-scale parallel evaluation. Parallel execution affects only evaluation throughput and does not change individual task configurations or scoring criteria. We report the end-to-end task success rate calculated by the official evaluators.

\paragraph{WindowsAgentArena.}
To further assess cross-platform generalization, we evaluate our model on WindowsAgentArena (WAA)~\cite{bonatti2025windowsagentarena}, which tests computer-use agents on Windows applications and system-specific configurations. We follow OS-SYMPHONY's evaluation protocol for WAA and use the corresponding OSWorld configurations as references for model-specific settings. Our implementation supports both WAA's original Docker-based environment and our internal sandbox. We report the end-to-end task success rate computed by WAA's task-specific evaluators.

\paragraph{OSWorkerBench.}
To support long-horizon state tracking, UI-Mate-27B retains its generated thinking throughout the interaction history, keeping intermediate constraints, decisions, and pending actions available during later stages of a workflow. For the baseline agents with instruction-only setup, we follow their corresponding OSWorld evaluation configurations and adapt them to the OSWorkerBench environment without changing their model-specific interaction mechanisms. All agents are evaluated with a maximum budget of 200 interaction steps per task.

\subsection{Main Results}
\subsubsection{OSWorld-Verified}

\paragraph{UI-Mate is competitive with general-purpose models and advances specialized computer-use agents.}
As shown in Table~\ref{tab:osworld_results}, UI-Mate-27B achieves an average score of $77.0\%$ on OSWorld-Verified, outperforming the general-purpose Kimi-K2.6 ($73.1\%$) and Qwen3.7-Plus ($73.3\%$) while approaching GPT-5.5 ($78.7\%$) with a gap of $1.7\%$. Among specialized agents, it exceeds ScaleCUA-Qwen3.5 ($68.7\%$), EvoCUA-32B ($56.7\%$), and UI-TARS-1.5 ($25.4\%$). UI-Mate-9B reaches $66.2\%$, slightly below ScaleCUA-Qwen3.5-9B but above the larger EvoCUA-32B by $9.5\%$. These comparisons indicate that environment-grounded computer-use training can deliver performance competitive with substantially larger general-purpose systems and that model scale alone does not determine agent capability.

\paragraph{Scaling primarily improves application-level workflow execution.}
UI-Mate-9B and UI-Mate-27B obtain the same OS score of $91.7\%$, suggesting that the smaller model already acquires strong basic operating-system interaction capabilities. The 27B model's overall gain of $10.8\%$ instead comes primarily from Office ($+11.9\%$), Daily ($+12.7\%$), Professional ($+6.1\%$), and Workflow ($+13.0\%$) tasks. Thus, increasing model capacity contributes less to atomic OS control than to coordinating longer, application-specific execution paths. UI-Mate-9B nevertheless reaches $66.2\%$ and surpasses EvoCUA-32B ($56.7\%$), showing that model size alone does not determine computer-use performance; the coverage and quality of agent training data are also critical.

\paragraph{UI-Mate's relative strengths lie in system interaction and workflow coordination.}
Compared with Kimi-K2.6, UI-Mate-27B achieves higher scores on OS ($91.7\%$ vs.\ $79.2\%$), Office ($85.4\%$ vs.\ $80.0\%$), and Workflow tasks ($63.7\%$ vs.\ $55.0\%$), while the two models perform similarly on Daily tasks ($76.8\%$ vs.\ $77.1\%$). In contrast, UI-Mate-27B remains weaker on Professional tasks ($75.5\%$ vs.\ $81.6\%$). This performance profile indicates that UI-Mate's training transfers particularly well to operating-system control, office applications, and multi-stage workflow execution, while specialized professional software remains an important direction for improving coverage and generalization.

\subsubsection{WindowsAgentArena}

\paragraph{UI-Mate establishes the strongest open-weight performance on WindowsAgentArena.}
In Table~\ref{tab:windowsagentarena_results}, UI-Mate-27B achieves a task success rate of $66.2\%$, outperforming all open-weight baselines from 7B to 1T parameters. It surpasses the substantially larger Kimi-K2.6 ($63.3\%$) by $2.9$ percentage points and exceeds specialized computer-use agents including EvoCUA-32B ($56.5\%$), UI-TARS-1.5-7B ($42.1\%$), and ScaleCUA-Qwen3.5-9B ($38.1\%$). UI-Mate-27B also approaches frontier closed-source models, trailing Claude Sonnet 5 ($68.8\%$), Claude Opus 4.8 ($69.3\%$), and GPT-5.5 ($70.4\%$) by only $2.6$, $3.1$, and $4.2$ percentage points, respectively. These comparisons demonstrate that environment-grounded computer-use training can produce an open-weight agent competitive with substantially larger general-purpose models and leading proprietary systems.

\paragraph{UI-Mate delivers substantial gains across model scales.}
UI-Mate-9B reaches a success rate of $61.7\%$, outperforming its Qwen3.5-9B base model ($37.5\%$) by $24.2$ percentage points and ScaleCUA-Qwen3.5-9B ($38.1\%$), a specialized agent at the same scale, by $23.6$ points. Despite using only 9B parameters, it also surpasses the larger EvoCUA-32B ($56.5\%$) by $5.2$ points and comes within $1.6$ points of the 1T-parameter Kimi-K2.6. Similarly, UI-Mate-27B improves upon Qwen3.6-27B ($47.1\%$) by $19.1$ percentage points while holding model scale fixed. Together, these results show that UI-Mate's gains arise not merely from model capacity, but from the coverage, quality, and executable feedback provided by our training stack.

\subsubsection{OSWorkerBench}

\paragraph{UI-Mate delivers competitive performance at the 27B scale.}
In Table~\ref{tab:OSWorkerBench_results}, UI-Mate-27B attains 41.00\% overall binary success and a 76.86\% progress score. Relative to its Qwen3.6-27B base model, our computer-use training improves these metrics by 17.67 and 24.51 percentage points, respectively, while holding architecture and parameter count fixed. This brings UI-Mate-27B slightly ahead of Kimi-K2.6 in both end-to-end completion (41.00\% vs.\ 40.67\%) and overall progress (76.86\% vs.\ 72.42\%). UI-Mate-27B also exceeds the best prior specialized agent in overall success, although a clear gap to frontier models remains.

\paragraph{UI-Mate achieves substantial improvements at the 9B scale.}
As shown in Table~\ref{tab:OSWorkerBench_results}, UI-Mate-9B improves overall binary success from 5.05\% to 34.00\% and progress from 18.11\% to 66.55\% over its Qwen3.5-9B base model, corresponding to gains of 28.95 and 48.44 percentage points. It also substantially outperforms ScaleCUA-Qwen3.5-9B, a specialized agent built on the same base model, in both overall success (34.00\% vs.\ 16.33\%) and progress (66.55\% vs.\ 38.27\%).

\paragraph{UI-Mate is stronger on long-horizon and cross-application tasks.}
UI-Mate-27B achieves 28.57\% success on Multi-App tasks and 32.84\% on Long-Memory tasks, substantially improving over its Qwen3.6-27B base model (7.48\% and 12.94\%) and outperforming Kimi-K2.6 (18.37\% and 25.37\%). Given the near tie between UI-Mate-27B and Kimi-K2.6 on overall success, this relative advantage suggests that our training particularly benefits workflows requiring information transfer across applications and sustained state tracking.

\paragraph{Trajectory-level evidence.}
Representative trajectories illustrate this difference. In a five-application pipeline-management workflow, UI-Mate preserves record-specific attributes through the final updates, whereas Kimi-K2.6 completes most stages but fails to persist required dates. In a multi-student report-card workflow, UI-Mate maintains the correspondence among records, reports, recipients, and status updates, while Kimi-K2.6 omits required attributes from the final email. These cases point to late-stage information preservation, rather than basic interface reachability, as an important source of the performance difference.

\begin{figure*}[!t]
  \centering

  \begin{subfigure}[t]{0.55\textwidth}
    \centering
    \includegraphics[width=\linewidth]{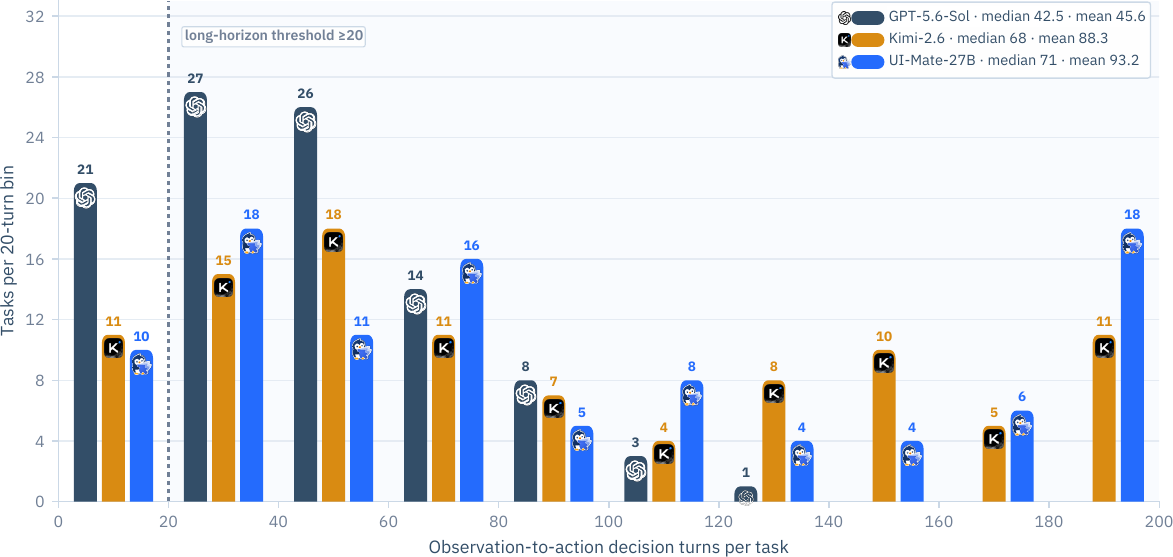}
    \caption{Trajectory-length histogram (20-turn bins)}
    \label{fig:WorkerBench_trajectory_histogram}
  \end{subfigure}
  \hfill
  \begin{subfigure}[t]{0.44\textwidth}
    \centering
    \includegraphics[width=\linewidth]{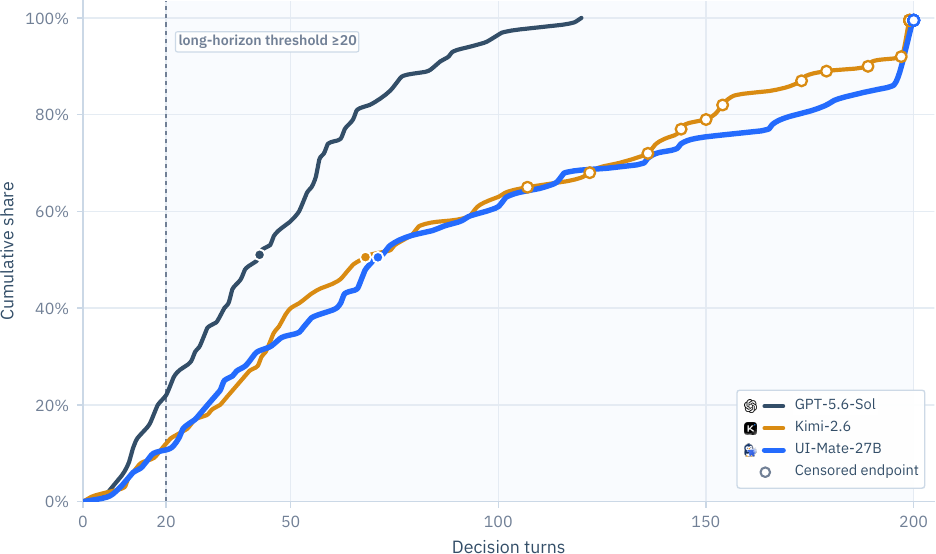}
    \caption{Cumulative distribution}
    \label{fig:WorkerBench_cumulative_distribution}
  \end{subfigure}

  \caption{\textbf{Decision-turn distributions for GPT-5.6-Sol, Kimi-K2.6, and UI-Mate-27B on the same 100 OSWorkerBench tasks.}
  A decision turn is one observation-to-action model response. A response containing multiple UI actions still counts as one turn because no new observation occurs within the batch; terminal responses and duplicate log records are excluded.
  Trajectory-length histograms using 20-turn bins show the prevalence of extended interactions, while empirical cumulative distributions summarize the full trajectory-length profiles; open triangles mark trajectories censored by the turn limit or technical failures.
  Each system has one rollout per task, and decision-turn counts characterize end-to-end interaction patterns rather than atomic UI-operation counts.}
  \label{fig:WorkerBench_trajectory_comparison}
\end{figure*}

\paragraph{Remaining failure modes.}
UI-Mate-27B's progress score exceeds its binary success rate by 35.86 percentage points, indicating substantial partial progress on tasks that are not completed end to end. The trajectory examples point to late-stage omissions, such as an unwritten field or missing final notification, as an important remaining failure mode. Closing the gap between partial progress and strict completion is therefore a primary opportunity for improvement. Long-Memory and Multi-App are overlapping task populations rather than matched variants, so their success rates characterize relative strengths but do not isolate the causal effect of any single training component.

\subsubsection{Decision-Turn Analysis on OSWorkerBench}
\label{sec:decision_turn_analysis}

\paragraph{Protocol and metric.}
Task-level scores summarize outcome quality but not the interaction horizon required to achieve it. We therefore analyze trajectories from GPT-5.6-Sol, Kimi-K2.6, and UI-Mate-27B on the same 100 OSWorkerBench tasks. A \emph{decision turn} consists of one interface observation followed by one model response. A response containing multiple UI actions still counts as a single turn because the model does not receive another observation or make another decision within the batch. Terminal responses and duplicate records are excluded. This metric therefore measures cycles of observation and decision making rather than atomic mouse or keyboard operations. All systems use screenshot observations, PyAutoGUI control, $1920{\times}1080$ environments, and a nominal 200-turn budget, with one rollout per task.

\paragraph{OSWorkerBench requires long-horizon interaction.}
Figure~\ref{fig:WorkerBench_trajectory_histogram} reports trajectory-length histograms, while Figure~\ref{fig:WorkerBench_cumulative_distribution} shows the corresponding cumulative distributions. The median trajectory contains 68 decision turns for Kimi-K2.6 and 71 for UI-Mate-27B, and 40 UI-Mate trajectories extend to at least 100 turns. Even GPT-5.6-Sol, whose trajectory distribution is shorter, requires sustained interaction on most tasks. Several Kimi and UI-Mate trajectories reach the turn limit or terminate because of technical failures, so their observed lengths are lower bounds. These distributions establish that OSWorkerBench evaluates extended workflows requiring sustained planning, state tracking, and execution rather than isolated GUI operations.

\paragraph{Shorter GPT trajectories largely reflect action batching.}
GPT's lower decision-turn count does not imply that the underlying workflows are shorter or require fewer atomic operations. A separate audit of its action logs shows that GPT produces an average of 3.83 action records per turn, with nearly half of its turns containing multiple non-wait UI actions. A substantial part of the distributional difference in Figure~\ref{fig:WorkerBench_trajectory_comparison} therefore reflects action batching and system-level interaction policies rather than a reduction in the underlying task horizon.

\paragraph{UI-Mate maintains execution over extended workflows.}
UI-Mate-27B operates across a median of 71 decision turns, with a substantial portion of its trajectories extending beyond 100 turns. Together with the outcome results in Table~\ref{tab:OSWorkerBench_results}, this distribution provides evidence that UI-Mate can preserve task state and accumulate verified progress across long, multi-stage workflows. The central conclusion is therefore not that shorter trajectories are inherently better, but that OSWorkerBench exposes long-term execution demands and that UI-Mate can operate effectively over these extended horizons.

\begin{table*}[t]
  \tablesize
  \centering
  \renewcommand{\arraystretch}{1.3}
  \caption{\textbf{Agent success rates on WindowsAgentArena.} All values are percentages, and higher is better. Our models are shaded.}
  \label{tab:windowsagentarena_results}

  \begin{tabularx}{0.75\textwidth}{
      >{\raggedright\arraybackslash}p{5.2cm}
      >{\centering\arraybackslash}X
      >{\centering\arraybackslash}X
  }
    \toprule
    \textbf{Model} & \textbf{Access / Size} & \textbf{Total Score (\%)} \\
    \midrule

    \rowcolor[gray]{0.92}
    \multicolumn{3}{l}{\emph{General-purpose multimodal models}} \\
    Kimi-K2.6                & Open, 1T-A32B & 63.3 \\
    Qwen3.6-27B             & Open, 27B     & 47.1 \\
    Qwen3.5-9B              & Open, 9B      & 37.5 \\
    GPT-5.5                 & Closed-source & 70.4 \\
    Claude Sonnet 5         & Closed-source & 68.8 \\
    Claude Opus 4.8         & Closed-source & 69.3 \\

    \midrule

    \rowcolor[gray]{0.92}
    \multicolumn{3}{l}{\emph{Specialized computer-use agents}} \\
    EvoCUA-8B               & Open, 8B           & 37.4 \\
    EvoCUA-32B              & Open, 32B           & 56.5 \\
    UI-TARS-1.5-7B          & Open, 7B            & 42.1 \\
    UI-TARS-2              & Closed-source, 230B-A23B            & 50.6 \\
    ScaleCUA-Qwen3.5-9B      & Open, 9B      & 38.1 \\

    \midrule

    \rowcolor[gray]{0.92}
    \multicolumn{3}{l}{\emph{Our models}} \\

    \rowcolor{our_blue!20}
    \textbf{UI-Mate-9B (Ours)}  & 9B  & \textbf{61.7} \\

    \rowcolor{our_blue!20}
    \textbf{UI-Mate-27B (Ours)} & 27B & \textbf{66.2} \\

    \bottomrule
  \end{tabularx}
\end{table*}

\subsection{DemoCUA Results}
\label{sec:democua-evaluation-setup}

All quantitative No-Demo versus Demo results in this subsection use the \textbf{\emph{self-demo}} setting, in which each target is paired with a demonstration of that same task.
For the 33-task OSWorker-Subset, each self-demo is a successful rollout produced by a stronger GUI agent on the corresponding target task.
This setting is separate from OSWorkerBench's \textbf{\emph{variant-demo}} resource, which contains 45 human-recorded demonstrations of related but non-identical source tasks (Section~\ref{sec:OSWorkerBench}).
We conduct preliminary variant-demo exploration on OSWorkerBench targets, where Section~\ref{sec:discussion} summarizes an observation from a 10-task pilot out of the 45 recorded demos.
We do not include variant-demo results in the aggregate tables, and leave systematic evaluation of all 45 pairs to future work.

\subsubsection{Benchmark}

We evaluate the effectiveness of DemoCUA in the self-demo setting on three benchmarks comprising 73 tasks in total: a 30-task subset of OSWorld, a newly constructed 33-task subset of OSWorkerBench, and our 10-task GameDev benchmark. 
Together, these benchmarks cover diverse applications, long-horizon interactions, and workflows involving repeated or branching subtasks, enabling a comprehensive evaluation of demonstration-guided computer use.

\paragraph{GameDev.}
We introduce a manually curated test set of 10 exceptionally long-horizon GUI tasks, each requiring more than 200 human actions on average. 
The benchmark centers on eight Godot tasks that collectively cover end-to-end 2D game development from scratch. 
In the demonstration-supported setting, we provide human-recorded demonstrations together with relevant online tutorials. Further details are provided in Appendix~\ref{sec:curated-task-cases}.

\paragraph{OSWorkerBench-Subset.}
OSWorkerBench-Subset comprises 33 multi-application tasks selected from OSWorkerBench, each requiring coordinated interaction across three to five applications. 
These tasks feature repeated subtask patterns and branching execution paths, making them particularly challenging for long-horizon agents. 
Without prior knowledge of application-specific operations and workflow structure, an agent may fail to discover necessary interactions, lose track of repeated objectives, or terminate after completing only part of the task. 
For each target, we pair the evaluated model with a successful stronger-agent rollout of that same task. This 33-task self-demo subset therefore provides a focused testbed for evaluating whether same-task workflow guidance improves task completeness.

\paragraph{OSWorld-Subset.}
OSWorld-Subset contains 30 feasible tasks that UI-Mate-27B fails without demonstration guidance but that a stronger reference agent can solve. 
This selection isolates tasks that are challenging yet demonstrably executable, enabling us to evaluate whether demonstrations can provide the procedural knowledge needed to close the capability gap. 
Tasks determined to be infeasible are excluded.

\subsubsection{Demonstration Construction.} 
We construct a demonstration for each task using a model-assisted pipeline. 
First, we deploy a capable GUI agent to interact with the target applications and record its interaction trajectories as screen-capture videos. 
We then convert the raw trajectories into structured action sequences paired with annotated screenshots. 
Human annotators subsequently refine each demonstration by (1) completing any unfinished portions of the task that the agent was unable to solve, (2) removing redundant interactions such as error-recovery loops, and (3) retaining the key actions that convey the essential workflow logic. 
The resulting demonstrations are concise and correct, cover the full scope of each task, and do not reveal task-specific answers such as underlying data values or classification labels.

\subsubsection{Evaluation Setup}

We evaluate Kimi K2.6 and UI-Mate-27B under both No-Demo and Demo conditions.
Unless explicitly stated otherwise, ``Demo'' in this subsection means self-demo.
The two conditions use identical task instructions, initialized environments, interaction budgets, and evaluators; demonstration guidance is provided only in the Demo condition. 
Both agents use the long-horizon context-management mechanism described in Section~\ref{sec:demo-inference} and are allowed up to 1,000 interaction steps per episode.

For proactive folding, each screenshot is assigned a fixed cost of 1,800 tokens when estimating context usage. Folded summaries are generated at a temperature of 0.2 and limited to 2,048 tokens. All other inference settings are held fixed between the No-Demo and Demo conditions for each agent.

Kimi K2.6 uses a 96K-token context window, with 32K tokens reserved for generation. It retains the eight most recent interaction steps verbatim and triggers context folding when estimated usage reaches 85\% of the available input budget. 
UI-Mate-27B uses a 128K-token context window, with 64K tokens reserved for generation. It retains up to 40 recent textual steps and the five most recent screenshots. Because UI-Mate-27B produces substantially longer reasoning traces, folding is triggered earlier, at 60\% of its input budget.

\subsubsection{Results}

\begin{table*}[htp]
  \tablesize
  \centering
  \renewcommand{\arraystretch}{1.3}
  \setlength{\tabcolsep}{2pt}
  \caption{
  \textbf{GameDev performance with and without demonstrations  on UI-Mate-27B.}
  Results are evaluated over five runs. 
  The two Avg.\ columns report the corresponding mean success rates, while $\Delta$ denotes the Demo - No-Demo difference in percentage points. Higher is better.
  }
  \label{tab:gamedev_demo_results}
  \begin{tabularx}{\textwidth}{
    >{\raggedright\arraybackslash}p{1.8cm}
    *{5}{>{\centering\arraybackslash}X}
    >{\columncolor[gray]{0.92}\centering\arraybackslash}X
    *{5}{>{\centering\arraybackslash}X}
    >{\columncolor[gray]{0.92}\centering\arraybackslash}X
    >{\columncolor{our_blue!20}\centering\arraybackslash}X
  }
    \toprule
    & \multicolumn{5}{c}{\textbf{No-Demo}}
    &
    & \multicolumn{5}{c}{\textbf{Demo}}
    &
    & \\
    \cline{2-6}\cline{8-12}
    \multirow{-2}{*}{\textbf{Task}}
    & \textbf{1} & \textbf{2} & \textbf{3} & \textbf{4} & \textbf{5}
    & \multirow{-2}{*}{\textbf{Avg.}}
    & \textbf{1} & \textbf{2} & \textbf{3} & \textbf{4} & \textbf{5}
    & \multirow{-2}{*}{\textbf{Avg.}}
    & \multirow{-2}{*}{$\Delta$} \\
    \midrule

    godot-01
    & 100.00 & 100.00 & 100.00 & 100.00 & 100.00
    & 100.00
    & 100.00 & 100.00 & 100.00 & 100.00 & 100.00
    & 100.00 & 0.00 \\

    godot-02
    & 71.43 & 64.29 & 78.57 & 78.57 & 57.14
    & 70.00
    & 71.43 & 71.43 & 64.29 & 71.43 & 64.29
    & 68.57 & -1.43 \\

    godot-03
    & 100.00 & 100.00 & 100.00 & 100.00 & 100.00
    & 100.00
    & 100.00 & 100.00 & 100.00 & 100.00 & 100.00
    & 100.00 & 0.00 \\

    godot-04
    & 22.22 & 38.89 & 44.44 & 38.89 & 38.89
    & 36.67
    & 55.56 & 50.00 & 55.56 & 66.67 & 50.00
    & 55.56 & 18.89 \\

    godot-05
    & 76.47 & 82.35 & 82.35 & 82.35 & 76.47
    & 80.00
    & 64.71 & 76.47 & 70.59 & 76.47 & 88.24
    & 75.29 & -4.71 \\

    godot-06
    & 86.67 & 86.67 & 80.00 & 80.00 & 80.00
    & 82.67
    & 93.33 & 93.33 & 73.33 & 86.67 & 86.67
    & 86.67 & 4.00 \\

    godot-07
    & 63.16 & 36.84 & 78.95 & 84.21 & 68.42
    & 66.32
    & 89.47 & 94.74 & 73.68 & 89.47 & 73.68
    & 84.21 & 17.89 \\

    godot-08
    & 100.00 & 100.00 & 100.00 & 100.00 & 100.00
    & 100.00
    & 100.00 & 100.00 & 100.00 & 100.00 & 100.00
    & 100.00 & 0.00 \\

    obsidian-01
    & 66.67 & 80.00 & 33.33 & 53.33 & 70.00
    & 60.67
    & 66.67 & 60.00 & 36.67 & 76.67 & 53.33
    & 58.67 & -2.00 \\

    qgis-01
    & 68.75 & 81.25 & 68.75 & 68.75 & 68.75
    & 71.25
    & 75.00 & 100.00 & 87.50 & 75.00 & 75.00
    & 82.50 & 11.25 \\

    \midrule
    \textbf{Average}
    & 75.54 & 77.03 & 76.64 & 78.61 & 75.97
    & \textbf{76.76}
    & 81.62 & 84.60 & 76.16 & 84.24 & 79.12
    & \textbf{81.15}
    & \textbf{4.39} \\
    \bottomrule
  \end{tabularx}
\end{table*}

\paragraph{GameDev.}
As shown in Table~\ref{tab:gamedev_demo_results}, demonstration guidance improves the average score of UI-Mate-27B from 76.76\% to 81.15\%, a gain of 4.39 percentage points. The largest improvements occur on godot-04 (+18.89 points), godot-07 (+17.89), and qgis-01 (+11.25), all of which require long, structured sequences of fine-grained operations. Performance remains unchanged on the three tasks already solved perfectly without demonstrations.
We observe a similar improvement with the stronger Kimi K2.6, whose results and trajectory lengths are reported in Appendix Tables~\ref{tab:kimi_gamedev_demo_results} and~\ref{tab:kimi_gamedev_steps_and_scores}.

\begin{table*}[htp]
  \tablesize
  \centering
  \setlength{\tabcolsep}{5pt}
  \renewcommand{\arraystretch}{1.28}

  \caption{
    \textbf{Paired self-demo performance on the OSWorld-Subset30 and OSWorkerBench-Subset33 with UI-Mate-27B.}
    Each target is evaluated under instruction-only and self-demo-guided
    conditions with identical initial states, budgets, and evaluators.
    The self-demo is a successful stronger-agent rollout of the same task.
    OSWorld-Subset results are averaged over five runs per target, while
    OSWorkerBench-Subset results are averaged over three runs per target.
    Paired gains are computed as self-demo guided minus instruction only
    and reported in percentage points (pp). Higher is better.
  }
  \label{tab:OSWorkerBench_demo_results}

  \begin{tabularx}{1.0\textwidth}{
    @{}
    >{\raggedright\arraybackslash}p{3.15cm}
    >{\raggedright\arraybackslash}p{2.75cm}
    *{4}{>{\centering\arraybackslash}X}
    @{}
  }
    \toprule
    \multirow{2}{*}{\makecell[l]{\textbf{Evaluation Set}}} &
    \multirow{2}{*}{\makecell[l]{\textbf{Condition}}} &
    \multicolumn{2}{c}{\textbf{Performance} (\%)} &
    \multicolumn{2}{c}{\textbf{Paired Gain} (pp)} \\
    \cmidrule(lr){3-4}
    \cmidrule(lr){5-6}
    & &
    \makecell{\textbf{Progress}\\\textbf{Score}} &
    \makecell{\textbf{Binary}\\\textbf{Success}} &
    \makecell{\textbf{$\Delta$}\\\textbf{Progress}} &
    \makecell{\textbf{$\Delta$}\\\textbf{Success}} \\
    \midrule

    \multirow{2}{*}{
      \makecell[l]{OSWorld\\\textnormal{(Subset-30)}}
    }
    & Instruction only
    & 40.27 & -- & -- & -- \\

    & \emph{Demo guided}
    & \textbf{65.75} & --
    & \textbf{$+25.48$} & -- \\

    \midrule

    \multirow{2}{*}{
      \makecell[l]{OSWorkerBench\\\textnormal{(Subset-33)}}
    }
    & Instruction only
    & 67.85 & 17.17 & -- & -- \\

    & \emph{Demo guided}
    & \textbf{81.14} & \textbf{35.35}
    & \textbf{$+13.29$} & \textbf{$+18.18$} \\

    \bottomrule
  \end{tabularx}
\end{table*}

\paragraph{OSWorkerBench-Subset.}
Table~\ref{tab:OSWorkerBench_demo_results} presents the self-demo results on the more demanding OSWorkerBench subset. 
Demonstrations improve the average normalized task score from 67.85\% to 81.14\%, a gain of 13.29 percentage points, with improvements on 28 of the 33 tasks. 
The number of tasks receiving a perfect score also increases from one to five. 
At the same time, the average trajectory length increases from 173.3 to 216.0 steps~(Table~\ref{tab:democua_osworker_steps_and_scores}). 
This increase does not simply indicate lower execution efficiency: OSWorkerBench tasks contain repetitive and branching subtasks, and agents without demonstrations frequently terminate after completing only part of the requested workflow. Demonstrations help the agent identify and execute the remaining branches or repeated operations, producing longer but more complete trajectories.

\paragraph{OSWorld-Subset.}
On the OSWorld subset, Table~\ref{tab:OSWorkerBench_demo_results} and Table~\ref{tab:osworld_subset_demo_results} in Appendix shows a substantially larger improvement: demonstrations raise the average score from 40.27\% to 65.75\%, a gain of 25.48 percentage points. 
Performance improves on 18 of the 30 tasks and remains unchanged on eight. Four tasks that receive zero score without demonstrations: chrome-02, chrome-03, multi-02, and os-01, are solved perfectly in all demonstration-conditioned runs. 
These results show that demonstrations can provide critical application-specific procedures that the agent is unlikely to discover reliably from the instruction alone. 
Nevertheless, performance decreases on four tasks, suggesting that demonstration guidance can be harmful when the demonstrated procedure is misapplied or conflicts with the current interface state.

\subsubsection{Case Study}

We present two representative cases illustrating demo-in-the-loop execution: one from our GameDev benchmark and one adapted from OSWorld2~\cite{yuan26osworld2}. 
For the latter visa-application case, we remove the requirement to invoke \texttt{ask\_user}.

\begin{figure}[t]
    \centering
    \includegraphics[width=\linewidth]{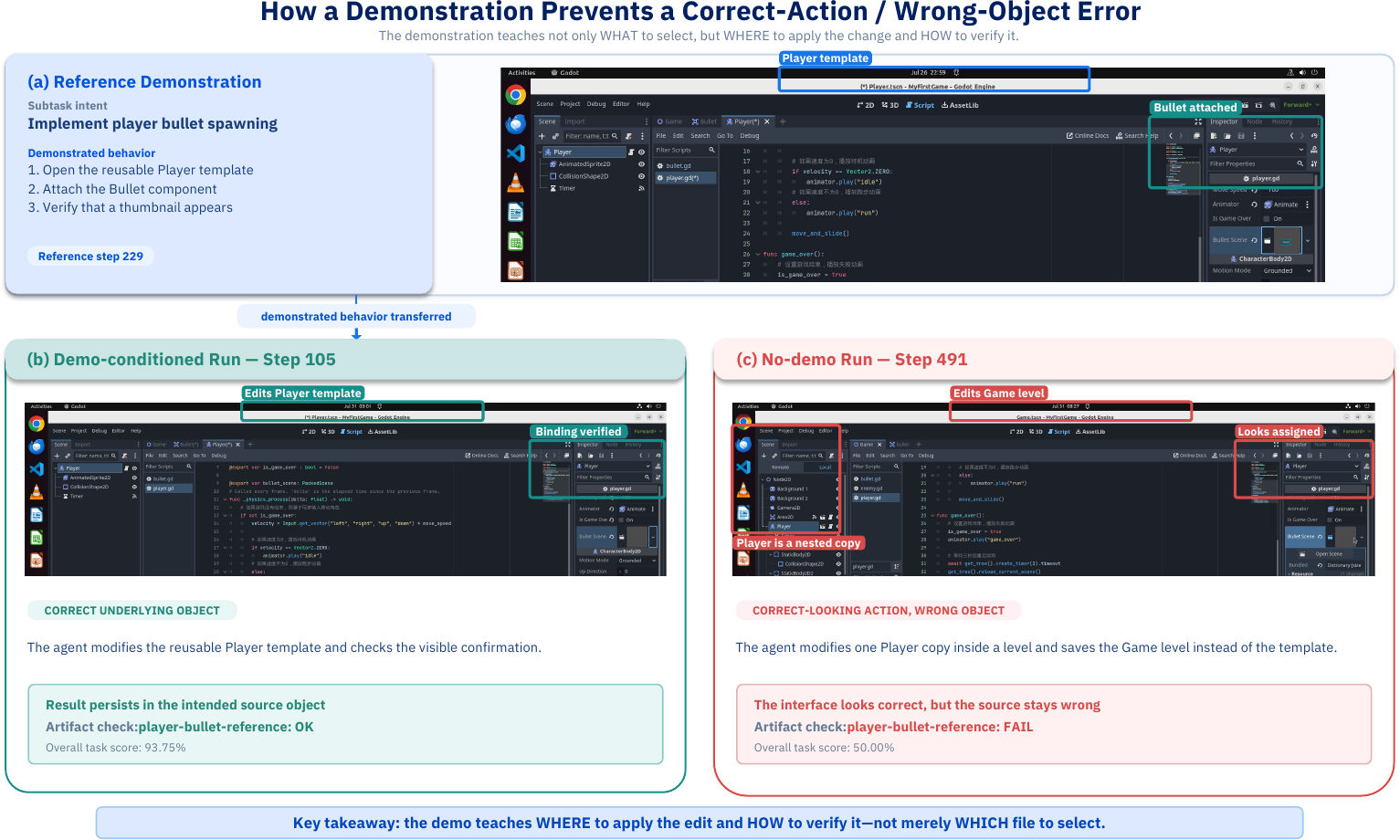}
    \caption{
        \textbf{Demonstrations help the agent modify the correct underlying object.}
        The reference demonstration and demo-conditioned run modify the reusable Player template, whereas the no-demo run modifies only one Player copy inside a game level.
        The two operations look similar in the interface, but only the former persists in the intended source object.
    }
    \label{fig:democua_godot_case}
\end{figure}

\paragraph{GameDev-Godot}
As illustrated in Fig.~\ref{fig:democua_godot_case}, the task requires configuring which bullet a game character should fire.
The no-demonstration agent correctly identified the bullet file, but attached it to a character instance in the current game level rather than to the reusable character template (Fig.~\ref{fig:democua_godot_case}(c)).
Because both operations look nearly identical in the interface, the agent incorrectly considered the task complete.
The demonstration showed not only which file to select, but also that the reusable character template should be opened, modified, and checked for a visible confirmation icon (Fig.~\ref{fig:democua_godot_case}(a)).
Consequently, the demo-conditioned agent followed the demonstrated procedure and modified the correct underlying object (Fig.~\ref{fig:democua_godot_case}(b)), whereas the no-demonstration agent performed an apparently correct operation in the wrong context.
This difference is confirmed by the final artifact evaluation: the demo-conditioned run passed the \texttt{player-bullet-reference} check, while the no-demonstration run failed it.

\begin{figure}[t]
    \centering
    \includegraphics[width=\linewidth]{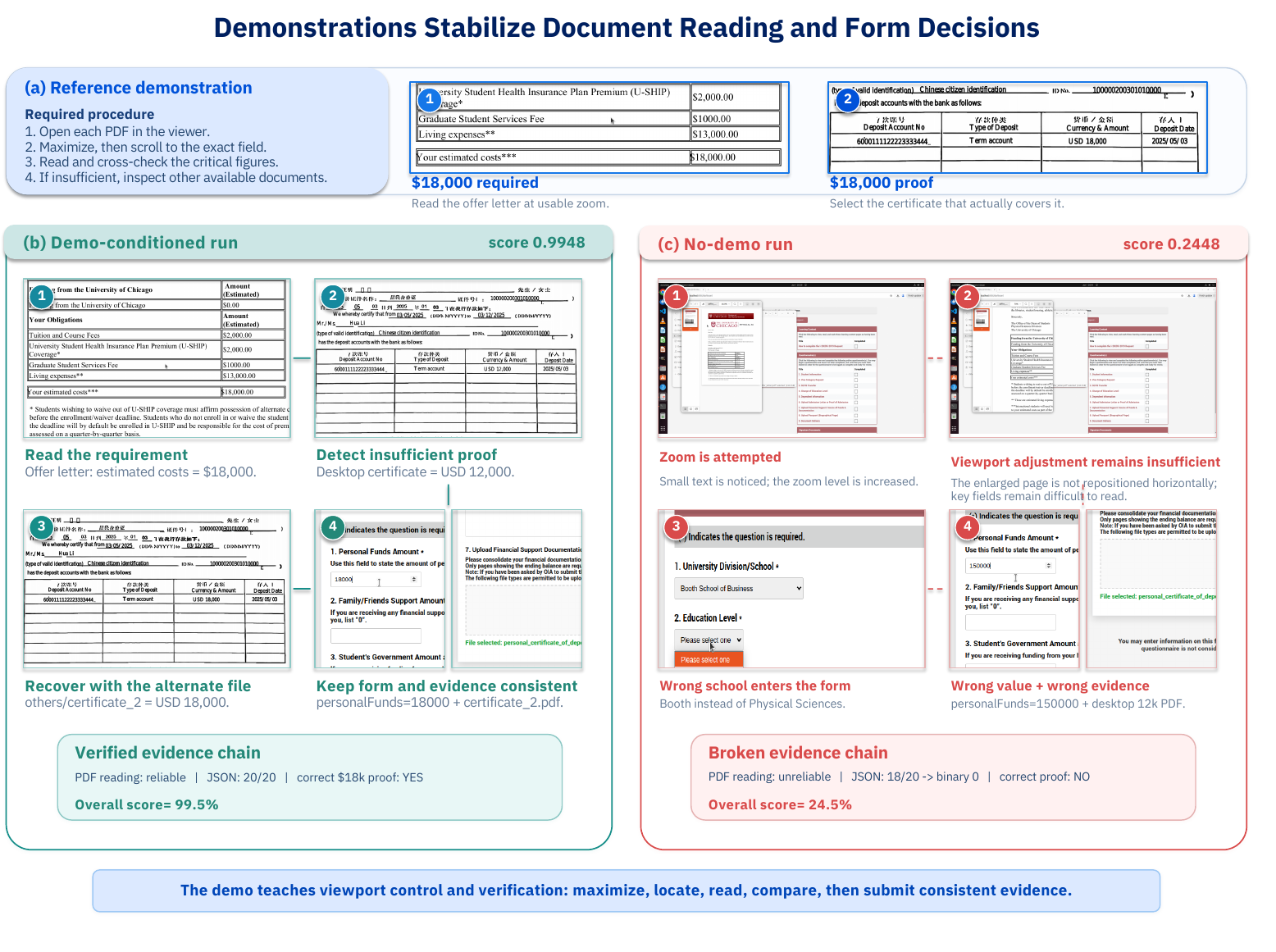}
    \caption{
        \textbf{Demonstrations teach reliable viewport control for document-grounded reasoning.}
        The reference demonstration and demo-conditioned run maximize the PDF viewer, locate and cross-check the critical values, and submit consistent form entries and supporting evidence.
        The no-demo run enlarges the document view without adequately repositioning the viewport, leading to incorrect visual/OCR readings and erroneous downstream decisions.
    }
    \label{fig:democua_visa_case}
\end{figure}

\paragraph{OSWorld2-Visa Application}
As illustrated in Fig.~\ref{fig:democua_visa_case}, the task~\cite{yuan26osworld2} requires reading the supporting PDF documents and completing a DS-2019 application with consistent form values and financial evidence.
The no-demonstration agent attempted to enlarge the document view, but did not adequately reposition the viewport to expose the relevant fields (Fig.~\ref{fig:democua_visa_case}(c)).
This resulted in incorrect visual/OCR readings and, consequently, incorrect form values and document selection.
The demonstration showed not only which documents to inspect, but also that the PDF viewer should be maximized, the viewport should be adjusted to locate the critical fields, and the extracted values should be cross-checked before submission (Fig.~\ref{fig:democua_visa_case}(a)).
Consequently, the demo-conditioned agent correctly identified the \$18,000 funding requirement, recognized that the initial \$12,000 certificate was insufficient, located the alternative \$18,000 certificate, and submitted consistent evidence (Fig.~\ref{fig:democua_visa_case}(b)).
This difference is confirmed by the final artifact evaluation: the demo-conditioned run passed all JSON checks and uploaded the correct financial certificate, achieving a score of 99.5\%, whereas the no-demonstration run submitted incorrect field values and the insufficient certificate, achieving only 24.5\%.

\subsection{Findings}

\subsubsection{Data}

\paragraph{Task difficulty lives in the environment, not the instruction.} The same
instruction can be trivial over a synthesized spreadsheet of a dozen uniform rows, but demands locating, disambiguating, and cross-checking content when the workbook is a real one with multiple sheets, merged headers, and columns irrelevant to the task. Concretely, for office-related tasks, retrieved real resources are roughly $6\times$ larger than their synthesized counterparts, and trajectories over them average $58.4$ steps against $38.5$, a 51.7\% increase. Environment realism, rather than instruction complexity alone, is therefore the primary target when we construct training data.

\paragraph{Capability-aware allocation matters more than raw data volume.}
As data production scales, uniform sampling increasingly revisits well-covered capabilities while leaving long-tail gaps obscured by application-level statistics. Thus, we route each instruction through the capability tree of Section~\ref{sec:data_capability} to a fine-grained operation and allocate data according to the coverage of the capability. This tree-guided rebalancing expands capability coverage, with the added coverage concentrated primarily in Multi-App workflows, and improves Multi-App performance by $15.5$ pp over training without capability-aware sampling. Gains are also larger on tasks whose capabilities were previously underrepresented. Although this evidence is correlational rather than a controlled isolation of capability tagging, it suggests that the tree provides an effective interface for diagnosing long-tail gaps and grouping superficially different tasks by shared operational structure. We therefore budget data by capability coverage rather than by raw trajectory count or application.

\paragraph{The hard part of a verifiable task is making its reward mean the instruction.}
The execution invariant of Equation~\ref{eq:invariant} removes inconsistent tasks but cannot certify that a reward means what its instruction says: it only checks the reward against a single reference completion, which says nothing about the other solutions it should accept or the wrong ones it should reject. An LLM audit of evaluators that had passed this filter found roughly $18\%$ still misaligned, with over-strict matching ($40\%$), semantically vacuous assertions ($23\%$), and checks applied to the wrong target ($19\%$) as the leading modes. Auditing therefore belongs in the pipeline, where flagged tasks enter repair rather than discard, as this preserves the hard, environment-rich tasks most likely to be misjudged, while what is not recovered is not promoted. We treat reward validity, not artifact checkability, as the binding constraint on synthesizing verifiable tasks at scale.

\subsubsection{Model Training}

\paragraph{Historical reasoning improves inference but undermines exploration during RL rollouts.}
Historical reasoning---preserving the reasoning traces from preceding interaction steps---consistently improves inference-time performance. Activating historical reasoning exclusively at evaluation yields gains of $+3.43$ pp for the SFT model and $+2.27$ pp for the RL model trained without historical reasoning. 
When historical reasoning is additionally incorporated during SFT, it yields a further improvement of approximately $+2.85$ pp on long-horizon, cross-application tasks, but with negligible gains on shorter and simpler tasks. Qualitative analysis attributes these improvements to more robust cross-application state tracking, consistent preservation of entity chains across Greenhouse, Gmail, Calendar, Slack, HubSpot, and QuickBooks, and stronger constraint adherence and source grounding in table aggregation and cross-application referencing. Historical reasoning also facilitates recovery from intermediate execution errors, reducing premature termination with \texttt{DONE}.

In contrast, incorporating historical reasoning into RL training affects the policy's exploration dynamics. Conditioning each decision on accumulated historical reasoning traces increases predictive confidence and accelerates the reduction of policy entropy. Empirically, this configuration exhibits signatures of entropy collapse, thereby restricting exploration during RL rollouts and resulting in lower evaluation performance. These results highlight the need for a more effective strategy to integrate historical reasoning into RL, one that preserves policy diversity without compromising its advantages for long-horizon state tracking.

\paragraph{Adaptive curriculum sampling and process credit improve training efficiency rather than final performance.}
Adaptive curriculum sampling and PCM do not consistently improve final task success over standard outcome-only training. Their main benefit is faster convergence: models using these mechanisms often reach comparable performance in fewer than half as many optimization updates. Adaptive curriculum sampling prioritizes weak domains, while PCM concentrates verifier-derived credit on decisions most relevant to success or failure. These mechanisms therefore improve how efficiently a fixed training corpus is learned, whereas further gains remain primarily constrained by data coverage and quality.

\paragraph{Smaller models benefit from explicit reasoning and staged, verifier-grounded training.} Model scale changes the training recipe that works best. During SFT, the 9B model benefits from training on trajectories with explicit reasoning, whereas the 27B model remains effective when trained on a mixture of trajectories with and without reasoning. During RL, the 9B model is more sensitive to evaluator correctness: incomplete success criteria or permissive reward proxies more readily reinforce partial or unintended behaviors. Finally, the 9B model benefits from revisiting the same tasks across multiple training stages. These observations suggest that smaller GUI agents require more explicit reasoning supervision, more carefully validated evaluators, and repeated on-policy exposure to the same task distribution.

\subsubsection{DemoCUA}

\paragraph{Subtask-level demonstrations keep the model on track.}
Providing the full task demonstration at once can confuse the model about its current progress, causing it to follow steps from later stages. 
We instead show only the active subtask in detail and summarize the others in a progress checklist. 
This keeps the model focused while preserving the overall workflow (see Figure~\ref{fig:democua_context_example}).

\paragraph{Less guidance helps training; More guidance helps inference.}
During training, extracting key actions from the current subtask forces the model to infer omitted intermediate steps from the screenshot rather than copy the demonstration workflow. 
Once trained, the model treats the demonstration as a fallible reference and resolves conflicts using the live screenshot. 
Key-action extraction is therefore unnecessary at inference; providing the full sequence is simpler, more informative, and empirically better.

\begin{table*}[t]
  \tablesize
  \centering
  \renewcommand{\arraystretch}{1.25}
  \caption{\textbf{Trajectory lengths and task scores on GameDev on UI-Mate-27B.}
  Human Demo reports the number of steps in the human demonstration, while No-Demo and Demo report the average model steps and normalized task scores over five runs. Tasks requiring longer trajectories are generally associated with lower scores.
  }
  \label{tab:gamedev_steps_and_scores}
  \begin{tabularx}{\textwidth}{
    >{\raggedright\arraybackslash}p{1.8cm}
    *{5}{>{\centering\arraybackslash}X}
  }
    \toprule
    \multirow{2}{*}{\textbf{Task}}
    & \multicolumn{1}{c}{\textbf{Human Demo}}
    & \multicolumn{2}{c}{\textbf{No-Demo}}
    & \multicolumn{2}{c}{\textbf{Demo}} \\
    \cmidrule(lr){2-2}
    \cmidrule(lr){3-4}
    \cmidrule(lr){5-6}
    & \textbf{Steps}
    & \textbf{Avg. Steps}
    & \textbf{Score (\%)}
    & \textbf{Avg. Steps}
    & \textbf{Score (\%)} \\
    \midrule

    godot-01
    & 35
    & 29.8
    & 100.00
    & 30.0
    & 100.00 \\

    godot-02
    & 223
    & 178.8
    & 70.00
    & 186.6
    & 68.57 \\

    godot-03
    & 202
    & 173.8
    & 100.00
    & 131.0
    & 100.00 \\

    godot-04
    & 247
    & 410.4
    & 36.67
    & 357.8
    & 55.56 \\

    godot-05
    & 380
    & 580.0
    & 80.00
    & 362.6
    & 75.29 \\

    godot-06
    & 323
    & 638.4
    & 82.67
    & 379.2
    & 86.67 \\

    godot-07
    & 376
    & 442.8
    & 66.32
    & 448.0
    & 84.21 \\

    godot-08
    & 46
    & 15.0
    & 100.00
    & 36.2
    & 100.00 \\

    obsidian-01
    & 255
    & 427.4
    & 60.67
    & 457.6
    & 58.67 \\

    qgis-01
    & 305
    & 139.6
    & 71.25
    & 142.2
    & 82.50 \\

    \midrule
    \textbf{Average}
    & \textbf{239.2}
    & \textbf{303.6}
    & \textbf{76.76}
    & \textbf{253.1}
    & \textbf{81.15} \\

    \bottomrule
  \end{tabularx}
\end{table*}

\paragraph{Longer trajectories are associated with greater task difficulty.} Table~\ref{tab:gamedev_steps_and_scores} shows that longer tasks tend to score lower on GameDev: godot-01 and godot-08 are finished in 29.8 and 15.0 steps at 100\%, whereas every task needing more than 400 steps stays below 83\%. 
The length of the human demonstration is a cleaner predictor of the steps the model needs (rank correlation $+0.83$), suggesting that trajectory length mainly reflects how much work a task inherently requires.

\paragraph{Demonstrations improve execution efficiency.} 
Demonstrations shorten the average trajectory from 303.6 to 253.1 steps on GameDev, a reduction of 50.5 steps or 16.6\%, while raising the average score from 76.76\% to 81.15\% (Tables~\ref{tab:gamedev_demo_results} and~\ref{tab:gamedev_steps_and_scores}), so shorter runs do not sacrifice completion. 
The saving is concentrated on the five tasks that exceed 400 steps without demonstrations, where the average falls from 499.8 to 401.0 steps, indicating that demonstrations mainly remove exploratory detours on long-horizon tasks.

\section{UI-Mate App}
\label{sec:uimate}

\subsection{Overview}
\label{sec:uimate:overview}

UI-Mate App turns a configured VLM into a computer-use agent (CUA) on the user's machine. Given a task, it observes the screen and operates applications through the mouse and keyboard. It also records demonstrations for in-context learning in the same environment. Because it drives the desktop directly, installed applications require no plugin, API or scripting interface.

\paragraph{Design.}
Two decisions define the architecture. First, \emph{the application contains no model}: it sends OpenAI-format requests to a configured endpoint, allowing one build to support every arrangement in \S\ref{sec:uimate:deployment}. Second, \emph{the agent logic is separate}: an API-connected harness handles prompting, action selection and completion judgement. Models and agent logic can thus change independently of desktop control, demonstration capture and execution visibility.

\begin{figure}[t]
  \centering
  \includegraphics[width=\linewidth]{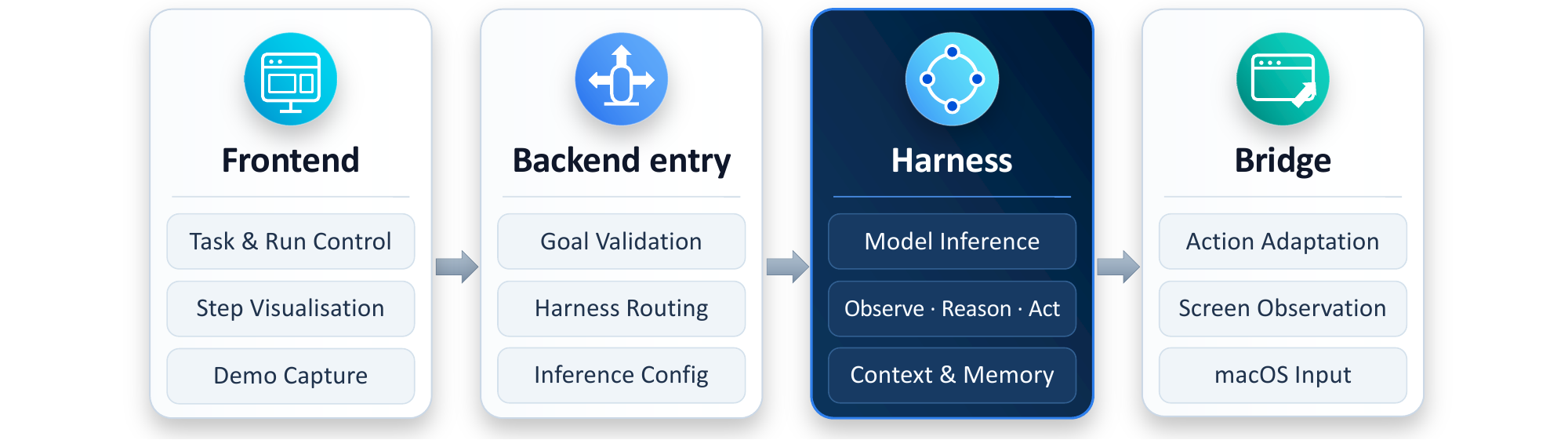}
  \caption{UI-Mate App's four-layer architecture: the frontend controls runs and demonstrations, the backend configures requests, the harness drives the agent loop, and the bridge provides macOS observation and input.}
  \label{fig:uimate:architecture}
\end{figure}

\paragraph{Architecture.}
Figure~\ref{fig:uimate:architecture} shows four layers: the \emph{frontend} controls and visualizes runs and records demonstrations; the \emph{backend entry} validates and configures requests; the \emph{harness} selects actions; and the platform-specific \emph{bridge} executes them. The same agent can therefore run in offline evaluation or a virtual machine. A stream of JSON-line events connects the frontend and backend processes, enabling live updates and mid-run commands. On macOS the bridge is a native Swift helper; Linux uses \texttt{pyautogui}, and Windows support is in progress.

\subsection{General CUA}
\label{sec:uimate:general}

\begin{figure}[t]
  \centering
  \includegraphics[width=\linewidth]{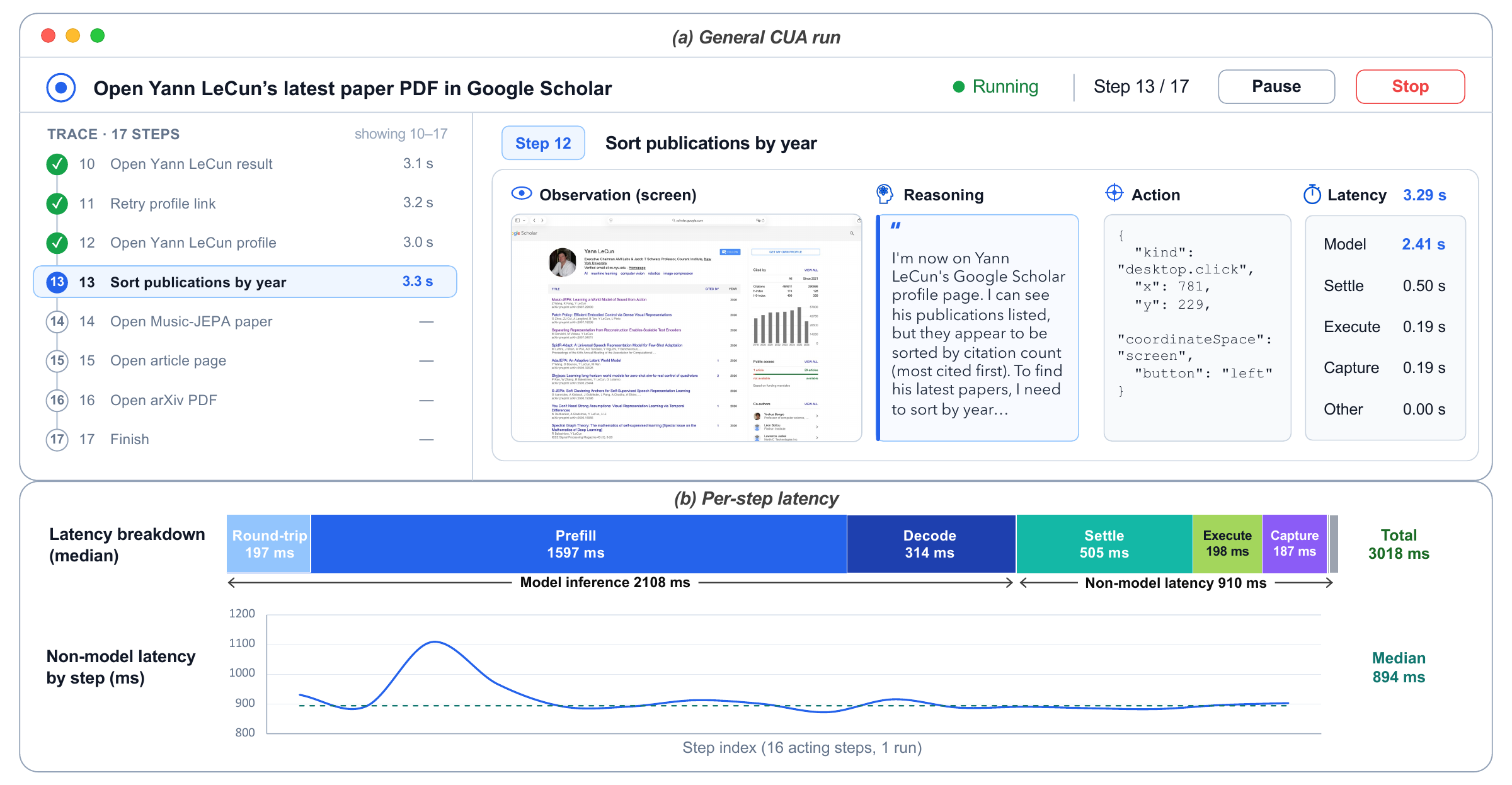}
  \caption{UI-Mate interface and runtime latency. (a) Execution trace with an expanded step. (b) Per-step latency and breakdown in a real CUA run.}
  \label{fig:uimate:general_cua}
\end{figure}

A general run executes a natural-language goal without a demonstration. The trace in Figure~\ref{fig:uimate:general_cua}(a) summarizes each step and expands to its observation, reasoning, action and latency. Users can pause, resume or stop the run and send a message for the next step. The application marks the controlled display and highlights actions while excluding its own windows from captures. Each run is saved incrementally as a JSONL trajectory of screenshots, model inputs and actions, exportable as training data or a benchmark task.

\paragraph{Step cycle.}
In each step, the application calls the model on the current screen, extracts and executes its action, waits for the interface to settle, then captures the observation for the next step. Engines such as UI-Mate and Kimi~\cite{moonshotai2026kimik26} differ in prompts and history policy but share the action space and execution path.

\paragraph{Action adaptation.}
The bridge grounds platform-neutral, \texttt{pyautogui}-style actions in the current desktop. It maps pointer locations from a 1,000-unit image grid through capture downscaling, Retina scaling and display origin into global coordinates. The macOS Accessibility (AX) API then identifies the element: actionable elements are invoked directly, tolerating small coordinate errors; otherwise the bridge simulates a click, supporting canvases and interfaces absent from AX. It also maps \textsf{ctrl} to \textsf{cmd} outside terminals.

\paragraph{Per-step latency.}
Figure~\ref{fig:uimate:general_cua}(b) profiles a run on the latency-tuned 27B endpoint of \S\ref{sec:uimate:deployment}. The median step takes $3030$\,ms: $2108$\,ms for the model call and $910$\,ms elsewhere. Prefill and decoding account for roughly $76\%$ and $15\%$ of the call; action settling, execution and screen capture dominate the remainder. Halving all non-model costs would reduce the median by only about $15\%$. UI-Mate therefore exposes image history, resolution, prompt-prefix reuse and reasoning-output controls that target prefill or decoding.

\subsection{Demo CUA}
\label{sec:uimate:demo}

Many desktop workflows involve internal tools, personal conventions or required action orders that a model does not know. A user can demonstrate such a workflow once: an offline stage converts the recording into a saved workflow, which then guides autonomous runs online.

\paragraph{Recording and Demo2Workflow.}
A native macOS recorder saves video and raw input, groups related events into actions, and extracts frames at and after each action to show its target and effect. Demo2Workflow then uses a VLM to describe each action from its frame pair and recorded facts, groups the steps into named subtasks, and lets the user revise the result. Manual edits remain separate from model output so neither overwrites the other.

\paragraph{Guided runs.}
Processed workflows form a personal library from which the user explicitly attaches one to a task. A guided run uses the general loop but also provides each step with workflow progress and the recorded steps of the current subtask. The agent reports subtask completion, advancing both the workflow and its displayed progress.

\subsection{Deployment}
\label{sec:uimate:deployment}

Model serving is separate from client installation. For each agent engine, the user configures the base URL and model name of an OpenAI-compatible endpoint, allowing one installation to switch among the arrangements in Table~\ref{tab:uimate:routes} without rebuilding.

\begin{table}[t]
  \centering
  \small
  \caption{Serving arrangements and approximate step times on our devices.}
  \label{tab:uimate:routes}
  \begin{tabular}{@{}lll@{}}
    \toprule
    \textbf{Where the model runs} & \textbf{Served as} & \textbf{Per step} \\
    \midrule
    Hosted gateway & Vendor's own & varies \\
    \addlinespace[2pt]
    Self-hosted, one 8-GPU machine & 27B, bf16 & 3--5\,s \\
                                   & 27B, ours & 2--3\,s \\
    \addlinespace[2pt]
    On device, one Apple Silicon Mac & 9B, 6-bit & 10--20\,s \\
    \bottomrule
  \end{tabular}
\end{table}

\paragraph{Accelerated self-hosted serving.}
The timings in \S\ref{sec:uimate:general} use QuaRot-style W8A8 quantization~\cite{ashkboos2024quarot} and speculative decoding with a DFlash-trained draft model~\cite{chen2026dflash}. Together they reduce a step from $3$--$5$\,s to $2$--$3$\,s on the same machine without application changes.

\paragraph{On-device serving.}
On a single Mac, visual encoding and prefill dominate. We use a six-bit 9B model of roughly $7$\,GB, reduce captures from 1080p to 720p, and patch \texttt{mlx-vlm} to reuse prompt prefixes and cached image features. These changes reduce typical step time from about $45$\,s to $10$--$20$\,s.

\paragraph{Client distribution.}
The signed and notarized disk image includes the native helper and Python runtime. After installation, UI-Mate requests the macOS Screen Recording and Accessibility permissions needed to observe and control the desktop.

\section{Related Work}

\label{sec:related}

\paragraph{Computer Use Agents.}
Computer-use agents have shifted from agentic scaffolds that compose
planning, grounding, and reflection modules around general-purpose
vision--language models~\cite{agashe2025agent,zhang2025ufo2,song2026coact}
toward native models that internalize perception, grounding, and action
end-to-end~\cite{hong2024cogagent,cheng2024seeclick,wu2025atlas,gou2025navigating,xu2024aguvis,lin2025showui}.
Native foundation agents such as UI-TARS~\cite{ui-tars2},
OpenCUA~\cite{wang2026opencua}, ScaleCUA~\cite{liu2026scalecua},
Mobile-Agent-v3~\cite{ye2025mobile}, UI-Venus-1.5~\cite{team2026ui},
and MAI-UI/Qwen-UI-Agent~\cite{zhou2025mai,zhou2026qwen_ui_agent} train on large
trajectory corpora and increasingly close the loop with environment
interaction: online and curriculum reinforcement
learning~\cite{bai2024digirl,DBLP:conf/iclr/QiLILSSYYY00D25,DBLP:journals/corr/abs-2505-16282}, multi-turn RL at
scale~\cite{ui-tars2}, EvoCUA's self-evolution from scalable synthetic
experience~\cite{xue2026evocua}, scalable environment
construction~\cite{wang2026cua}, and verifiable task synthesis with efficient
online RL~\cite{lv2026scalecuascalingcomputeruse}. Yet the resulting corpora
drift toward what is cheap to instantiate, and most standard training and
evaluation protocols condition primarily on task instructions. UI-Mate
advances both fronts: its closed-loop data engine budgets generation by a
hierarchical capability tree and promotes only audited task--verifier bundles
to RL; and it consumes in-context multimodal demonstrations, distilled from a
single human recording into a subtask-level workflow that is followed where the
user's steps carry procedural intent and overridden by the live screen where
the target task diverges, targeting forms of prompt ambiguity and execution
unreliability that scaling instruction-only training alone does not directly
address.
 
\paragraph{Computer Use Benchmarks.}
GUI-agent evaluation spans static grounding
suites~\cite{li2025screenspot}, offline web-navigation
benchmarks~\cite{deng2023mind2web}, and interactive environments with
executable, state-based verification. These environments cover the
web~\cite{zhou2024webarena,koh2024visualwebarena},
mobile devices~\cite{rawles2025androidworld,kong2026mobileworld}, and
desktop operating systems, anchored by OSWorld~\cite{xie2024osworld} and
WindowsAgentArena~\cite{bonatti2025windowsagentarena}. Recent benchmarks further
raise realism and interaction horizon:
OSWorld~2.0~\cite{yuan2026osworld2} evaluates 108 hour-scale workflows
with dynamic mid-task events, WeaveBench~\cite{li2026weavebench} forces
interleaved GUI--CLI execution under trajectory-level auditing,
ScienceBoard~\cite{sun2026scienceboard} targets scientific workflows,
and office-oriented suites assess professional deliverables across
business applications~\cite{xu2026theagentcompany,wang2024officebench}.
Most standard protocols nevertheless specify target tasks through
instructions and environment-provided artifacts, without explicitly
controlling demonstration availability for the same target. Even
tutorial-following tasks~\cite{yuan2026osworld2} test adherence to materials
supplied by the environment rather than acquisition of a user's own procedure.
OSWorkerBench addresses this gap with 100 long-horizon office tasks across 41
applications, requirement-level Long-Memory and Multi-App subsets, and
validated checkpoint evaluators. 
It separates two demonstration-guided settings: self-demos from successful strong-agent rollouts of the same tasks, and variant-demos from multimodal human demonstrations of related but non-identical tasks.
Both support controlled comparison with instruction-only execution under the same target instruction, environment, budget, and verifier.
This controlled protocol measures demonstration-guided
procedural generalization and task completion.

\paragraph{Demonstration-Guided Computer Use.}
Traditional robotic process automation (RPA) records or scripts fixed
sequences of interface operations, providing efficient and deterministic
execution for repetitive workflows but remaining brittle to changes in task
logic or UI state~\cite{vanderAalst2018robotic}. Recent work instead exposes
procedural knowledge as reusable agent skills. CUA-Skill represents
human-authored desktop procedures with parameterized execution and composition
graphs~\cite{chen2026cuaskill}, while MMSkills augments textual procedures with
state cards and visual keyframes so that agents can recognize when a skill
applies and verify its progress~\cite{zhang2026mmskills}. Most closely related,
ShowUI-Aloha converts a human screen recording into a semantic action trace
that guides planning and execution~\cite{showui_aloha}; its demonstrated
transfer primarily reuses one taught procedure across task instances sharing
the same workflow logic. In contrast, UI-Mate distills either a human recording or an agent rollout into subtask-level procedural intent, selectively follows, skips, or adapts demonstrated steps, and grounds every decision in the live interface, thereby enabling demonstration-guided procedural generalization rather than workflow replay.

\section{Discussion and Future Work}
\label{sec:discussion}

\paragraph{Verifier-Grounded Progress Credit.}
PCM's efficiency gains motivate verifier-grounded progress estimation from its milestone structure. A task-conditioned model could reward milestone completion and recovery, penalize regression, and retain signal in homogeneous-outcome groups. The environment would remain authoritative: a terminal residual would keep total process reward equal to the executable outcome. Deterministic checks would score verifiable milestones, with learned judges used only for uncertain states. Neutral spans could remain as context but be masked from actor loss, shortening the optimization horizon without losing causal evidence.

\paragraph{Beyond Self-Demo: the Variant-Demo Setting.}
All quantitative DemoCUA results in this report use the self-demo setting: the demonstration and evaluation target are the same task.
On OSWorker-Subset, this means 33 targets are paired with successful same-task rollouts from a stronger GUI agent.
These results establish the value of execution guidance, but should not be interpreted as procedural transfer across task variants.

The open problem is the variant-demo setting.
OSWorkerBench provides 45 human-recorded demonstrations whose source tasks are related to, but non-identical to, their paired targets.
We do not report a systematic aggregate result on this 45-task collection.
In a preliminary pilot on ten of these targets, replicating the demonstrated segment to match the true number of target entities makes the variant setting net positive, but performance is not yet sufficiently stable for a main benchmark claim. 
Three directions follow: 
(1) model capability, training agents to extract the transferable structure of a partially matching demonstration and to recognize what it does not cover, rather than replay it step by step; 
(2) offline acquisition, harvesting demonstrations at scale from books, curated documentation and instructional video instead of recording one per task; 
(3) retrieval, querying a demonstration corpus against the live state so that coverage grows with the corpus rather than with annotation effort.

\section{Author Contributions}
\label{sec:contributions}
We would like to express our sincere gratitude to all contributors, including those not listed in the paper, for their invaluable support and efforts. Authors within each role are listed alphabetically by their last name.

\subsection*{Core Contributors}

\begin{multicols}{3}\raggedcolumns
\begin{itemize}[leftmargin=*,label={},itemsep=1pt,topsep=0pt,parsep=0pt]
  \item Zihan Ding
  \item Longxu Dou\textsuperscript{*}
  \item Qi Gao
  \item Xiangwu Guo
  \item Shengchao Hu
  \item Zilong Huang\textsuperscript{\dag}
  \item Zihang Jiang\textsuperscript{*}
  \item Lei Ke\textsuperscript{*}
  \item Mengcheng Lan
  \item Weixian Lei\textsuperscript{*}
  \item Hanxuan Li
  \item Honglin Li
  \item Xiyun Li
  \item Zaitang Li
  \item Leowei Liang\textsuperscript{\ddag}
  \item Xin Luo
  \item Haozhe Ma
  \item Jiayi Mao
  \item Zhoujie Pan
  \item Can Qin
  \item Tianyuan Qu
  \item Weiqi Wang
  \item Wenkai Wang
  \item Yonglin Wang
  \item Yuxin Wang
  \item Chenxu Wu
  \item Yingchen Yu\textsuperscript{*}
  \item Chenyu Zhang
  \item Yuhao Zheng
\end{itemize}
\end{multicols}

\subsection*{Contributors}
\begin{multicols}{3}\raggedcolumns
\begin{itemize}[leftmargin=*,label={},itemsep=1pt,topsep=0pt,parsep=0pt]
  \item Tianqing Fang
  \item Zhenpeng Huang
  \item Zhongwei Wan
  \item Jiahao Xu
  \item Ruihan Yang
  \item Yidi Zhang
\end{itemize}
\end{multicols}

\vspace{0.5em}
\small{\noindent\textsuperscript{*}Project Co-Lead.\quad
\textsuperscript{\dag}Project Lead.\quad
\textsuperscript{\ddag}Project Supervisor.}

\bibliographystyle{unsrt}
\bibliography{paper}


\clearpage
\newpage
\beginappendix
\section{Source of Task Instructions}
\label{appendix:task_instructions}
\paragraph{Converted open-source instructions.} We adapt instructions from computer-use datasets, including AgentNet~\cite{wang2026opencua} and ScaleCUA~\cite{liu2025scalecua}, to our platforms. The selected instructions undergo both cleaning and platform conversion. We remove tasks requiring unavailable applications or containing ambiguous targets or unverifiable outcomes. We preserve operational intent while adapting application and platform references; for example, an Ubuntu application may be replaced by a Windows equivalent. This retains existing capability coverage while making instructions executable and unambiguous.

\paragraph{Atomic capabilities decomposed from existing rollouts.} Following~\cite{xue2026evocua}, we analyze failed or stalled rollouts, identify the operation or decision responsible, and isolate it as an independent subtask. This turns difficult portions of long workflows into targeted instructions for evaluating and improving atomic capabilities while directing data collection toward interactions that agents demonstrably find challenging. Decomposition also expands coverage of fundamental operating-system interactions and makes failure modes easier to evaluate in isolation.

\paragraph{Instructions grounded in real content.} Open-source and rollout-derived tasks cover atomic operations well but provide limited support for authentic cross-application workflows. We address this gap by generating instructions from real documents, spreadsheets, presentations, and static websites from InfiniteWeb~\cite{zhang2026infiniteweb}. Content previews let instructions reference concrete entities and relationships rather than artificial placeholders. Operational patterns from application manuals guide realistic long-horizon sequences, such as transferring website information into a spreadsheet and using the analysis to update a document or presentation. Grounding instructions in real content makes the tasks more realistic and their outcomes easier to evaluate.

\paragraph{Capability-tree-driven task generation.} Finally, we construct capability trees from application specifications, organizing functionality into operational modes and fine-grained capabilities. This top-down process complements collection from datasets, rollouts, and real files by surfacing underrepresented operations that common workflows might crowd out. The resulting targeted instructions systematically fill remaining capability gaps.

\newpage

\section{Details of DemoCUA Data Generation}

To supplement the three-stage data generation pipeline outlined in Section~\ref{sec:train_w_demo}, we provide fine-grained specifications for the evaluation (Score) and post-processing (Filter \& Repair) phases.

During Score, each rollout is evaluated across three dimensions via a hybrid mechanism (Table~\ref{tab:scoring_rules}). First, for trajectory-level Whole-Task Completion (S1), a VLM judge with a Skeptical-Auditor prompt assesses task success from a visual chain. Second, for subtask-level Per-Subtask Completion (S2), the VLM evaluates subtasks from boundary frames and self-reports. Third, for Demo-following adherence (S3--S4), rule-based metrics calculate Subtask Completion Rate and measure Execution-Order Alignment against the reference workflow via Longest Common Subsequence (LCS).

During Filter \& Repair, rather than discarding trajectories with fixable errors, we recover valid training data by applying deterministic and LLM-based repairs (Table~\ref{tab:repair_rules}). Deterministic program rules (R1, R2, R4) handle format normalization by collapsing duplicate reports, closing unclosed trajectories, re-deriving subtask IDs, and stripping leading WAIT or empty actions. For missing boundary reports where subtask transitions occur without explicit calls (R3), an LLM synthesizes the missing report from surrounding context.

\begin{table}[h]
\centering
\caption{%
\textbf{Scoring rules used to evaluate each rollout.}
Every rollout is assessed along three orthogonal dimensions.
$K$ is a small constant specifying how many trailing frames are appended to
the visual chain used by the VLM in rule~S1.
Let $\mathcal{S}$ be the set of subtasks defined by the workflow and
$|\mathcal{S}|$ its cardinality (i.e., the total number of subtasks).
}
\label{tab:scoring_rules}
\renewcommand{\arraystretch}{1.3}
\setlength{\tabcolsep}{6pt}
\begin{tabularx}{\textwidth}{@{}c l p{2.6cm} c X@{}}
\toprule
\textbf{ID} & \textbf{Dimension} & \textbf{Item} & \textbf{Type} & \textbf{Judging Method} \\
\midrule
S1 & Trajectory-level & \makecell[l]{Whole-Task\\Completion} & VLM &
A Skeptical-Auditor prompt is fed to the VLM together with a multi-frame
visual chain (initial $+$ uniformly sampled intermediate $+$ last-$K$ frames);
the VLM returns a binary verdict on whether the full task is completed. \\
\addlinespace[2pt]
S2 & Subtask-level & \makecell[l]{Per-Subtask\\Completion} & VLM &
For every subtask, its boundary frames and its self-report sentences are
passed to the VLM, which independently returns
$\mathrm{passed}(s)\!\in\!\{\text{True},\text{False}\}$ for each subtask
$s\!\in\!\mathcal{S}$. \\
\addlinespace[2pt]
S3 & Demo-following & \makecell[l]{Subtask\\Completion Rate} & Rule &
Fraction of subtasks judged as passed by S2, computed as
$\displaystyle
\#\{\,s\!\in\!\mathcal{S}\mid\mathrm{passed}(s)=\text{True}\,\}\,/\,|\mathcal{S}|$. \\
\addlinespace[2pt]
S4 & Demo-following & \makecell[l]{Execution-Order\\Alignment} & Rule &
Longest-common-subsequence similarity between the agent's actual visit
order $\mathbf{o}_{\text{agent}}$ (extracted from the trajectory) and the
reference visit order $\mathbf{o}_{\text{demo}}$ defined by the demo workflow,
normalized as
$\displaystyle
\mathrm{LCS}(\mathbf{o}_{\text{agent}},\,\mathbf{o}_{\text{demo}})\,/\,|\mathcal{S}|$. \\
\bottomrule
\end{tabularx}
\end{table}

\begin{table}[h]
\centering
\caption{%
\textbf{Repair rules applied to rollouts.}
Each rule targets one action/report issue, executed by a deterministic program (\textbf{Rule}) or language model (\textbf{LLM}). Subtask reports refer to \texttt{report(subtask\_id, status)} calls at boundaries.
}
\label{tab:repair_rules}
\renewcommand{\arraystretch}{1.3}
\setlength{\tabcolsep}{6pt}
\begin{tabularx}{\textwidth}{@{}c p{2.6cm} c X@{}}
\toprule
\textbf{ID} & \textbf{Item} & \textbf{Type} & \textbf{Repair Method} \\
\midrule
R1 & \makecell[l]{Duplicate\\Reports} & Rule &
Detect adjacent \texttt{report} calls that share the same
\texttt{subtask\_id} and \texttt{status}, and collapse them into one. \\
\addlinespace[2pt]
R2 & \makecell[l]{Unclosed\\Trajectory} & Rule &
If the trajectory terminates without a final
\texttt{report(subtask\_id,\,done)}, append one to close the last subtask. \\
\addlinespace[2pt]
R3 & \makecell[l]{Missing\\Boundary} & LLM &
When a subtask transition is present in the action stream but the
corresponding boundary \texttt{report} is absent, an LLM synthesizes the
missing report from surrounding context. \\
\addlinespace[2pt]
R4 & \makecell[l]{Step-Level\\Cleanup} & Rule &
Re-derive subtask boundaries from the report stream and rewrite each
step's \texttt{subtask\_id} to align with the correct segment; in the same
pass, strip any leading \texttt{WAIT} or empty action so that the first
step is action-effective. \\
\bottomrule
\end{tabularx}
\end{table}

\newpage

\section{Details of GameDev Tasks}
\label{sec:curated-task-cases}

\begingroup
\providecolor{ctGray}{RGB}{105,105,105}
\captionsetup{singlelinecheck=false,font=footnotesize,skip=2pt}
\setlength{\parindent}{0pt}
\setlength{\parskip}{0pt}
\setlength{\tabcolsep}{4pt}
\renewcommand{\arraystretch}{1.03}
\hbadness=10000

\newcommand{\ctCode}[1]{\texttt{\detokenize{#1}}}
\newcommand{\ctApp}[1]{%
  \textcolor{headingblue}{\sffamily\bfseries #1}%
}
\newcommand{\ctCase}[4]{%
  \multicolumn{2}{@{}l@{}}{%
    \begingroup
    \setlength{\fboxsep}{1pt}%
    \colorbox{headingblue!4}{%
      \parbox{\dimexpr\linewidth-2\fboxsep\relax}{%
        \fontsize{6.8pt}{7.2pt}\selectfont
        \ctApp{#2}\quad
        {\sffamily\bfseries Case #1: #3}\hfill
        {\color{ctGray}#4 checks}%
      }%
    }%
    \endgroup
  }\\[0.5pt]
}
\newcommand{\ctRubric}[2]{%
  \textcolor{headingblue}{\sffamily\bfseries #1} #2\par\vspace{0.25pt}
}

\thispagestyle{empty}
\centering
\captionof{table}{Ten curated task case studies (1/2). Instructions are verbatim; rubric entries enumerate equal-weight artifact checks.}
\label{tab:curated-tasks-1}
\begin{adjustbox}{width=\linewidth,center}
\begin{minipage}{1.17\linewidth}
{\fontsize{7.2pt}{7.8pt}\selectfont
\begin{tabularx}{\linewidth}{@{}>{\raggedright\arraybackslash}X>{\raggedright\arraybackslash}p{0.315\linewidth}@{}}
\toprule
\rowcolor{headingblue!7}
\sffamily\bfseries Original instruction (verbatim) &
\sffamily\bfseries Detailed artifact-based rubric \\
\midrule

\ctCase{1}{obsidian-01}{Appearance and workspace}{30}
Starting from the default Ubuntu desktop, install Obsidian and launch its AppImage from a terminal with the \ctCode{--password-store=basic} flag (for example, \ctCode{./Obsidian-1.12.7.AppImage --password-store=basic}), then create or open an Obsidian vault. In Obsidian, enable community plugins, install and enable the Style Settings plugin, then install and enable the AnuPpuccin theme. Make sure the Fira Code and Maple Mono NF CN fonts are actually installed on the system. In Obsidian's Appearance settings, set the interface font to Maple Mono NF CN, set the monospace font to Fira Code, select AnuPpuccin as the theme, and use the dark appearance mode. In Style Settings, configure the AnuPpuccin options as follows: set the dark theme flavor to Frappe, set the active line highlight to the border style, and set callouts to the block style. Enable file icons, the floating header, collapsed folder styling, and custom vault title styling. Set rainbow folders to the simple rainbow style, and enable rainbow coloring for folder titles, collapse icons, indentation lines, folder icons, and subfolders. Also enable the floating status bar, mini tabs, and card layout. Finally, arrange the Obsidian right sidebar into two vertical sections, with the graph view in the upper section and the outline view in the lower section. Make sure all plugin, theme, font, Style Settings, and sidebar layout changes are saved.
&
\ctRubric{Installation (6):}{Obsidian executable plus registered vault; Style Settings enabled; real plugin installation; real AnuPpuccin installation; Fira Code installed; Maple Mono NF CN installed.}
\ctRubric{Appearance (4):}{AnuPpuccin selected; dark mode; Fira Code monospace font; Maple Mono NF CN interface font.}
\ctRubric{Style Settings (16):}{Frappe; border active line; block callouts; file icons; floating header; collapsed folders; custom vault title; simple-rainbow style; rainbow titles; collapse icons; indentation lines; folder icons; subfolders; floating status bar; mini tabs; card layout.}
\ctRubric{Workspace (4):}{right sidebar exists; top/bottom split; Graph in upper branch; Outline in lower branch.}
\\
\midrule

\ctCase{2}{qgis-01}{CSV to labeled map}{16}
Starting from the default Ubuntu desktop, download and install QGIS, then download the cities dataset CSV file (\ctCode{Cities_Asia.csv}, containing Name, Latitude, and Longitude columns for 14 Asian cities) from this Google Drive folder into the Downloads folder: \url{https://drive.google.com/drive/folders/1sp-LIgRxUvkHaGEG16nToIW4NjLC77U2?usp=sharing}. In QGIS, install and enable the QuickMapServices plugin from the official plugin repository, then add the OSM Standard basemap to the map and set its transparency to 50\%. Import the CSV as a delimited-text point layer using the Longitude and Latitude fields with the WGS 84 (EPSG:4326) coordinate reference system. Style the city points using a simple marker with a point size of 4 millimeters, and enable single labels on the layer using the Name field. Export the point layer to the Desktop as an ESRI Shapefile named \ctCode{cities_asia} (producing the \ctCode{.shp/.shx/.dbf/.prj} files) in EPSG:4326, keeping the Name, Latitude, and Longitude attributes. Finally, save the QGIS project to the Desktop as \ctCode{cities_asia.qgz}. Make sure QGIS, the downloaded CSV, the installed plugin, the exported shapefile, and the saved project are all persisted to disk.
&
\ctRubric{Installation (1):}{real QGIS executable plus profile or parseable-project evidence.}
\ctRubric{Shapefile (6):}{all four files/openable layer; Point geometry; exactly 14 features; EPSG:4326; all coordinates match without longitude/latitude swap; exactly the Name/Latitude/Longitude fields and matching values.}
\ctRubric{Project (7):}{Desktop QGZ exists; project CRS is EPSG:4326; vector layer points to \ctCode{cities_asia}; 4\,mm SimpleMarker; single labels from Name; OSM Standard tile source; 50\% opacity/transparency.}
\ctRubric{Plugin (1):}{QuickMapServices both installed and enabled.}
\ctRubric{Source data (1):}{exact CSV exists in Downloads, or all core exported data is correct.}
\\
\midrule

\ctCase{3}{godot-01}{Install and initialize}{7}
Install the supplied offline Godot 4.3 Linux build from \ctCode{/home/user/Desktop/Godot_v4.3-stable_linux.x86_64.zip}. Extract the archive into \ctCode{/home/user/Desktop/Godot} so the executable is exactly \ctCode{/home/user/Desktop/Godot/Godot_v4.3-stable_linux.x86_64}, make that file executable, and launch it successfully.
\par\smallskip
In the Godot Project Manager, create a new project named MyFirstGame at \ctCode{/home/user/Desktop/myfirstgame}. Use the Godot 4.3 Forward Plus renderer and create the project in its own folder with Git version-control metadata and the normal default project files, including \ctCode{project.godot}, \ctCode{icon.svg}, \ctCode{.gitignore}, and \ctCode{.gitattributes}. Open the new project in the editor and leave all project files saved on disk. Use only the supplied offline build; do not install a different Godot version.
&
\ctRubric{Installation (3):}{dedicated Desktop directory; pinned executable at the exact path; executable reports Godot 4.3.}
\ctRubric{Project (4):}{project at the required path; name is MyFirstGame; Forward Plus configuration; default icon reference plus \ctCode{.gitignore} and \ctCode{.gitattributes}.}
\\
\midrule

\ctCase{4}{godot-02}{Assemble the game scene}{14}
Using Godot 4.3, create the Forward Plus project MyFirstGame at \ctCode{/home/user/Desktop/myfirstgame} and copy the supplied \ctCode{/home/user/Desktop/AssetBundle} folder into the project root so its Sprites, Audio, and \ctCode{Uranus_Pixel_11Px.ttf} resources are available under \ctCode{res://AssetBundle}. Keep canvas texture filtering on nearest/no filtering so the pixel art remains crisp.
\par\smallskip
Create a Scenes folder and save a main 2D scene as \ctCode{Scenes/Game.tscn}. Use a Node2D root. Add Sprite2D children named Background 1 and Background 2 that both use \ctCode{AssetBundle/Sprites/ForestBackground.png}. Place them side by side with no gap and center the combined continuous forest/road map around the world origin. Add a Camera2D at the origin with uniform zoom (2.415, 2.415) so the complete map is framed. Set \ctCode{Game.tscn} as \ctCode{run/main_scene}.
\par\smallskip
Create and save \ctCode{Scenes/Player.tscn} with a CharacterBody2D root named Player and an AnimatedSprite2D child. Create a SpriteFrames animation named idle from \ctCode{AssetBundle/Sprites/Foxy.png}: slice \ctCode{Foxy.png} along its natural, evenly spaced character-frame grid without crossing frame boundaries, use the first four cells of the first row in order, loop the animation at 12 fps, and make idle autoplay. Instance \ctCode{Player.tscn} into \ctCode{Game.tscn}, save both scenes and the project, keep the Player instance at the Game origin, and verify the main scene runs with the idle fox visible on the two-part background.
&
\ctRubric{Project (4):}{MyFirstGame/Forward Plus identity; \ctCode{Game.tscn} is main; nearest texture filter; complete sprite/audio/font asset bundle.}
\ctRubric{Scene roots (2):}{Game has Node2D root; Player has CharacterBody2D root.}
\ctRubric{Idle animation (3):}{four frames at 12 fps; autoplay; 33\(\times\)32 first-row atlas cells.}
\ctRubric{World (3):}{two ForestBackground sprites; one left of center; one right of center.}
\ctRubric{Camera/player (2):}{Camera2D at approximately \(2.4\times\) zoom; Game instances \ctCode{Player.tscn}.}
\\
\midrule

\ctCase{5}{godot-03}{Implement WASD movement}{10}
In the supplied MyFirstGame project, configure four Input Map actions with the default 0.5 deadzone: left bound to the physical A key, right to D, up to W, and down to S. Preserve the existing Game and Player scenes and idle animation.
\par\smallskip
Create \ctCode{Scripts/player.gd}, make it extend CharacterBody2D, and attach it to the Player root in \ctCode{Scenes/Player.tscn}. Declare \ctCode{@export var move_speed: float = 50.0} as the script default, then set the Player scene's Inspector override for \ctCode{move_speed} to 100 so the saved scene value differs from the script default. In \ctCode{_process(delta)}, read \ctCode{Input.get_vector("left", "right", "up", "down")}, multiply the result by \ctCode{move_speed}, assign it to \ctCode{velocity}, and call \ctCode{move_and_slide()}. Save \ctCode{player.gd}, \ctCode{Player.tscn}, and \ctCode{project.godot}, then ensure \ctCode{Scripts/player.gd} contains no debug print statements or hard-coded velocity assignment in \ctCode{_ready()}, run the game, and verify WASD moves the player in all four directions.
&
\ctRubric{Input Map (4):}{left=A; right=D; up=W; down=S using physical key codes.}
\ctRubric{Script/configuration (2):}{\ctCode{player.gd} attached to CharacterBody2D; exported speed default 50 and scene override about 100.}
\ctRubric{Movement (3):}{all four actions feed an input vector and velocity; movement runs in a frame/physics callback; \ctCode{move_and_slide()} is called.}
\ctRubric{Cleanup (1):}{no temporary debug print.}
\\
\bottomrule
\end{tabularx}
}
\end{minipage}
\end{adjustbox}

\clearpage

\thispagestyle{empty}
\centering
\captionof{table}{Ten curated task case studies (2/2). Cases are independently evaluated; all listed rubric checks have equal weight within each case.}
\label{tab:curated-tasks-2}
\begin{adjustbox}{width=\linewidth,center}
\begin{minipage}{1.17\linewidth}
{\fontsize{7.2pt}{7.8pt}\selectfont
\begin{tabularx}{\linewidth}{@{}>{\raggedright\arraybackslash}X>{\raggedright\arraybackslash}p{0.315\linewidth}@{}}
\toprule
\rowcolor{headingblue!7}
\sffamily\bfseries Original instruction (verbatim) &
\sffamily\bfseries Detailed artifact-based rubric \\
\midrule

\ctCase{6}{godot-04}{Animate running and add borders}{18}
Download and install Godot 4, then open the MyFirstGame project located at \ctCode{/home/user/Desktop/myfirstgame} (a top-down 2D game with a fox player on a forest/road map). Make the following two changes and save everything to disk.
\par\smallskip
1) Player run animation. On the player's AnimatedSprite2D, add a SpriteFrames animation named "run" using the \ctCode{Foxy.png} sprite sheet from the asset bundle on the Desktop (the AssetBundle folder at \ctCode{/home/user/Desktop/AssetBundle}): figure out the sprite sheet's grid, slice it into cells, and take the second row (the running frames) as its 6 frames, and set the playback speed to 12 fps. Keep the existing "idle" animation, and keep autoplay on idle (run must not auto-play). In \ctCode{player.gd}, run the movement in \ctCode{_physics_process}: drive velocity from the existing left/right/up/down input actions, and switch the AnimatedSprite2D between "idle" when stopped and "run" when moving, then call \ctCode{move_and_slide()}. Expose the AnimatedSprite2D to the script (e.g. an \ctCode{@export/@onready} reference) and make sure it is actually assigned to that node, so the animation calls work. Also add a CollisionShape2D to the player with a CircleShape2D whose center is moved down toward the fox's feet.
\par\smallskip
2) Map borders. In the Game scene add four WorldBoundaryShape2D static walls confining the player - near the bottom, left, and right map edges, and a top wall along the middle of the map (the forest/road divide, not the very top) - each oriented so its solid side faces inward. Group the four wall bodies under one parent Node2D and lock that node.
&
\ctRubric{Animation (6):}{"run" exists; "idle" remains; run has 6 frames; run speed passes the requested minimum; run is not autoplay; frames are sliced atlas regions.}
\ctRubric{Controller (5):}{script derives velocity from input; switches run/idle; calls \ctCode{move_and_slide()}; uses \ctCode{_physics_process}; animator reference is bound.}
\ctRubric{Collider (2):}{player collision shape exists; center is shifted toward the feet.}
\ctRubric{Borders (5):}{bottom, left, right, and middle-top boundaries are correctly placed; all four are grouped under a locked parent.}
\\
\midrule

\ctCase{7}{godot-05}{Trigger collision and game over}{17}
Create and save \ctCode{Scenes/Slime.tscn} as a reusable Area2D enemy scene. Add an AnimatedSprite2D using \ctCode{AssetBundle/Sprites/Slimer.png}, split the strip into its complete sequence of evenly sized slime poses, and create a looping animation using all frames at 12 fps with autoplay. Add a CollisionShape2D with a CircleShape2D sized to the slime's solid body and shift its center downward toward the slime's base. Create \ctCode{Scripts/enemy.gd}, attach it to the Area2D, and declare \ctCode{@export var slime_speed: float = -100.0} as the script default. Use \ctCode{slime_speed} to move the slime left at a frame-rate-independent pace during physics updates, then set the Slime scene's Inspector override for \ctCode{slime_speed} to about -50 so the saved scene value differs from the script default.
\par\smallskip
Connect the Slime Area2D \ctCode{body_entered} signal to \ctCode{enemy.gd}. In the handler, detect a CharacterBody2D player and call its \ctCode{game_over()} function. Instance one fully visible Slime in \ctCode{Game.tscn} on the road to the right of the player for testing, while preserving Y-sorted gameplay rendering.
\par\smallskip
Extend \ctCode{Player.tscn} with a looping \ctCode{game_over} animation built from the complete failure-action row in \ctCode{Foxy.png}, ordered from left to right and played at about 6 fps. In \ctCode{player.gd}, add an \ctCode{is_game_over} boolean initially false. Only process normal input, idle/run animation, and \ctCode{move_and_slide} while the game is not over. Implement \ctCode{game_over()} to set the flag, play the \ctCode{game_over} animation, await a SceneTree timer of about three seconds, and reload the current scene. Save and verify touching the slime stops player movement, shows the failure animation, and restarts.
&
\ctRubric{Slime scene (6):}{Area2D root with script; eight ordered 41\(\times\)38 first-row frames at 12 fps; autoplay; circle collider; downward collider offset; speed default \(-100\) and override about \(-50\).}
\ctRubric{Signal/movement (4):}{\ctCode{body_entered} connection; physics movement; frame-rate-independent use of exported speed; handler calls Player \ctCode{game_over()}.}
\ctRubric{Placement (2):}{Slime instance exists in Game; instance lies in the lower-right road region.}
\ctRubric{Animation (2):}{six-frame looping \ctCode{game_over}; speed about 6 fps.}
\ctRubric{Player logic (3):}{game-over state/function; normal movement guard; delayed scene reload.}
{\fontsize{6.2pt}{6.7pt}\selectfont\color{ctGray}Preserved main-scene Y-sort is an unweighted regression gate.}
\\
\midrule

\ctCase{8}{godot-06}{Fire bullets on a timer}{15}
Create \ctCode{Scenes/Bullet.tscn} with an Area2D root named Bullet. Add a Sprite2D using \ctCode{AssetBundle/Sprites/Bullet.png} and a CollisionShape2D using a RectangleShape2D fitted closely to the visible bullet sprite. Create and attach \ctCode{Scripts/bullet.gd}, and declare \ctCode{@export var bullet_speed: float = 100.0} as the script default. Set the Bullet scene's Inspector override for \ctCode{bullet_speed} to about 300 so the saved scene value differs from the script default, then move the bullet right in \ctCode{_physics_process} with \ctCode{position += Vector2(bullet_speed, 0) * delta}. In \ctCode{_ready}, await a SceneTree timer of about three seconds and queue\_free the bullet so missed shots cannot accumulate indefinitely.
\par\smallskip
In \ctCode{Scripts/player.gd}, declare \ctCode{@export var bullet_scene: PackedScene}. In \ctCode{Player.tscn}, bind that exported property to \ctCode{Bullet.tscn}. Add an automatically starting, continuously repeating Timer child with an approximately one-second trigger interval and connect timeout to \ctCode{player.gd::_on_fire}. In \ctCode{_on_fire}, return without shooting whenever velocity is not \ctCode{Vector2.ZERO} or \ctCode{is_game_over} is true. Otherwise instantiate \ctCode{bullet_scene}, place the new bullet at the player's right-side gun muzzle so it visibly emerges from the character's weapon, and add it to \ctCode{get_tree().current_scene}. Ensure \ctCode{Game.tscn} contains no pre-placed Bullet instance, save all files, and verify the stationary player fires repeatedly while moving or game-over states suppress firing and old bullets self-destruct.
&
\ctRubric{Bullet scene (7):}{Area2D root with \ctCode{bullet.gd}; \ctCode{Bullet.png}; closely fitted rectangle collider; fast scene speed override; speed-driven physics movement; timer plus \ctCode{queue_free}; lifetime about 3 seconds.}
\ctRubric{Reference (1):}{Player exports \ctCode{bullet_scene} and binds it to \ctCode{Bullet.tscn}.}
\ctRubric{Timer (4):}{autostart; about 1 second; repeating; timeout connected to \ctCode{_on_fire}.}
\ctRubric{Firing (3):}{dynamic instantiation with no pre-placed Bullet; blocked while moving/game-over; player-relative right-muzzle offset.}
\\
\midrule

\ctCase{9}{godot-07}{Add death and dynamic spawning}{19}
Add the Bullet Area2D root to a Godot group named bullet. In \ctCode{Slime.tscn}, rename the existing looping animation to idle and keep it autoplaying. Add a non-looping death animation using every pose in \ctCode{AssetBundle/Sprites/SlimerDeath.png} at the demonstrated 12 fps pace. Connect the Slime Area2D \ctCode{area_entered} signal to \ctCode{enemy.gd}.
\par\smallskip
In \ctCode{enemy.gd}, add an \ctCode{is_dead} boolean initially false and only move the slime while it is not dead. At the start of the area-entered handler, return immediately when \ctCode{is_dead} is already true. Otherwise check \ctCode{area.is_in_group("bullet")}; on the first hit play death, set \ctCode{is_dead} true, queue\_free the bullet area, await about 0.6 seconds for the death animation, and then queue\_free the slime itself. Preserve the separate \ctCode{body_entered} player-collision game-over behavior.
\par\smallskip
Create \ctCode{Scripts/GameManager.gd} extending Node2D and attach it to the Game scene root. Export \ctCode{slime_scene} as PackedScene and \ctCode{spawn_timer} as Timer, and bind them to \ctCode{Slime.tscn} and a Game Timer child. Configure that Timer with \ctCode{wait_time} 3, autostart enabled, and timeout connected to \ctCode{_spawn_slime}. Remove the old pre-placed Slime instance. In \ctCode{_spawn_slime}, instantiate \ctCode{slime_scene}, place it just beyond the map's right edge at a random height that keeps the complete sprite within the playable road band, and add it to the current scene. In \ctCode{_process}, gradually subtract about \ctCode{0.2 * delta} from \ctCode{spawn_timer.wait_time} and clamp the interval between 1 and 3 seconds. Save and verify slimes spawn with increasing frequency and bullets trigger one clean death sequence.
&
\ctRubric{Bullet group (1):}{Bullet Area2D belongs to group \ctCode{bullet}.}
\ctRubric{Spawner (9):}{GameManager attached; scene/Timer bindings; no pre-placed Slime; 3-second autostart Timer; timeout signal; instantiate/add logic; beyond-right randomized road-safe position; interval adjusted and clamped; decay about 0.2/s within 1--3 seconds.}
\ctRubric{Animations (3):}{eight-frame \ctCode{idle} at 12 fps; idle autoplay; seven-frame non-looping \ctCode{death} at 12 fps.}
\ctRubric{Death logic (6):}{\ctCode{area_entered} connection; bullet-group test; repeated-hit guard and death state; bullet and slime cleanup; about 0.6-second delay; dead slime stops moving.}
{\fontsize{6.2pt}{6.7pt}\selectfont\color{ctGray}The inherited \ctCode{body_entered} game-over path is an unweighted regression gate.}
\\
\midrule

\ctCase{10}{godot-08}{Clean up off-screen enemies}{2}
Add off-screen slime cleanup in \ctCode{Scripts/enemy.gd}. The map's left collision wall is approximately x = -235; use x = -267 as the off-screen cleanup threshold. In \ctCode{_physics_process}, after updating slime movement, call \ctCode{queue_free()} when \ctCode{position.x < -267}. Save \ctCode{enemy.gd}.
&
\ctRubric{Boundary (1):}{\ctCode{enemy.gd} checks that the slime has moved beyond the left boundary.}
\ctRubric{Cleanup order (1):}{the physics callback updates movement before freeing a slime that crossed the boundary.}
\\
\bottomrule
\end{tabularx}
}
\end{minipage}
\end{adjustbox}

\endgroup

\newpage

\section{GameDev Performance of Kimi-K2.6}

\begin{table*}[htp]
  \tablesize
  \centering
  \renewcommand{\arraystretch}{1.3}
  \setlength{\tabcolsep}{2pt}
  \caption{
  \textbf{Kimi K2.6 performance on GameDev without and with demonstrations.}
  Results are evaluated over five runs.
  The two Avg.\ columns report the corresponding mean success rates, while $\Delta$ denotes the Demo - No-Demo difference in percentage points. Higher is better.
  }
  \label{tab:kimi_gamedev_demo_results}
  \begin{tabularx}{\textwidth}{
    >{\raggedright\arraybackslash}p{1.8cm}
    *{5}{>{\centering\arraybackslash}X}
    >{\columncolor[gray]{0.92}\centering\arraybackslash}X
    *{5}{>{\centering\arraybackslash}X}
    >{\columncolor[gray]{0.92}\centering\arraybackslash}X
    >{\columncolor{our_blue!20}\centering\arraybackslash}X
  }
    \toprule
    & \multicolumn{5}{c}{\textbf{No-Demo}}
    &
    & \multicolumn{5}{c}{\textbf{Demo}}
    &
    & \\
    \cline{2-6}\cline{8-12}
    \multirow{-2}{*}{\textbf{Task}}
    & \textbf{1} & \textbf{2} & \textbf{3} & \textbf{4} & \textbf{5}
    & \multirow{-2}{*}{\textbf{Avg.}}
    & \textbf{1} & \textbf{2} & \textbf{3} & \textbf{4} & \textbf{5}
    & \multirow{-2}{*}{\textbf{Avg.}}
    & \multirow{-2}{*}{$\Delta$} \\
    \midrule

    godot-01
    & 100.00 & 100.00 & 100.00 & 100.00 & 100.00
    & 100.00
    & 100.00 & 100.00 & 100.00 & 100.00 & 100.00
    & 100.00 & 0.00 \\

    godot-02
    & 92.86 & 71.43 & 92.86 & 78.57 & 92.86
    & 85.71
    & 92.86 & 71.43 & 50.00 & 85.71 & 85.71
    & 77.14 & -8.57 \\

    godot-03
    & 100.00 & 100.00 & 100.00 & 100.00 & 100.00
    & 100.00
    & 100.00 & 100.00 & 100.00 & 100.00 & 100.00
    & 100.00 & 0.00 \\

    godot-04
    & 78.95 & 52.63 & 84.21 & 63.16 & 68.42
    & 69.47
    & 94.74 & 84.21 & 68.42 & 84.21 & 84.21
    & 83.16 & 13.68 \\

    godot-05
    & 82.35 & 41.18 & 64.71 & 82.35 & 29.41
    & 60.00
    & 70.59 & 58.82 & 70.59 & 76.47 & 70.59
    & 69.41 & 9.41 \\

    godot-06
    & 86.67 & 93.33 & 86.67 & 80.00 & 93.33
    & 88.00
    & 93.33 & 86.67 & 86.67 & 100.00 & 86.67
    & 90.67 & 2.67 \\

    godot-07
    & 84.21 & 89.47 & 36.84 & 78.95 & 84.21
    & 74.74
    & 78.95 & 78.95 & 84.21 & 89.47 & 52.63
    & 76.84 & 2.11 \\

    godot-08
    & 100.00 & 100.00 & 100.00 & 100.00 & 100.00
    & 100.00
    & 100.00 & 100.00 & 100.00 & 100.00 & 100.00
    & 100.00 & 0.00 \\

    obsidian-01
    & 80.00 & 90.00 & 90.00 & 83.33 & 90.00
    & 86.67
    & 76.67 & 90.00 & 76.67 & 96.67 & 96.67
    & 87.33 & 0.67 \\

    qgis-01
    & 75.00 & 68.75 & 68.75 & 68.75 & 68.75
    & 70.00
    & 100.00 & 100.00 & 100.00 & 100.00 & 100.00
    & 100.00 & 30.00 \\

    \midrule
    \textbf{Average}
    & 88.00 & 80.68 & 82.40 & 83.51 & 82.70
    & \textbf{83.46}
    & 90.71 & 87.01 & 83.66 & 93.25 & 87.65
    & \textbf{88.46}
    & \textbf{5.00} \\
    \bottomrule
  \end{tabularx}
\end{table*}

\begin{table*}[htp]
  \tablesize
  \centering
  \renewcommand{\arraystretch}{1.25}
  \caption{
  \textbf{Trajectory lengths and task scores for Kimi K2.6 on GameDev.}
  Human Demo reports the number of steps in the human demonstration, while No-Demo and Demo report the average model steps and normalized task scores over five runs. Tasks requiring longer trajectories are generally associated with lower scores.
  }
  \label{tab:kimi_gamedev_steps_and_scores}
  \begin{tabularx}{\textwidth}{
    >{\raggedright\arraybackslash}p{1.8cm}
    *{5}{>{\centering\arraybackslash}X}
  }
    \toprule
    \multirow{2}{*}{\textbf{Task}}
    & \multicolumn{1}{c}{\textbf{Human Demo}}
    & \multicolumn{2}{c}{\textbf{No-Demo}}
    & \multicolumn{2}{c}{\textbf{Demo}} \\
    \cmidrule(lr){2-2}
    \cmidrule(lr){3-4}
    \cmidrule(lr){5-6}
    & \textbf{Steps}
    & \textbf{Avg. Steps}
    & \textbf{Score (\%)}
    & \textbf{Avg. Steps}
    & \textbf{Score (\%)} \\
    \midrule

    godot-01
    & 35
    & 22.2
    & 100.00
    & 31.8
    & 100.00 \\

    godot-02
    & 223
    & 253.4
    & 85.71
    & 190.6
    & 77.14 \\

    godot-03
    & 202
    & 103.0
    & 100.00
    & 145.8
    & 100.00 \\

    godot-04
    & 247
    & 387.4
    & 69.47
    & 418.2
    & 83.16 \\

    godot-05
    & 380
    & 450.6
    & 60.00
    & 445.0
    & 69.41 \\

    godot-06
    & 323
    & 213.8
    & 88.00
    & 233.6
    & 90.67 \\

    godot-07
    & 376
    & 433.2
    & 74.74
    & 388.6
    & 76.84 \\

    godot-08
    & 46
    & 104.8
    & 100.00
    & 121.0
    & 100.00 \\

    obsidian-01
    & 255
    & 193.0
    & 86.67
    & 463.2
    & 87.33 \\

    qgis-01
    & 305
    & 100.4
    & 70.00
    & 136.8
    & 100.00 \\

    \midrule
    \textbf{Average}
    & \textbf{239.2}
    & \textbf{226.2}
    & \textbf{83.46}
    & \textbf{257.5}
    & \textbf{88.46} \\

    \bottomrule
  \end{tabularx}
\end{table*}

Table~\ref{tab:kimi_gamedev_demo_results} shows that Kimi K2.6 already has strong long-horizon task-completion ability without demonstrations, achieving a mean score of 83.46 and fully solving three tasks. Demonstrations raise the mean from 83.46 to 88.46, a gain of 5.00 points, and improve six of the seven tasks not already at ceiling. The largest gains are 30.00 points on QGIS, 13.68 on \texttt{godot-04}, and 9.41 on \texttt{godot-05}, indicating better coverage of fine-grained artifact requirements. Table~\ref{tab:kimi_gamedev_steps_and_scores} provides a trajectory-level view: without demonstrations, we observe that Kimi often uses CLI commands to bypass lengthy GUI sequences, allowing it to complete tasks in fewer steps on average than the pure-GUI human reference. With demonstrations that primarily follow GUI workflows, Kimi takes 257.5 steps on average but achieves higher scores, suggesting that these workflows help it better satisfy fine-grained task requirements.
\newpage

\newpage
\section{Details of the OSWorld-Subset under DemoCUA setting}

Table~\ref{tab:osworld_subset_demo_results} reports per-task results on the 30-task OSWorld subset. Self-demonstrations increase the average score from 40.27\% to 65.75\%, a gain of 25.48 percentage points. Performance improves on 18 tasks, remains unchanged on eight, and decreases on four, demonstrating a substantial overall benefit despite several cases of negative transfer.

Table~\ref{tab:osworld_subset_steps_and_scores} further compares trajectory lengths with the human demonstrations. Self-demonstrations reduce the average model trajectory from 33.3 to 30.0 steps while improving task completion. 

\begin{table*}[htp]
  \tablesize
  \centering
  \renewcommand{\arraystretch}{1.3}
  \setlength{\tabcolsep}{2pt}
  \caption{
  \textbf{OSWorld Per-task performance with and without demonstrations on UI-Mate-27B.}
  Results are evaluated over five runs.
  The two Avg.\ columns report the corresponding mean success rates, while $\Delta$ denotes the Demo - No-Demo difference in percentage points. Higher is better.
  }
  \label{tab:osworld_subset_demo_results}
  \begin{tabularx}{\textwidth}{
    >{\raggedright\arraybackslash}p{1.8cm}
    *{5}{>{\centering\arraybackslash}X}
    >{\columncolor[gray]{0.92}\centering\arraybackslash}X
    *{5}{>{\centering\arraybackslash}X}
    >{\columncolor[gray]{0.92}\centering\arraybackslash}X
    >{\columncolor{our_blue!20}\centering\arraybackslash}p{1.05cm}
  }
    \toprule
    & \multicolumn{5}{c}{\textbf{No-Demo}}
    &
    & \multicolumn{5}{c}{\textbf{Demo}}
    &
    & \\
    \cline{2-6}\cline{8-12}
    \multirow{-2}{*}{\textbf{Task}}
    & \textbf{1} & \textbf{2} & \textbf{3} & \textbf{4} & \textbf{5}
    & \multirow{-2}{*}{\textbf{Avg.}}
    & \textbf{1} & \textbf{2} & \textbf{3} & \textbf{4} & \textbf{5}
    & \multirow{-2}{*}{\textbf{Avg.}}
    & \multirow{-2}{*}{$\Delta$} \\
    \midrule

    chrome-01
    & 100.00 & 0.00 & 0.00 & 0.00 & 100.00
    & 40.00
    & 100.00 & 100.00 & 100.00 & 100.00 & 100.00
    & 100.00 & 60.00 \\

    chrome-02
    & 0.00 & 0.00 & 0.00 & 0.00 & 0.00
    & 0.00
    & 100.00 & 100.00 & 100.00 & 100.00 & 100.00
    & 100.00 & 100.00 \\

    chrome-03
    & 0.00 & 0.00 & 0.00 & 0.00 & 0.00
    & 0.00
    & 100.00 & 100.00 & 100.00 & 100.00 & 100.00
    & 100.00 & 100.00 \\

    chrome-04
    & 0.00 & 0.00 & 0.00 & 0.00 & 0.00
    & 0.00
    & 0.00 & 0.00 & 0.00 & 0.00 & 0.00
    & 0.00 & 0.00 \\

    gimp-01
    & 0.00 & 0.00 & 0.00 & 0.00 & 0.00
    & 0.00
    & 0.00 & 100.00 & 0.00 & 0.00 & 100.00
    & 40.00 & 40.00 \\

    calc-01
    & 100.00 & 100.00 & 100.00 & 100.00 & 100.00
    & 100.00
    & 100.00 & 100.00 & 100.00 & 100.00 & 100.00
    & 100.00 & 0.00 \\

    calc-02
    & 0.00 & 0.00 & 0.00 & 0.00 & 0.00
    & 0.00
    & 0.00 & 0.00 & 0.00 & 100.00 & 0.00
    & 20.00 & 20.00 \\

    calc-03
    & 100.00 & 100.00 & 100.00 & 100.00 & 100.00
    & 100.00
    & 100.00 & 100.00 & 100.00 & 100.00 & 100.00
    & 100.00 & 0.00 \\

    calc-04
    & 0.00 & 0.00 & 0.00 & 0.00 & 0.00
    & 0.00
    & 0.00 & 100.00 & 0.00 & 0.00 & 0.00
    & 20.00 & 20.00 \\

    calc-05
    & 100.00 & 100.00 & 100.00 & 100.00 & 100.00
    & 100.00
    & 100.00 & 100.00 & 100.00 & 100.00 & 100.00
    & 100.00 & 0.00 \\

    impress-01
    & 0.00 & 100.00 & 0.00 & 0.00 & 0.00
    & 20.00
    & 100.00 & 100.00 & 100.00 & 100.00 & 100.00
    & 100.00 & 80.00 \\

    impress-02
    & 100.00 & 0.00 & 0.00 & 100.00 & 100.00
    & 60.00
    & 100.00 & 100.00 & 100.00 & 100.00 & 100.00
    & 100.00 & 40.00 \\

    writer-01
    & 0.00 & 100.00 & 0.00 & 100.00 & 0.00
    & 40.00
    & 100.00 & 100.00 & 100.00 & 100.00 & 0.00
    & 80.00 & 40.00 \\

    writer-02
    & 0.00 & 100.00 & 0.00 & 0.00 & 100.00
    & 40.00
    & 0.00 & 0.00 & 0.00 & 0.00 & 0.00
    & 0.00 & $-$40.00 \\

    writer-03
    & 100.00 & 0.00 & 0.00 & 0.00 & 100.00
    & 40.00
    & 0.00 & 0.00 & 0.00 & 100.00 & 0.00
    & 20.00 & $-$20.00 \\

    multi-01
    & 0.00 & 0.00 & 0.00 & 0.00 & 0.00
    & 0.00
    & 0.00 & 0.00 & 0.00 & 0.00 & 0.00
    & 0.00 & 0.00 \\

    multi-02
    & 0.00 & 0.00 & 0.00 & 0.00 & 0.00
    & 0.00
    & 100.00 & 100.00 & 100.00 & 100.00 & 100.00
    & 100.00 & 100.00 \\

    multi-03
    & 100.00 & 100.00 & 100.00 & 100.00 & 100.00
    & 100.00
    & 0.00 & 0.00 & 0.00 & 0.00 & 0.00
    & 0.00 & $-$100.00 \\

    multi-04
    & 0.00 & 0.00 & 0.00 & 100.00 & 0.00
    & 20.00
    & 0.00 & 100.00 & 100.00 & 100.00 & 0.00
    & 60.00 & 40.00 \\

    multi-05
    & 83.99 & 58.60 & 62.49 & 84.11 & 52.29
    & 68.30
    & 93.85 & 93.85 & 93.85 & 93.85 & 93.85
    & 93.85 & 25.55 \\

    multi-06
    & 100.00 & 0.00 & 0.00 & 0.00 & 100.00
    & 40.00
    & 0.00 & 100.00 & 100.00 & 100.00 & 0.00
    & 60.00 & 20.00 \\

    multi-07
    & 0.00 & 0.00 & 0.00 & 0.00 & 0.00
    & 0.00
    & 0.00 & 0.00 & 0.00 & 0.00 & 0.00
    & 0.00 & 0.00 \\

    multi-08
    & 100.00 & 100.00 & 100.00 & 100.00 & 100.00
    & 100.00
    & 0.00 & 100.00 & 100.00 & 100.00 & 100.00
    & 80.00 & $-$20.00 \\

    multi-09
    & 0.00 & 0.00 & 0.00 & 0.00 & 0.00
    & 0.00
    & 0.00 & 0.00 & 0.00 & 0.00 & 0.00
    & 0.00 & 0.00 \\

    os-01
    & 0.00 & 0.00 & 0.00 & 0.00 & 0.00
    & 0.00
    & 100.00 & 100.00 & 100.00 & 100.00 & 100.00
    & 100.00 & 100.00 \\

    vlc-01
    & 100.00 & 100.00 & 100.00 & 100.00 & 100.00
    & 100.00
    & 100.00 & 100.00 & 100.00 & 100.00 & 100.00
    & 100.00 & 0.00 \\

    vlc-02
    & 100.00 & 100.00 & 0.00 & 0.00 & 100.00
    & 60.00
    & 100.00 & 100.00 & 100.00 & 100.00 & 100.00
    & 100.00 & 40.00 \\

    vlc-03
    & 98.70 & 0.00 & 0.00 & 0.00 & 0.00
    & 19.74
    & 98.63 & 98.63 & 98.63 & 98.70 & 98.63
    & 98.64 & 78.90 \\

    vlc-04
    & 100.00 & 100.00 & 0.00 & 100.00 & 100.00
    & 80.00
    & 100.00 & 100.00 & 100.00 & 100.00 & 100.00
    & 100.00 & 20.00 \\

    vlc-05
    & 100.00 & 100.00 & 100.00 & 100.00 & 0.00
    & 80.00
    & 100.00 & 100.00 & 100.00 & 100.00 & 100.00
    & 100.00 & 20.00 \\

    \midrule
    \textbf{Average}
    & 49.42 & 41.95 & 25.42 & 39.47 & 45.08
    & \textbf{40.27}
    & 56.42 & 73.08 & 66.42 & 73.08 & 59.75
    & \textbf{65.75}
    & \textbf{25.48} \\
    \bottomrule
  \end{tabularx}
\end{table*}

\begin{table*}[htp]
  \tablesize
  \centering
  \renewcommand{\arraystretch}{1.25}
  \caption{\textbf{Trajectory lengths and task scores with and without demonstrations on OSWorld-Subset 30 problems with UI-Mate-27B.}
  Human Demo reports the number of steps in the human demonstration, while No-Demo and Demo report the average model steps and normalized task scores over five runs. Demonstrations make the model trajectory length track the human demonstration much more closely (rank correlation with the human step count rises from 0.66 to 0.86).
  }
  \label{tab:osworld_subset_steps_and_scores}
  \begin{tabularx}{\textwidth}{
    >{\raggedright\arraybackslash}p{1.8cm}
    *{5}{>{\centering\arraybackslash}X}
  }
    \toprule
    \multirow{2}{*}{\textbf{Task}}
    & \multicolumn{1}{c}{\textbf{Human Demo}}
    & \multicolumn{2}{c}{\textbf{No-Demo}}
    & \multicolumn{2}{c}{\textbf{Demo}} \\
    \cmidrule(lr){2-2}
    \cmidrule(lr){3-4}
    \cmidrule(lr){5-6}
    & \textbf{Steps}
    & \textbf{Avg. Steps}
    & \textbf{Score (\%)}
    & \textbf{Avg. Steps}
    & \textbf{Score (\%)} \\
    \midrule

    chrome-01
    & 5
    & 9.0
    & 40.00
    & 8.8
    & 100.00 \\

    chrome-02
    & 13
    & 87.4
    & 0.00
    & 21.6
    & 100.00 \\

    chrome-03
    & 24
    & 10.8
    & 0.00
    & 18.0
    & 100.00 \\

    chrome-04
    & 17
    & 34.4
    & 0.00
    & 64.0
    & 0.00 \\

    gimp-01
    & 4
    & 4.8
    & 0.00
    & 6.6
    & 40.00 \\

    calc-01
    & 17
    & 22.8
    & 100.00
    & 21.8
    & 100.00 \\

    calc-02
    & 14
    & 25.2
    & 0.00
    & 35.6
    & 20.00 \\

    calc-03
    & 7
    & 9.0
    & 100.00
    & 10.2
    & 100.00 \\

    calc-04
    & 45
    & 26.2
    & 0.00
    & 50.6
    & 20.00 \\

    calc-05
    & 10
    & 22.6
    & 100.00
    & 15.6
    & 100.00 \\

    impress-01
    & 16
    & 55.0
    & 20.00
    & 26.4
    & 100.00 \\

    impress-02
    & 45
    & 44.8
    & 60.00
    & 54.8
    & 100.00 \\

    writer-01
    & 4
    & 12.8
    & 40.00
    & 7.0
    & 80.00 \\

    writer-02
    & 16
    & 85.2
    & 40.00
    & 31.0
    & 0.00 \\

    writer-03
    & 22
    & 40.0
    & 40.00
    & 25.6
    & 20.00 \\

    multi-01
    & 32
    & 60.8
    & 0.00
    & 48.4
    & 0.00 \\

    multi-02
    & 21
    & 36.0
    & 0.00
    & 44.0
    & 100.00 \\

    multi-03
    & 26
    & 45.0
    & 100.00
    & 50.6
    & 0.00 \\

    multi-04
    & 39
    & 84.2
    & 20.00
    & 49.6
    & 60.00 \\

    multi-05
    & 20
    & 37.4
    & 68.30
    & 25.6
    & 93.85 \\

    multi-06
    & 21
    & 19.8
    & 40.00
    & 27.0
    & 60.00 \\

    multi-07
    & 11
    & 11.4
    & 0.00
    & 15.2
    & 0.00 \\

    multi-08
    & 39
    & 69.8
    & 100.00
    & 73.6
    & 80.00 \\

    multi-09
    & 13
    & 28.4
    & 0.00
    & 22.4
    & 0.00 \\

    os-01
    & 42
    & 35.8
    & 0.00
    & 54.0
    & 100.00 \\

    vlc-01
    & 8
    & 12.8
    & 100.00
    & 15.2
    & 100.00 \\

    vlc-02
    & 3
    & 5.6
    & 60.00
    & 10.0
    & 100.00 \\

    vlc-03
    & 8
    & 18.4
    & 19.74
    & 13.0
    & 98.64 \\

    vlc-04
    & 12
    & 30.2
    & 80.00
    & 26.8
    & 100.00 \\

    vlc-05
    & 14
    & 13.8
    & 80.00
    & 25.8
    & 100.00 \\

    \midrule
    \textbf{Average}
    & \textbf{18.9}
    & \textbf{33.3}
    & \textbf{40.27}
    & \textbf{30.0}
    & \textbf{65.75} \\

    \bottomrule
  \end{tabularx}
\end{table*}

\newpage

\section{Details of the OSWorker-Subset under DemoCUA setting}

\begin{table*}[htp]
  \tablesize
  \centering
  \renewcommand{\arraystretch}{1.25}
  \caption{\textbf{Trajectory lengths and task scores for UI-Mate-27B on the 33-task OSWorkerBench self-demo subset.}
For each target, Demo provides a successful stronger-agent rollout of that same task (self-demo setting), represented by screenshots and action types without concrete pixel coordinates.
No-Demo and Demo report the average model steps and normalized task scores over three runs under otherwise identical conditions. Demos improve average task score by $13.29$ percentage points but also increase trajectory length, because these tasks involve repetitive or branching subtasks that the model may fail to infer from instructions alone and thus declares completion prematurely. The longer trajectories reflect higher task completeness rather than lower efficiency.
  }
  \vspace{-0.1em}
  \label{tab:democua_osworker_steps_and_scores}
  \begin{tabularx}{\textwidth}{
    >{\raggedright\arraybackslash}p{1.8cm}
    *{4}{>{\centering\arraybackslash}X}
  }
    \toprule
    \multirow{2}{*}{\textbf{Task}}
    & \multicolumn{2}{c}{\textbf{No-Demo}}
    & \multicolumn{2}{c}{\textbf{Demo}} \\
    \cmidrule(lr){2-3}
    \cmidrule(lr){4-5}
    & \textbf{Avg. Steps}
    & \textbf{Score (\%)}
    & \textbf{Avg. Steps}
    & \textbf{Score (\%)} \\
    \midrule

    task-01
    & 250.3
    & 95.67
    & 265.3
    & 98.33 \\

    task-02
    & 145.0
    & 78.89
    & 203.3
    & 68.33 \\

    task-03
    & 224.0
    & 53.33
    & 313.3
    & 18.33 \\

    task-04
    & 174.3
    & 43.75
    & 264.0
    & 41.67 \\

    task-05
    & 270.3
    & 79.58
    & 278.0
    & 81.25 \\

    task-06
    & 139.7
    & 93.33
    & 123.3
    & 93.33 \\

    task-07
    & 384.3
    & 0.00
    & 454.3
    & 15.67 \\

    task-08
    & 220.7
    & 93.33
    & 393.3
    & 97.67 \\

    task-09
    & 80.3
    & 86.56
    & 120.3
    & 100.00 \\

    task-10
    & 230.7
    & 43.33
    & 191.7
    & 49.17 \\

    task-11
    & 196.7
    & 72.33
    & 213.3
    & 76.00 \\

    task-12
    & 319.0
    & 61.69
    & 205.7
    & 93.35 \\

    task-13
    & 205.0
    & 76.58
    & 253.0
    & 78.48 \\

    task-14
    & 164.3
    & 32.67
    & 300.7
    & 58.00 \\

    task-15
    & 168.7
    & 98.33
    & 244.3
    & 100.00 \\

    task-16
    & 125.3
    & 54.17
    & 301.3
    & 94.78 \\

    task-17
    & 133.7
    & 57.22
    & 99.7
    & 93.33 \\

    task-18
    & 106.7
    & 52.97
    & 165.7
    & 97.92 \\

    task-19
    & 165.0
    & 72.00
    & 171.3
    & 97.33 \\

    task-20
    & 273.7
    & 63.29
    & 329.0
    & 85.62 \\

    task-21
    & 200.7
    & 62.00
    & 293.0
    & 84.00 \\

    task-22
    & 157.3
    & 48.83
    & 219.7
    & 90.00 \\

    task-23
    & 82.3
    & 78.36
    & 83.7
    & 84.17 \\

    task-24
    & 156.7
    & 68.55
    & 140.7
    & 100.00 \\

    task-25
    & 183.0
    & 58.67
    & 214.7
    & 84.33 \\

    task-26
    & 109.0
    & 93.33
    & 122.7
    & 100.00 \\

    task-27
    & 160.0
    & 53.96
    & 223.7
    & 62.50 \\

    task-28
    & 101.0
    & 98.33
    & 112.0
    & 100.00 \\

    task-29
    & 140.3
    & 46.19
    & 198.0
    & 66.81 \\

    task-30
    & 122.0
    & 100.00
    & 89.0
    & 82.22 \\

    task-31
    & 140.3
    & 56.67
    & 283.7
    & 96.19 \\

    task-32
    & 59.3
    & 82.19
    & 114.3
    & 91.26 \\

    task-33
    & 130.0
    & 82.92
    & 142.3
    & 97.50 \\

    \midrule
    \textbf{Average}
    & \textbf{173.3}
    & \textbf{67.85}
    & \textbf{216.0}
    & \textbf{81.14} \\

    \bottomrule
  \end{tabularx}
  \vspace{-0.1em}

  \begin{minipage}{\textwidth}
    \small
    \textbf{Example.}
    Task 18 requires the model to classify six inbound emails by tier and then, for each qualifying lead, update Salesforce, create calendar events, and post messages to Slack. 
    Without a demonstration, the model correctly classifies all six emails but processes only one of the two ``Hot'' leads, losing track of the second in the long interaction history before incorrectly declaring the task complete (with an average of 106.7 steps and a score of 52.97\%).
    Although the demonstration does not explicitly specify which leads should be processed, it provides a reference workflow that illustrates how the same sequence of actions should be repeated for every qualifying lead. 
    This guides the model to process both ``Hot'' leads more completely, including updating Salesforce, creating follow-up tasks, and scheduling calendar events, resulting in an average of 165.7 steps and a score of 97.92\%.
    
  \end{minipage}
  \vspace{-0.1em}
\end{table*}

\newpage


\end{document}